\documentclass[preprint,12pt]{elsarticle}
\usepackage[pdfborder={0 0 0}]{hyperref}
\usepackage{amsthm,amsmath,amsfonts,latexsym,amssymb}
\usepackage{graphics,fancyhdr,graphicx}
\usepackage[ruled,linesnumbered]{algorithm2e}
\usepackage[margin=1in]{geometry}
\usepackage{multirow,booktabs}
\usepackage[monochrome]{xcolor}
\usepackage[table]{xcolor}
\usepackage{listings}
\usepackage{tabularx}
\usepackage{placeins}
\usepackage{comment}
\usepackage{lineno}
\usepackage{lscape}
\usepackage{bm}
\usepackage{hyperref}
\usepackage{mathrsfs}
\usepackage{arydshln}
\usepackage{appendix}
\usepackage{textcomp}
\usepackage{bm}

\usepackage{booktabs}
\usepackage{pifont}

\usepackage[ruled,linesnumbered]{algorithm2e}
\usepackage{arydshln} 
\usepackage{float}
\definecolor{custom-blue}{RGB}{3,69,173}
\hypersetup{
    colorlinks = true,
    urlcolor   = blue,
    citecolor  = custom-blue,
}
\usepackage{subcaption}
\biboptions{numbers,sort&compress}
\definecolor{listinggray}{gray}{0.9}
\definecolor{lbcolor}{rgb}{0.9,0.9,0.9}
\definecolor{Darkgreen}{RGB}{0,100,0}

\makeatletter
\def\ps@pprintTitle{%
 \let\@oddhead\@empty
 \let\@evenhead\@empty
 \def\@oddfoot{}%
 \let\@evenfoot\@oddfoot}
\makeatother
\usepackage{enumitem}

\newcommand{\D}[1]{\textcolor{black}{#1}}
\newcommand{\DNew}[1]{\textcolor{blue}{#1}}

\begin{document}
\begin{frontmatter}		
\title{Learning Hidden Physics and System Parameters with Deep Operator Networks}
\author[add1]{Dibakar Roy Sarkar\fnref{fn1}}
\ead{droysar1@jh.edu}
\author[add2]{Vijay Kag\fnref{fn1}}
\ead{Vijay.Kag@in.bosch.com}
\author[add2]{Birupaksha Pal}
\ead{Birupaksha.Pal@in.bosch.com}
\author[add1]{Somdatta Goswami\corref{cor1}}
\ead{sgoswam4@jhu.edu}
\fntext[fn1]{Authors contributed equally to this work.}
\address[add1]{Johns Hopkins Whiting School of Engineering, Baltimore \\
3400 N. Charles St., MD 21218, United States}
\address[add2]{Robert Bosch Research and Technology Center, Bangalore, \\
123 Industrial layout, Karnataka 560095, India } 	
\cortext[cor1]{Corresponding author}

\begin{abstract}

\noindent Discovering hidden physical laws and identifying governing system parameters from sparse observations are central challenges in computational science and engineering. Existing data-driven methods, such as physics-informed neural networks (PINNs) and sparse regression, are limited by their need for extensive retraining, sensitivity to noise, or inability to generalize across families of partial differential equations (PDEs). In this work, we introduce two complementary frameworks based on deep operator networks (DeepONet) to address these limitations. The first, termed the Deep Hidden Physics Operator (DHPO), extends hidden-physics modeling into the operator-learning paradigm, enabling the discovery of unknown PDE terms across diverse equation families by identifying the mapping of unknown physical operators. The second is a parameter identification framework that combines pretrained DeepONet with physics-informed inverse modeling to infer system parameters directly from sparse sensor data. We demonstrate the effectiveness of these approaches on benchmark problems, including the Reaction-Diffusion system, Burgers’ equation, the 2D Heat equation, and 2D Helmholtz equation. Across all cases, the proposed methods achieve high accuracy, with relative solution errors on the order of $\mathcal{O}(10^{-2})$ and parameter estimation errors on the order of  $\mathcal{O}(10^{-3})$, even under limited and noisy observations. By uniting operator learning with physics-informed modeling, this work offers a unified and data-efficient framework for physics discovery and parameter identification, paving the way for robust inverse modeling in complex dynamical systems.
\end{abstract}

\end{frontmatter}

\nolinenumbers

\section{Introduction}
Recent advances in machine learning, coupled with increasingly sophisticated data acquisition technologies, are transforming how we model and understand complex physical systems. In domains such as fracture analysis in electromechanical systems~\cite{tarafder2023cohesive}, fluid dynamics~\cite{perdikaris2016model} in aerospace engineering, or predicting cardiovascular blood flow in medical research~\cite{kennedy2017emergence, gennisson2013ultrasound}, and failure in materials~\cite{chen2021learning}, classical first-principles models often struggle to capture nonlinear dynamics or account for incomplete knowledge of governing processes. Sparse or noisy observations further challenge conventional numerical and statistical methods. As a result, there has been a growing interest in integrating machine learning with applied mathematics to bridge these gaps and accelerate scientific discovery. The landscape of data-driven dynamical systems discovery is rich and multifaceted, encompassing diverse methodological innovations. Researchers have developed approaches such as Equation-free modeling~\cite{kevrekidis2004equation}, deep learning techniques~\cite{RaissiDHP}, non-linear regression~\cite{motulsky1987fitting}, empirical dynamic modeling~\cite{chang2017empirical}, automated dynamics inference~\cite{daniels2015automated}, symbolic regression~\cite{koza1992programming}, and Koopman analysis~\cite{brunton2021modern} to address the fundamental challenge of modeling complex systems with incomplete knowledge and limited observational data. \\
A critical frontier in this domain is discovering closed-form mathematical models of real-world systems described by partial differential equations (PDEs) and identifying crucial system parameters from sparse, spatio-temporally scattered data. Traditional approaches have struggled to bridge the gap between theoretical understanding and practical measurement. Several prominent approaches have emerged to address these challenges. Existing methods like Sparse Identification of Non-linear Dynamics~\cite{brunton2016discovering} (SINDy) and Deep Hidden Physics Models~\cite{RaissiDHP} (DHPM) have made significant strides, but they face inherent limitations. SINDy approximates system behavior using sparse linear combinations of predefined candidate terms, but the framework requires prior knowledge of system non-linearities and is sensitive to measurement noise as well as is typically designed for recovering explicit governing equations for a single dynamical system. Deep Hidden Physics Models (DHPM) offers a promising alternative by utilizing two deep neural networks. One network captures the system state, while another approximates the unknown physics of the model. The DHPM framework demonstrates remarkable stability, leveraging automatic differentiation techniques to compute gradient terms, thus effectively bypassing the numerical differentiation challenges inherent in traditional approaches. DHPM offers improved stability through dual neural networks - one capturing system state, another approximating unknown physics - but cannot identify unknown system parameters from labeled datasets. Physics-Informed Neural Networks (PINNs)~\cite{raissi2017physics,PINNreview,wang2024causality} marked a breakthrough in parameter estimation by introducing innovative functional approximation techniques. These methods simultaneously minimize both the residual loss from governing PDEs and the data loss from sparse sensor recordings, demonstrating success across diverse applications - from estimating hemodynamic parameters in cardiovascular systems~\cite{kissas2020machine} to reconstructing elasticity fields in heterogeneous materials~\cite{chen2021learning}, as well as inferring flow parameters in fluid mechanics~\cite{raissi2020hidden, zhang2023physics}. However, a fundamental limitation persists: existing methods typically require re-training and re-evaluation for each system parameter variation, limiting their generalizability and computational efficiency. \\
These challenges motivate the development of more flexible frameworks that (i) generalize across families of PDEs, (ii) infer hidden physics and system parameters simultaneously, and (iii) remain robust under sparse and noisy observations. Neural operators \cite{lu2021learning,goswami2022physics2,li2021fourier,tripura2023wavelet,raonic2023convolutional,cao2024laplace} offer a compelling solution by learning mappings between infinite-dimensional functional spaces, thereby enabling operator-level generalization across different initial conditions, boundary conditions, or source terms. Among these, deep operator networks (DeepONet) \cite{lu2021learning} are particularly appealing due to their architectural flexibility and proven success in modeling high-dimensional dynamical systems \cite{kontolati2024learning, agarwal2025multimodal, wang2025time, sarkar2025learning}. Recently, Molinaro et al. \cite{pmlr-v202-molinaro23a} proposed a Neural Inverse Operator (NIO) methodology that learns operator-to-function mappings for canonical inverse problems, including \DNew{e}lectrical \DNew{i}mpedance \DNew{t}omography, wave scattering, and seismic imaging. Their approach leverages multiple boundary measurements to characterize the observation operator and requires paired datasets of measurements and ground truth parameters generated from numerical simulations. Similarly, Long et al. \cite{pmlr-v258-long25a} introduced an invertible Fourier Neural Operator (iFNO) that jointly learns forward and inverse mappings through a reversible architecture combined with variational autoencoders, enabling bidirectional predictions given complete paired training data. While these methods demonstrate the potential of neural operators for inverse problems, they fundamentally rely on the availability of labeled parameter datasets or complete field observations for training.

In contrast, our work addresses scenarios where ground truth parameters are unavailable and only sparse sensor measurements are accessible, leveraging physics-informed constraints to enable parameter discovery directly from governing equations rather than paired supervision. In this study, we introduce two complementary physics-informed operator learning frameworks, built upon recent advances in neural operators: \vspace{-6pt}
\begin{enumerate}[leftmargin=*]
    \item The Deep Hidden Physics Operator (DHPO) network: The framework integrates DHPM with DeepONet to discover unknown governing physics directly from sparse data. Unlike DHPM, DHPO generalizes across multiple PDE families and system conditions, while maintaining interpretability by approximating the hidden terms of the governing equations. \vspace{-6pt}
    \item Physics-Informed Parameter Identification: A framework for inferring system parameters (\textit{e.g.}, viscosity, diffusivity, thermal conductivity, source function) using sparse sensor recordings. Here, a pretrained DeepONet reconstructs the solution field from limited data, which then serves as input to a physics-constrained inverse model for parameter estimation. \D{Unlike data-driven inverse operator approaches that require labeled parameter datasets, our framework addresses the more challenging scenario where ground truth parameters are unavailable during training. We leverage the known physics of the governing equations as constraints, enabling parameter identification purely from sparse sensor observations and PDE residuals without requiring supervised parameter labels.}  \vspace{-6pt}
\end{enumerate}

\noindent By uniting operator learning with physics-informed modeling, this work contributes a step toward generalizable, data-efficient solvers for inverse problems in scientific computing. The proposed frameworks provide new opportunities for tackling hidden-physics discovery and parameter inference across diverse applications in engineering and the physical sciences. \DNew{The training requirements, generalization capabilities, and key limitations of existing and proposed methods are summarized in Table \ref{tab:comparison}.}

\begin{table}[!htb]
\centering
{\color{blue}
\caption{\DNew{Comparison of methods for hidden physics discovery and parameter identification.
Existing approaches typically operate at the instance level or require fully paired input--output labels. 
In contrast, the proposed frameworks achieve operator-level generalization directly from sparse observations without requiring explicit paired labels.}}
\label{tab:comparison}
\footnotesize
\renewcommand{\arraystretch}{1.2}
\setlength{\tabcolsep}{5pt}
\begin{tabularx}{\textwidth}{@{} p{2.5cm} X X X X @{}}
\toprule
\textbf{Method} & 
\textbf{Problem} & 
\textbf{Training} & 
\textbf{Generalization} & 
\textbf{Key Limitation} \\
& \textbf{Setting} & \textbf{Requirement} & & \\
\midrule

Weak-SINDy \cite{messenger2021weak} & 
Single trajectory with sparse, noisy observations & 
Single snapshot \quad per instance & 
None \quad \quad \quad \quad (instance-specific) & 
Requires pre-specified candidate dictionary \\

\midrule

DHPM \cite{RaissiDHP} & 
Single PDE instance with sparse observations & 
Per-instance \quad \quad \quad retraining & 
None \quad \quad \quad \quad \quad(fixed ICs/BCs) & 
Limited to black-box discovery for a single PDE realization \\

\midrule

iFNO \cite{pmlr-v258-long25a} & 
Known parameterized PDE family with paired data & 
Fully paired labeled dataset 
$\{(\boldsymbol{\theta}_i, \mathbf{u}_i)\}$ & 
Within prescribed parameter space & 
Requires ground-truth parameters and dense supervision \\

\midrule

Invertible \quad  DeepONet \cite{kaltenbach2023semi} & 
Known parameterized PDE family with paired data & 
Fully paired labeled dataset 
$\{(\boldsymbol{\theta}_i, \mathbf{u}_i)\}$ & 
Within prescribed parameter space & 
Inapplicable when parameters are unknown or observations are sparse \\

\midrule

DHPO [Ours] & 
Unknown PDE terms with sparse multi-instance observations & 
Sparse data + partially known physics & 
Operator-level (within parameter space) & 
Limited to black-box physics discovery \\

\midrule

Parameter \quad  Identification Operator [Ours] & 
Known PDE structure with unknown parameters & 
Sparse data + physics constraints & 
Operator-level (within parameter space) & 
Requires known governing equation structure \\

\bottomrule
\end{tabularx}}
\end{table}
Both frameworks achieve high accuracy with limited observations, delivering relative $L_2$ errors on the order of $10^{-2}$ for hidden-physics discovery and absolute errors on the order of $10^{-3}$ for parameter identification. We demonstrate their performance on benchmark problems including the Burgers' equation, a Reaction--Diffusion system, the 2D Heat equation, \D{and the 2D Helmholtz equation} and showing that our methods remain accurate and robust even under out-of-distribution scenarios. The remainder of the paper is organized as follows. Section~\ref{sec:method} outlines the framework for building surrogates for hidden physics discovery and parameter identification. Section~\ref{sec:examples} demonstrates the applicability of the proposed framework using benchmark examples. Section~\ref{sec:discussions&conclusion} summarizes our study and highlights potential future research directions.

\section{\DNew{Methodology}}
\label{sec:method}

The application of deep learning to the physical sciences hinges on two essential capabilities: uncovering governing equations directly from data when the underlying physics is only partially known, and estimating system parameters that control the dynamics of complex processes. Conventional data-driven approaches often fall short in these tasks, as they require large amounts of training data and generalize poorly across varying boundary conditions, source terms, or material properties. To address these limitations, we propose two complementary operator-learning frameworks based on DeepONet. The first, the Deep Hidden Physics Operator (DHPO), extends hidden-physics modeling into the operator-learning paradigm, enabling the discovery of unknown PDE terms across multiple equation families. The second is a parameter identification framework that leverages pretrained DeepONet to reconstruct solution fields from sparse sensor measurements and infer system parameters in a physics-consistent manner. Together, these approaches provide a unified, data-efficient strategy for physics discovery and parameter estimation in complex dynamical systems.

\subsection{Deep Hidden Physics Operator (DHPO) - Discovering physics using DeepONet}
\label{subsec:dphm}

Building upon the concept of deep hidden physics models (DHPM) introduced by Raissi et al. \cite{RaissiDHP}, we introduce deep hidden physics operator (DHPO) - a framework that leverages the power of operator learning to discover underlying physical laws by identifying the mapping of unknown physical operators. While DHPM has proven effective in learning physics from data corresponding to a single PDE system under varying boundaries or initial conditions, our approach extends this capability to learn from multiple PDE systems simultaneously.

\begin{figure}[!htb]
    \centering
    \includegraphics[width=0.9\linewidth]{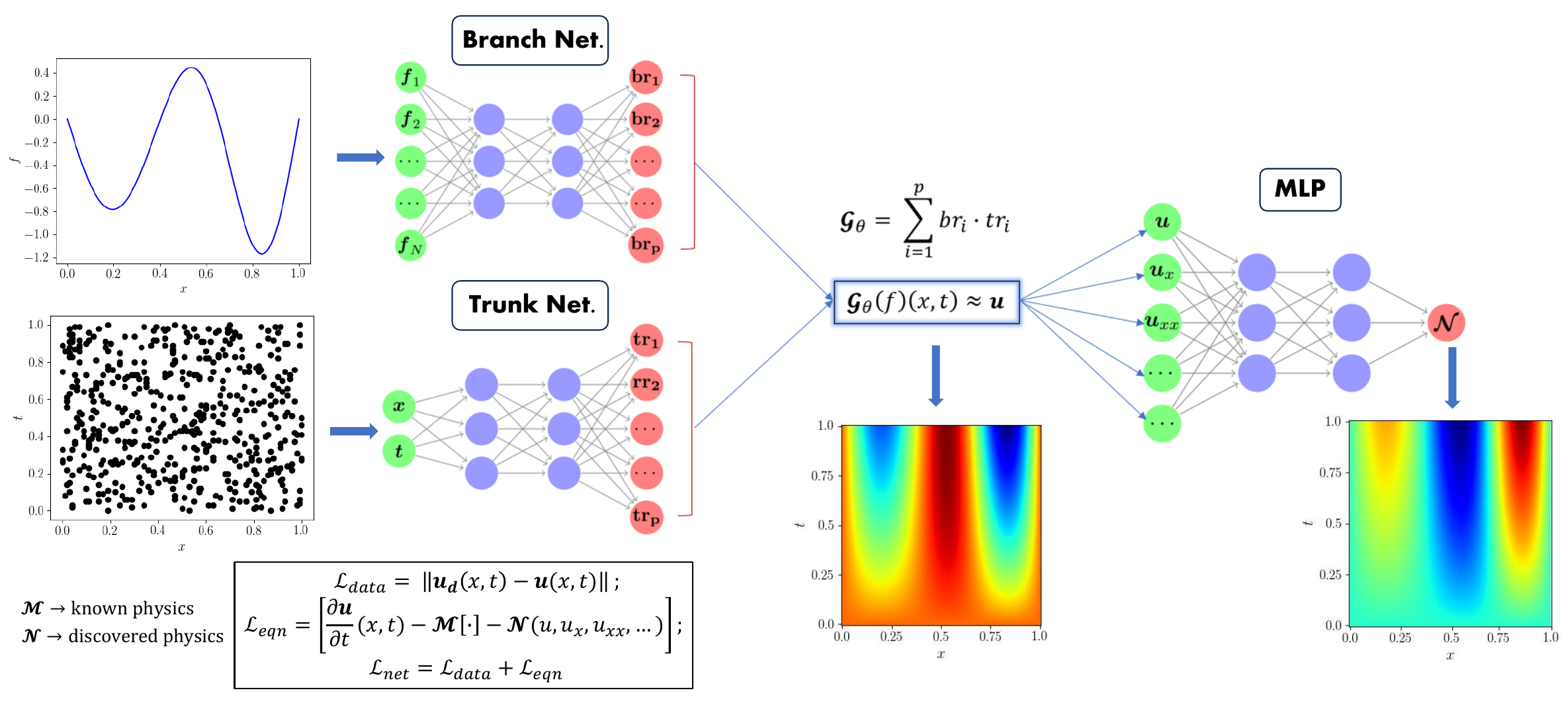}
    \caption{Architecture of the proposed Deep Hidden Physics Operator framework developed to discover the unknown physics leveraging sparsely labeled dataset. The DeepONet learns the solution operator for varying system conditions to predict the desired output. The solution as well as the gradients are considered as inputs to the MLP and it predicts the unknown physics. The entire framework is trained in a single training session with physics loss as defined in Equation~\ref{eq:eqn_loss} and the data loss.}
    \label{HPM_DON_arch}
\end{figure}

\noindent Consider a general nonlinear PDE of the form:
\begin{equation}
\frac{\partial{u}}{\partial{t}} = \mathcal{N}(t,x,u,u_x,u_{xx},..) + f(x),
\label{nonlin_eqn}
\end{equation}
for $x \in \Omega, t \in [0,T], u: \Omega \times t \rightarrow \mathbb{R}$. Here, $\mathcal{N}$ represents the unknown physics operator we aim to discover, and $f(x)$ is a variable source term. Our framework, DHPO, employs a DeepONet architecture consisting of two primary components: a branch network that processes the input function $f(x)$ at fixed sensor locations, and a trunk network that handles the spatio-temporal coordinates $(x,t)$. The solution field $u(x)$ is approximated through the interaction of these networks via a dot product operation. The gradient, $u_t, u_x, u_{xx}$, is computed using the automatic differentiation approach. To discover the underlying physics, we introduce a hidden physics neural network that approximates $\mathcal{N}$ by taking the tuple $[u, u_x, u_{xx}]$ as inputs. This network learns to represent the unknown terms in the PDE while maintaining the interpretability of the discovered physics. It is important to note that \textit{interpretability} here does not refer to symbolic equation recovery, but to mechanistic interpretability: identifying which physical processes are active, their relative contributions, and their functional dependence on solution gradients. The training of this framework is governed by a composite loss function that encompasses: \vspace{-8pt}
\begin{enumerate}[leftmargin=*]
    \item Initial condition constraints:
    \begin{equation}
    \mathcal{L}_{ic} = \frac{1}{N_{ic} b} \sum_{i=1}^{b} \sum_{j=1}^{N_{ic}}  |u_d(x_j^i ,0)- u(x_j^i ,0)|^2,
    \end{equation}
    \item Boundary condition constraints:
    \begin{equation}\label{BC_loss}
    \mathcal{L}_{bc} = \frac{1}{N_{bc} b} \sum_{i=1}^{b} \sum_{j=1}^{N_{bc}}  |u_d(0 ,t_j^i)- u(0 ,t_j^i)|^2 +  |u_d(1 ,t_j^i)- u(1 ,t_j^i)|^2,
    \end{equation}
    \item PDE residual:
    \begin{equation}\label{eq:eqn_loss}
    \mathcal{L}_{eqn} = \frac{1}{N_{Coll} b} \sum_{i=1}^{b} \sum_{j=1}^{N_{Coll}}  | \left( \frac{\partial u}{\partial t} \right)_{(x_j^i ,t_j^i)} -  \left(\mathcal{N}(u,u_x,u_{xx}) \right)_{(x_j^i ,t_j^i)} - f(x_j^i) |^2,
    \end{equation}
    \item Data fidelity:
    \begin{equation}
    \mathcal{L}_{data} = \frac{1}{N_d b} \sum_{i=1}^{b} \sum_{j=1}^{N_d}  |u(x_j^i ,t_j^i)- u_d(x_j^i ,t_j^i)|^2,
    \end{equation}
\end{enumerate}
where $u_d(x_j^i ,t_j^i)$ and $u(x_j^i ,t_j^i)$ denote the reference and predicted solutions at the $j^{th}$ point for the $i^{th}$ sample, respectively. The total loss function combines these components and is defined as:
\begin{equation}
\mathcal{L}_{total} = \mathcal{L}_{ic} + \mathcal{L}_{bc} + \mathcal{L}_{eqn} + \mathcal{L}_{data}.
\end{equation}
The parameters of the trunk, branch, and hidden physics networks are optimized simultaneously by minimizing $\mathcal{L}_{total}$. A schematic of the DHPO framework is shown in Figure \ref{HPM_DON_arch} and the detailed training procedure is presented in Algorithm \ref{training_algorithm_hidden_physics}.

Our framework improves upon traditional DHPM through enhanced generalization that simultaneously handles multiple PDE systems across diverse input functions, boundary conditions, and geometries without retraining. It achieves improved data efficiency by leveraging physics-informed constraints and knowledge transfer, enabling learning from sparse or noisy measurements. Computational advantages emerge from physics-guided optimization, leading to faster convergence, better-conditioned loss landscapes, and reduced computational resources, delivering stable and consistent predictions across varying data scenarios and out-of-distribution conditions.

\begin{algorithm}[!htb]
\caption{Training Scheme for the \DNew{D}eep Hidden Physics Operator (DHPO).}
\label{training_algorithm_hidden_physics}
\SetAlgoLined
\textbf{Inputs:} sensor measurements $f(x^s)$ and $u(x^{s'},t^{s'})$ at $n$ fixed locations, collocation points $x,t$ , partially known physics $\mathcal{M}$, initial conditions, boundary conditions. $f(x)$ can be IC/BC/source or forcing $g(x)$. \\

Initialize DeepONet $\mathcal{G}_{\bm{\theta}}$ with: \\
\hspace{1cm}Branch network: process sensor inputs $f(x^s)$\\
\hspace{1cm}Trunk network: process coordinates $(x, t)$ \\
Initialize MLP network for physics discovery with $\DNew{\bm\phi}$ \\
\While{not converged}{
    \For{each batch of sensor data}{
        {\it{Forward pass through DeepONet}}\\
        $u_{\text{pred}}$ = $\mathcal{G}_{\bm{\theta}}(f(x^s)(x,t))$\\
        $u_{\text{ic}}$ = $\mathcal{G}_{\bm{\theta}}(f(x^s)(x,0))$\\
        $u_{\text{bc}}$ = $\mathcal{G}_{\bm{\theta}}(f(x^s)(x,t))$\\
        {\it{Compute gradients}} \\
        $u_{x_{\text{pred}}} = d u_{\text{pred}}/dx$ \\
        $u_{xx_{\text{pred}}} = d^2 u_{\text{pred}}/dx^2$ \\
        $\cdots$ \\
        {\it{Forward pass through MLP}} \\
        $\mathcal{N} = \text{MLP}(u_{\text{pred}}, u_{x_{\text{pred}}}, u_{xx_{\text{pred}}}, ...)$
        
        {\it{Compute loss}} \\
        $\mathcal{L}^{\text{pde}} = \frac{1}{N} \sum_{i=1}^N || \mathcal{M}_i - \mathcal{N}_i - g(x)_i ||^2$ ~~{\it{physics loss}}\\
        $\mathcal{L}^{\text{ic}} = \frac{1}{N} \sum_{i=1}^N || u_{\text{ic}_i} - u(x,0)||^2$ ~~{\it{initial condition loss}}\\
        $\mathcal{L}^{\text{bc}} = \frac{1}{N} \sum_{i=1}^N || u_{\text{bc}_i} - u(x,t)||^2$ ~~{\it{boundary condition loss}}\\
        $\mathcal{L}^{\text{data}} = \frac{1}{N} \sum_{i=1}^N || u_{\text{pred}_i} - u(x^{s'},t^{s'})||^2$ ~~{\it{data loss}}\\
        $\mathcal{L}(\bm\theta,\DNew{\bm\phi}) = \mathcal{L}^{\text{pde}}(\bm\theta,\DNew{\bm\phi})(x,t) + \mathcal{L}^{\text{ic}}(\bm\theta)(x,0) + \mathcal{L}^{\text{bc}}(\bm\theta)(x,t) + \mathcal{L}^{\text{data}}(\bm\theta)(x^{s'},t^{s'})$ \\
        Backpropagate and update $\bm\theta$ and $\DNew{\bm\phi}$ using Adam
    }
}
{\it{Final optimized $\DNew{\bm\theta}$, $\DNew{\bm\phi}$}}\\
$\DNew{\bm\theta^*}$, $\DNew{\bm\phi^*}$ = argmin $\mathcal{L}(\DNew{\bm\theta}, \DNew{\bm\phi})$
\end{algorithm}

\D{We have considered the loss tolerance} $\epsilon$ of $\mathcal{O}(10^{-3})$. To measure the deviation of the predicted solution from reference we use metric relative $L_2$ error and mean absolute error. 

\begin{equation}\label{l2_error_def}
\text{Relative} \ L_2 \ \text{Error} = \frac{\sqrt{\sum_{i=1}^{N} |u_i- u_i^*|^2 }} {\sqrt{\sum_{i=1}^{N} |u_i^*|^2 }},
\end{equation}
\begin{equation}\label{mae_def}
\text{Mean Absolute Error}  =  \frac{1}{N_s} {\sum_{i=1}^{N_s}  E_i; \text{where} \   E_i=  \sum_{k=1}^{N_i} \frac{|u_i^k- u_i^{*k}|}{N_i}},
\end{equation}
where $u^*,\;u $ are the reference and the predicted solution respectively. $N_i$ is number of data points for $i^{th}$ sample, $N_s$ is the total number of samples. 
We also define, the error distribution by $\mu \pm \sigma$,  where $\mu = mean\{E_i\}, \sigma= std.dev\{E_i\}$ are mean and standard deviation of errors over the samples. \D{Unless otherwise noted, the mean test error refers to the mean absolute error (MAE) on the test set.
}

\subsection{System Parameter Identification using DeepONet}
\label{subsec:parameter_identification}

\subsubsection{\D{Deterministic Parameter Estimation}}
\begin{figure}[!htb]
    \centering
    \includegraphics[width=1\linewidth]{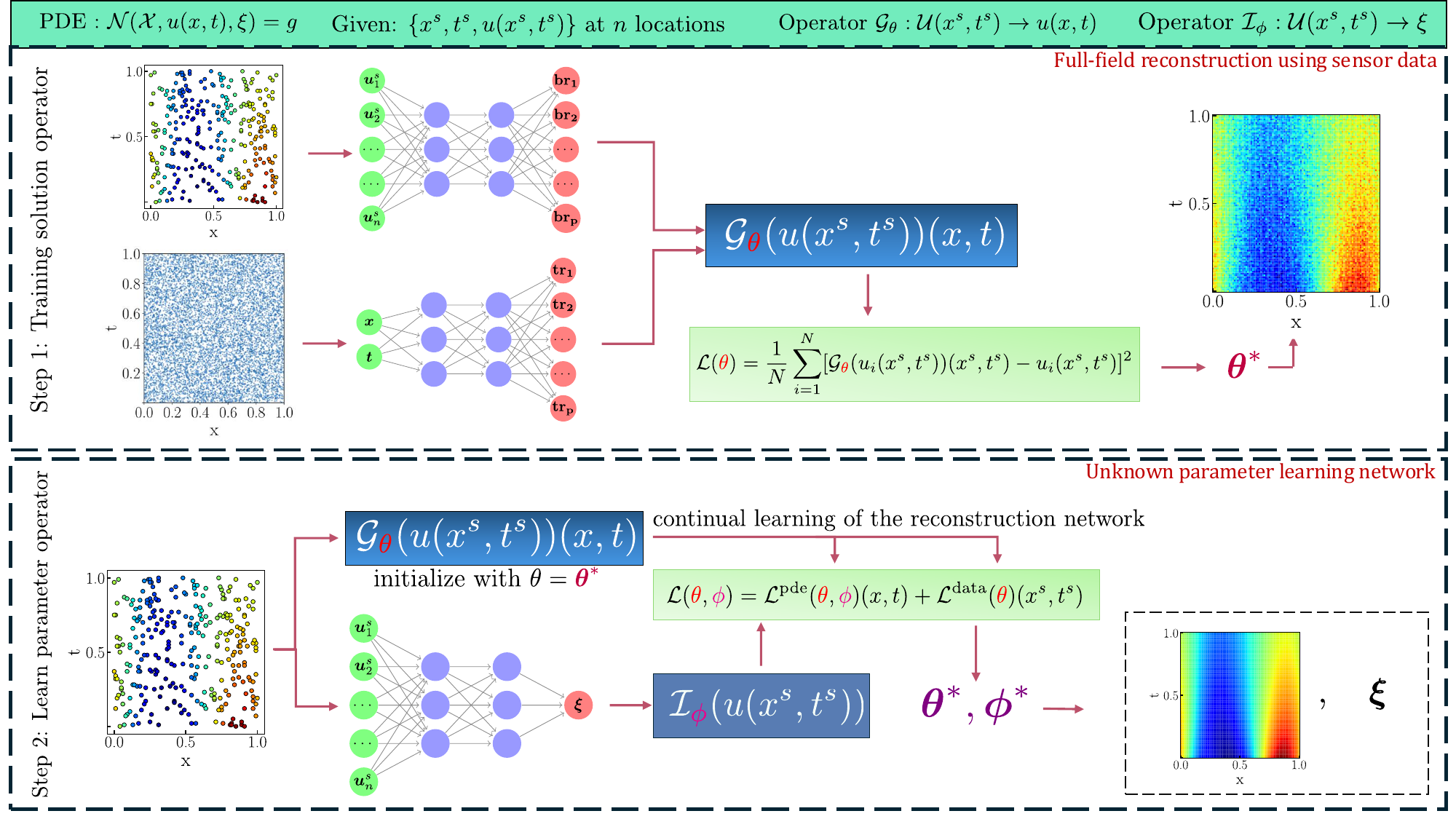}
    \caption{Schematic representation of the proposed architecture for unknown system parameters identification, $\nu$ for the Burgers' equation. In the schematic, $u^s$ denotes the known velocity field at fixed sensor locations. In step 1, a data-driven DeepONet is trained with loss defined at just the sensor locations to reconstruct the field based on sparse measurements. In the second step, the trained DeepONet is given a warm start, and the learning of the solution is improved in conjunction with an MLP which predicts the unknown $\nu$ employing a physics loss. $\bm\theta^*, \bm\phi^*$ are the optimized parameters of the networks.}
    \label{DON_inv_arch}
\end{figure}

Building upon our proposed architecture, we present a modified framework for parameter estimation in governing PDEs through surrogate modeling. While the underlying physics are known in this scenario, our objective is to determine the system parameters that govern the process (i.e., solving the parameter estimation inverse problem). Figure \ref{DON_inv_arch} illustrates the schematic of this modified architecture. For this framework, we have considered cases with just sensor recordings and no additional labeled datasets.

The framework consists of two key components. First, a DeepONet is pre-trained to learn the forward mapping from sparse input data to the complete solution field. Second, we introduce an additional MLP designed to learn the inverse mapping from the solution space to the parameter space. These two networks - the DeepONet and MLP - are then trained through continual learning, using the PDE residual as the primary loss term. To ensure solution accuracy and physical consistency, we incorporate additional loss terms for known data.

The fundamental advantage of this framework lies in its ability to simultaneously learn both forward and inverse mappings for any given physical system, without any additional labeled data beyond the sensor measurements. This dual-surrogate approach enables efficient parameter estimation while maintaining physical consistency through the embedded governing equations.

\subsubsection{ \D{Probabilistic Parameter Estimation}}

\D{The deterministic framework provides a single point estimate of the unknown parameter $\xi$. However, inverse problems often suffer from non-uniqueness where multiple parameter values can produce similar observations given sparse and noisy sensor data. To capture this inherent ambiguity, we extend our framework to estimate the full probability distribution of parameters. The MLP now outputs both the mean $\bm{\mu}_\xi$ and standard deviation $\bm{\sigma}_\xi$ of a Gaussian distribution: $\xi \sim \mathcal{N}(\bm{\mu}_\xi, \text{diag}(\bm{\sigma}_\xi^2))$, where $[\bm{\mu}_\xi, \bm{\sigma}_\xi] = \mathcal{I}_\phi(\mathbf{u}^s)$. During training, we employ the reparameterization trick ($\xi = \bm{\mu}_\xi + \bm{\sigma}_\xi \odot \bm{\epsilon}$ with $\bm{\epsilon} \sim \mathcal{N}(0, \mathbf{I})$) to enable backpropagation through the stochastic sampling process. The total loss function includes an additional KL divergence regularization term: $\mathcal{L}_{\text{total}} = \mathcal{L}_{\text{pde}}(\bm{\theta}, \bm{\phi}) + \lambda_{\text{data}} \mathcal{L}_{\text{data}}(\bm{\theta}) + \lambda_{\text{KL}} D_{\text{KL}}\left( \mathcal{N}(\bm{\mu}_\xi, \text{diag}(\bm{\sigma}_\xi^2)) \,\|\, p_{\text{prior}}(\xi) \right)$, where the KL term constrains the learned distribution to remain within physically meaningful bounds based on a prior $p_{\text{prior}}(\xi)$ informed by domain knowledge.}

\D{The probabilistic framework offers distinct advantages, it naturally quantifies epistemic uncertainty (narrow distributions indicate high confidence, broad distributions signal ambiguity), explicitly captures parameter non-uniqueness rather than forcing potentially misleading point estimates, accommodates measurement noise through learned variance that increases with data corruption, and enables risk-aware decision making through access to the full parameter distribution for uncertainty propagation. The training procedure follows Algorithm~\ref{training_algorithm_SPI} with minor modifications: the MLP output dimension is doubled to predict both $\bm{\mu}_\xi$ and $\bm{\sigma}_\xi$, parameter sampling via reparameterization is performed before physics loss evaluation, and the KL divergence term is added to the total loss. In Section~\ref{sec:examples}, we demonstrate both deterministic and probabilistic parameter identification, showing that the probabilistic framework provides calibrated uncertainty estimates that accurately reflect parameter ambiguity in sparse observation scenarios.}

\begin{algorithm}[!htb]
\caption{Training Scheme to obtain an operator framework for system parameter identification.}
\label{training_algorithm_SPI}
\SetAlgoLined
\D{\textbf{Inputs and Training Data:} sensor measurements $u(x^s,t^s)$ at $n$ fixed locations, collocation points $(x,t)$, PDE $\mathcal{N}(\mathcal{X},u(x,t),\xi)$, and training pairs $(u_i(x^s,t^s),\xi_i)_{i=1}^{N_{\text{train}}}$ consisting of $N_{\text{train}}$ system realizations with corresponding true parameters.} \\
\textbf{Step 1:} Training solution operator $\mathcal{G}_{\bm{\theta}}$\\
Initialize DeepONet $\mathcal{G}_{\bm{\theta}}$ with: \\
\hspace{1cm}Branch network: process sensor inputs $u(x^s,t^s)$\\
\hspace{1cm}Trunk network: process \D{coordinates} $(x, t)$ \\
\While{not converged}{
    \For{\D{for each batch of K system realizations}}{
        {\it{Forward pass through DeepONet at sensor locations}}\\
        prediction = $\mathcal{G}_{\bm{\theta}}(u(x^s, t^s)(x^s,t^s))$\\
        \D{{\it{Compute loss}} \\
        $\mathcal{L}(\theta) = \frac{1}{K} \sum_{i=1}^{K} \left\lVert \text{prediction}_{i} - u_{i}(x^{s}, t^{s}) \right\rVert^{2}$}\\

        Backpropagate and update $\theta$ using Adam
    }
}
{\it{Store optimised $\bm\theta$}}\\
\hspace{0.5cm} $\bm\theta^*= \text{ argmin} \mathcal{L}(\bm\theta)$\\
\textbf{Step 2:} Learn parameter operator $\mathcal{I}_{\bm\phi}$\\
Initialize DeepONet $\mathcal{G}_{\bm{\theta}}$ with $\bm\theta = \bm\theta^*$ from step 1. \\
Initialize Parameter network $\mathcal{I}_\phi$: maps sensor data $\{ u(x^s,t^s)\}$ to unknown parameter $\xi$ \\
\While{not converged}{
    \For{each batch of data}{
        {\it{Forward pass through both networks}} \\
        $u_{\text{pred}}$ = $\mathcal{G}_{\bm{\theta}}(u(x^s, t^s)(x,t))$ \\
        $\xi_{\text{pred}} = \mathcal{I}_{\bm\phi}(u(x^s, t^s))$ \\
        {\it{Compute combine loss}} \\
        $\mathcal{L}^{\text{pde}} = \frac{1}{N} \sum_{i=1}^N || \mathcal{N}(\mathcal{X}, u_{\text{pred}_i}, \xi_{\text{pred}_i}) - g_i ||^2$ ~~{\it{physics loss}}\\
        $\mathcal{L}^{\text{data}} = \frac{1}{N} \sum_{i=1}^N || u_{\text{pred}_i} - u(x^s,t^s)||^2$ ~~{\it{data loss}}\\
        $\mathcal{L}(\bm\theta,\bm\phi) = \mathcal{L}^{\text{pde}}(\bm\theta,\bm\phi)(x,t) +\mathcal{L}^{\text{data}}(\bm\theta)(x^s,t^s)$ \\
        Back-propagate and update $\bm\theta$ and $\bm\phi$ using Adam
    }
}

{\it{Final optimized $\bm\theta$, $\bm\phi$}}\\
$\bm\theta^*$, $\bm\phi^*$ = argmin $\mathcal{L}(\bm\theta, \bm\phi)$
\end{algorithm}

 \D{Algorithm \ref{training_algorithm_SPI} presents our two-stage training procedure for learning the parameter identification operator. The algorithm operates on a training dataset comprising multiple system realizations, where each realization corresponds to distinct parameter values and yields sensor measurements at fixed spatial locations over time. During training, the networks process mini-batches of these realizations, optimizing the loss functions through standard stochastic gradient descent until convergence.}

\section{Illustrative Problems}
\label{sec:examples}

To examine our approach, we perform tests on four \D{sets} of examples including the Reaction Diffusion equation, Burgers' equation, Heat equation and Helmholtz Equation. \D{ The network architectures were systematically selected following established DeepONet guidelines \cite{lu2021learning}, with branch and trunk networks employing 3-4 hidden layers of 64-128 neurons. Hyperparameter sweep analysis demonstrates robust performance across reasonable architectural variations. Complete details are provided in Appendix~A (Table~A.1). For each problem, we demonstrate the applicability of our frameworks for hidden physics discovery, parameter identification, or both depending on the problem 
configuration.}

\subsection{Reaction Diffusion System}
\label{subsec:reaction_diffusion}

We first examine our approaches using a reaction-diffusion equation, which models various physical phenomena in nature such as oscillating chemical concentrations and spatial waves as seen in the Belousov-Zhabotinsky reaction, population dynamics, pattern formation in biological systems and morphology in crystals and alloys~\cite{Kop04}. The following equation describes the spatiotemporal evolution of the system and is given by:
\begin{equation}\label{reaction_diffusion_eqn_IBC}
    \begin{split}
    &\frac{\partial u}{\partial t} (x,t)=  D \frac{\partial^2 u}{\partial x^2} (x,t)+ K u^2(x,t) + f(x) \text{ on }  \ \Omega: (x,t) \in [0,1]^2 , \\
    & IC: u(x,0)= 0 ,\\
    & BC: u(0,t)= u(1,t) =  0 ,\\
  \end{split}
\end{equation}
where $D$ and $K$ denote the diffusion and the reaction coefficients, respectively. In \autoref{reaction_diffusion_eqn_IBC}, \textit{IC} denotes the initial condition and \textit{BC} denotes the Dirichlet boundary condition. Going forward, we will first present the results of discovering the unknown physics using DeepONet, followed by presenting the results of employing DeepONet to characterize the diffusion coefficient of the system using sparse measurements.

\vspace{6pt}
\noindent \textbf{Discovering physics using DeepONet}
We re-write \autoref{reaction_diffusion_eqn_IBC} to denote the known and the unknown part of the PDE, as
\begin{equation}\label{reaction_diffusion_eqn_hidden}
\frac{\partial u}{\partial t} =  \mathcal{N}(u,u_x,u_{xx}) + f(x), 
\end{equation}
we assume that we know the time derivative of the problem and the remaining part of the governing equation shown in \autoref{reaction_diffusion_eqn_IBC} is unknown as well as approximated by $\mathcal{N}$, employing the proposed deep hidden physics framework within DeepONet. Our objective is to learn a solution operator $\mathcal{G}_{\boldsymbol \theta}$ that maps the source term $f(x)$ to the solution field $u(x,t)$ using partially known physics and sparse labeled data. For this setup, we have used $D = 0.01$ $K = 0.01$. To generate the training and testing data, the system in \autoref{reaction_diffusion_eqn_IBC} was solved using the finite difference method, with a spatial resolution of $\delta x = 0.01$ units and a time step of $\delta t = 0.001$ units. The final solution was evaluated on a $101 \times 101$ grid in the space-time domain. We investigate three different input function spaces for source term $f(x)$. We train the model only using sine basis functions and test for all three function spaces.
\begin{itemize}[leftmargin=*]
    \item \textbf{Sine Basis Functions: }
          \begin{equation}\label{source_terms_sine_basis}
           f(x) = \sum_{k=1}^{N_f} A_k sin(\pi k x),
          \end{equation}
          where $A_k$ are coefficients from a Gaussian normal distribution with mean 0 and variance 1 and the number of frequencies $N_f = 5$.
    \item \textbf{Gaussian Random Fields (GRFs): }
        \begin{equation}\label{grf_eqn}
        GRF(x) \sim { \mathcal{G}(0,k_l(x_1,x_2))},
        \end{equation}
        with RBF kernel $k_l(x_1,x_2) = exp(-|| x_1 - x_2||^2/(2l^2))$ with a length-scale parameter $l > 0$. The $l$ determines the oscillatory nature of the sampled function, a lower value of $l$ leads to higher oscillations. We consider GRFs with RBF kernel length-scale of 0.2,  unlike sine basis functions GRFs are not zero at boundaries. 
    \item \textbf{Modified GRFs: }
        \begin{equation}\label{modified_grf_eqn}
        \textnormal{Modified} \ GRF(x) =  \alpha(x-x^2)GRF(x)
        \end{equation}
        where $\alpha = 8.0$ ensures similar scale as sine basis functions.
\end{itemize}

To train the framework, the collocation points in the differential equation, initial conditions, and boundary conditions are defined as $N_{Coll}$, $N_{ic}$, and $N_{bc}$, respectively. These points are randomly generated for each sample using Latin hypercube sampling \cite{LATINHYCUBE}, creating synthetic domain data that can vary at each gradient descent step. The training process is shown in Algorithm \ref{training_algorithm_hidden_physics}. For this problem, we use $N_{Coll} = 2000$, $N_{ic} = 200$, and $N_{bc} = 250$. To determine the optimal amount of known labeled dataset required, we experiment with data point sets $N_d = \{200, 500\}$ and training sample sets $N_{train} = \{50, 100, 200, 500\}$, with a fixed number of test samples, $N_{test} = 1000$. 

The branch net takes the source term discretized at 101 spatial sensor locations, while the trunk net takes the spatial and temporal locations while the hidden physics MLP acts as a surrogate model for the right-hand side (RHS) of \autoref{reaction_diffusion_eqn_hidden}. It accomplishes this by taking the solution $u$ and its spatial gradients, $[u, u_x,u_{xx}]$ as inputs and producing the prediction $\mathcal{N}$.
The optimal model was found using training size of $N_{train} = 500, N_d = 500$. The loss curve over the training process is shown in Figure~\ref{fig:loss_plot}(a). The  test mean absolute error for each of the cases is presented in Table~\ref{tab:test_error_different basis}, The solution is predicted within training time domain unless specifically mentioned.

\begin{table}[h!]
    \centering
    \caption{Reaction Diffusion system: Mean and standard deviation of the test error for all the three different input function spaces for the source term, $f(x)$. The experiments were carried out for five independent runs.}
    \begin{tabular}{l c}
        \hline
        \textbf{Method} & \textbf{Test Error (Mean $\pm$ Std)} \\
        \hline
        Sine basis & $0.00871 \pm 0.00424$ \\
        GRF & $0.03918 \pm  0.02095$ \\
        Modified GRF & $0.01653 \pm 0.00856$ \\
        \hline
    \end{tabular}\label{tab:test_error_different basis}
\end{table}

\begin{figure}[!htb]
\centering
\begin{subfigure}{0.45\textwidth}
    \centering
    \includegraphics[width=\linewidth]{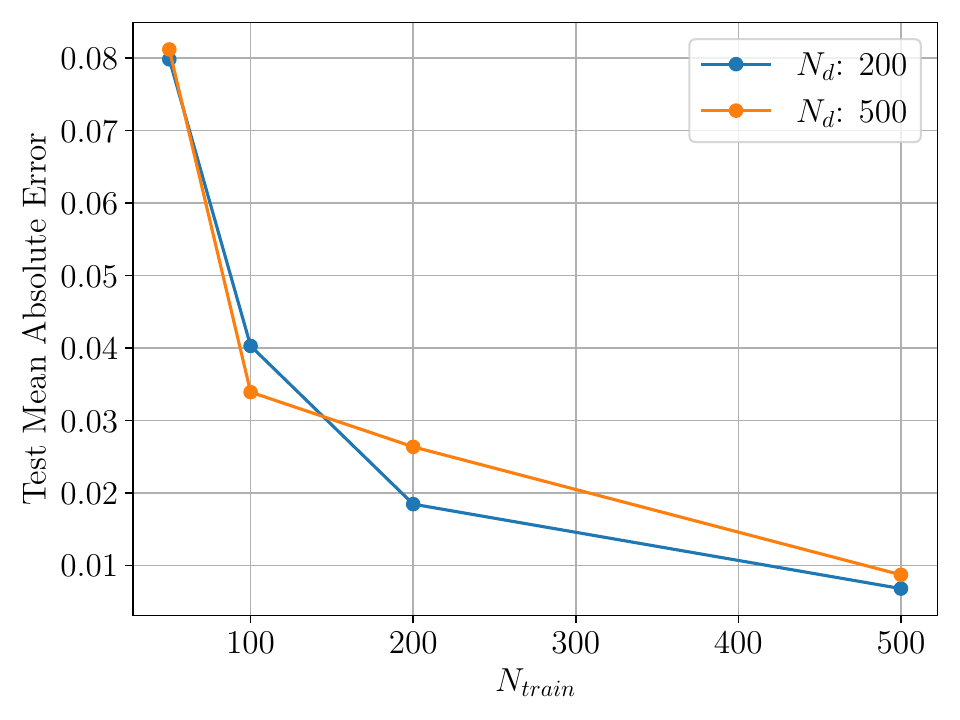}  
\end{subfigure}
\begin{subfigure}{0.45\textwidth}
    \centering
    \includegraphics[width=\linewidth]{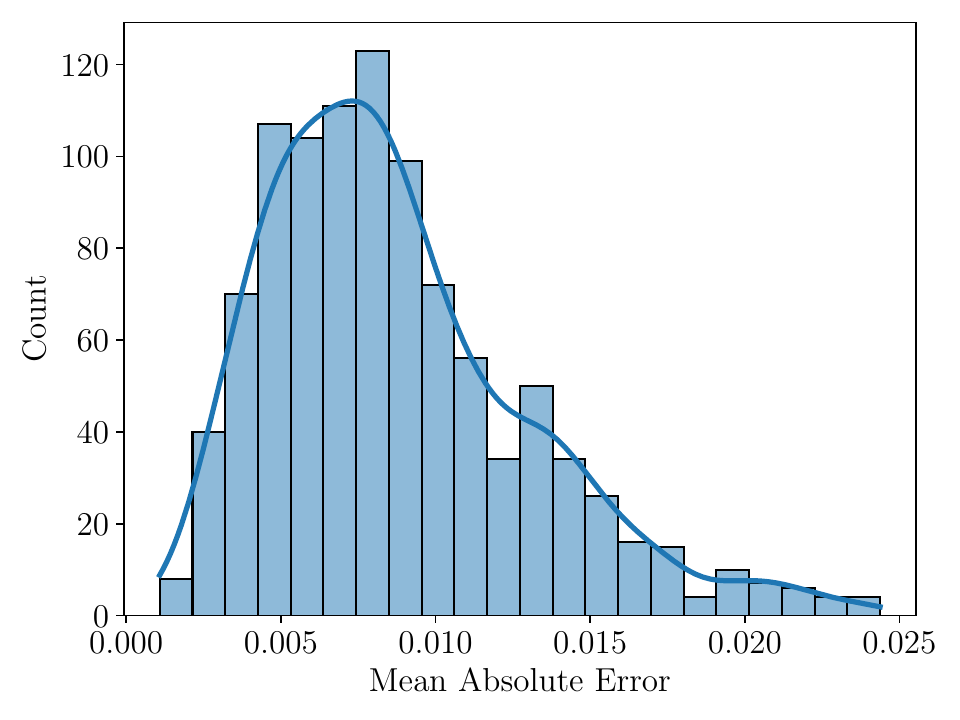}  
\end{subfigure}
\caption{Performance analysis for Reaction diffusion equation: (a) Mean test error over varying $N_{\text{train}}$. (b) Test error distribution of optimal model that achieved an error $0.00871 \pm 0.00424$ with $N_{train} = 500$, $N_{d} = 500$.}
\label{fig:error_reaction_diffusion}
\end{figure}

Using sine basis functions for the source term yields the highest prediction accuracy among tested methods. Figure~\ref{fig:error_reaction_diffusion}(a) illustrates the mean test error over various sample sizes and training data points, while Figure~\ref{fig:error_reaction_diffusion}(b) shows the distribution of test errors for this optimal model with sine basis input function. We further evaluate the model's predictive capability by using the learned hidden physics MLP $\mathcal{N}(x,t_n;f)$ to solve the governing PDE via time integration beyond the training data (temporal extrapolation). Starting from the initial condition $u_0$, we apply forward Euler integration with the discovered physics term. For the Reaction-Diffusion system, the update rule is $u_{n+1}(x) = u_n(x) + (\mathcal{N}(x,t_n;f) + f(x)) \cdot dt$, while for Burgers' equation it simplifies to $u_{n+1}(x) = u_n(x) + \mathcal{N}(x,t_n;f) \cdot dt$. We compare two solution approaches: the direct DHPO prediction $u_{\text{pred}}$ (querying the trained DeepONet at future times) versus the time-integrated solution $u_{\text{hpm}}$ (obtained by stepping forward using the learned physics). We assess both approaches on unseen sine basis source terms as defined in \autoref{source_terms_sine_basis}.

\begin{figure}[!htb]
\centering

\begin{subfigure}{0.495\textwidth}
    \centering
    \includegraphics[width=1.15\linewidth]{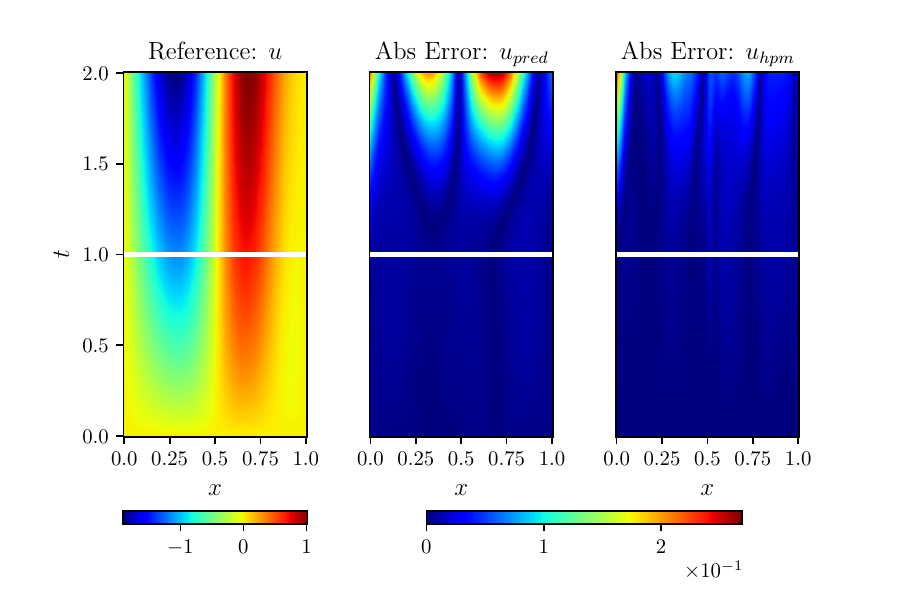}
    \caption{Sample 1}
\end{subfigure}
\hfill
\begin{subfigure}{0.495\textwidth}
    \centering
    \includegraphics[width=1.15\linewidth]{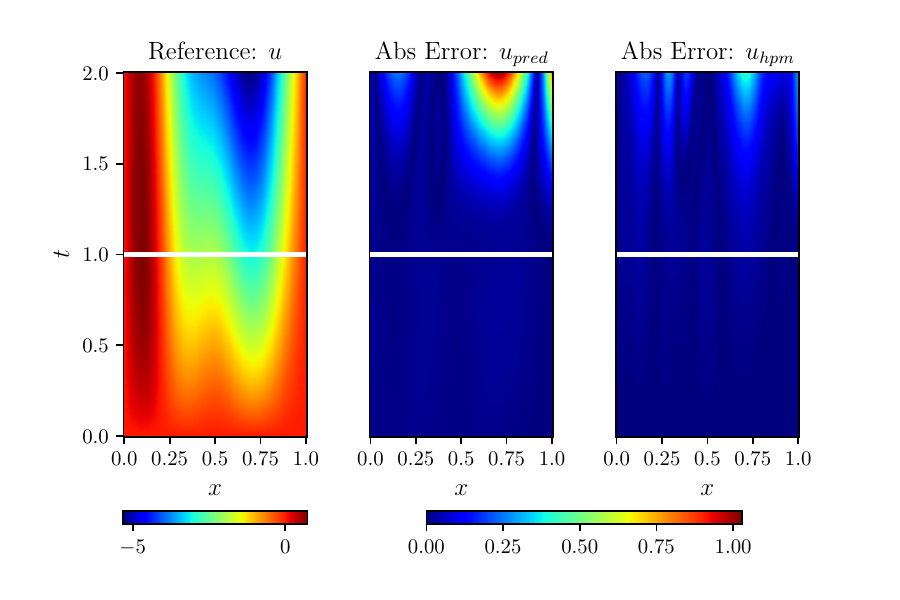}
    \caption{Sample 2}
\end{subfigure}

\caption{\D{Reaction-diffusion equation: reference vs.\ DHPO predictions and HPM-integrated trajectories for two representative test samples drawn from the sine-function input space.  
Training data spans $t\le1$, while the extrapolation region ($t>1$) is indicated by white horizontal lines.  
For Sample 1, mean absolute errors are  
$u_{pred}(t\le1)=0.00432$, $u_{pred}(t>1)=0.04360$,  
$u_{hpm}(t\le1)=0.00226$, $u_{hpm}(t>1)=0.01917$.  
For Sample 2, mean absolute errors are  
$u_{pred}(t\le1)=0.01299$, $u_{pred}(t>1)=0.12473$,  
$u_{hpm}(t\le1)=0.00655$, $u_{hpm}(t>1)=0.05840$.}}
\label{fig:RD_sine_extp_samples}
\end{figure}

As shown in Figure~\ref{fig:RD_sine_extp_samples}, the model accurately predicts  the DHPO solution ($u_{pred}$) and DHPO MLP integrated solution ($u_{hpm}$) at the training region. We also observe, in extrapolation region, the model accurately predicts the solution up to a specific time threshold, the accuracy in integrated solution is higher than DHPO solution, Hence we can infer that the Hidden Physics MLP is accurately capturing the dynamics. Figure~\ref{fig:RD_sine_hid_phy_samples} shows the contours of reference and predicted PDE terms. The inferred hidden physics term is given by,
$$\mathcal{N} \approx D \frac{\partial^2 u}{\partial x^2} + K u^2.$$

\begin{figure}[!!htb]
\centering
\begin{subfigure}{0.45\textwidth}
    \centering
    \includegraphics[width=\linewidth]{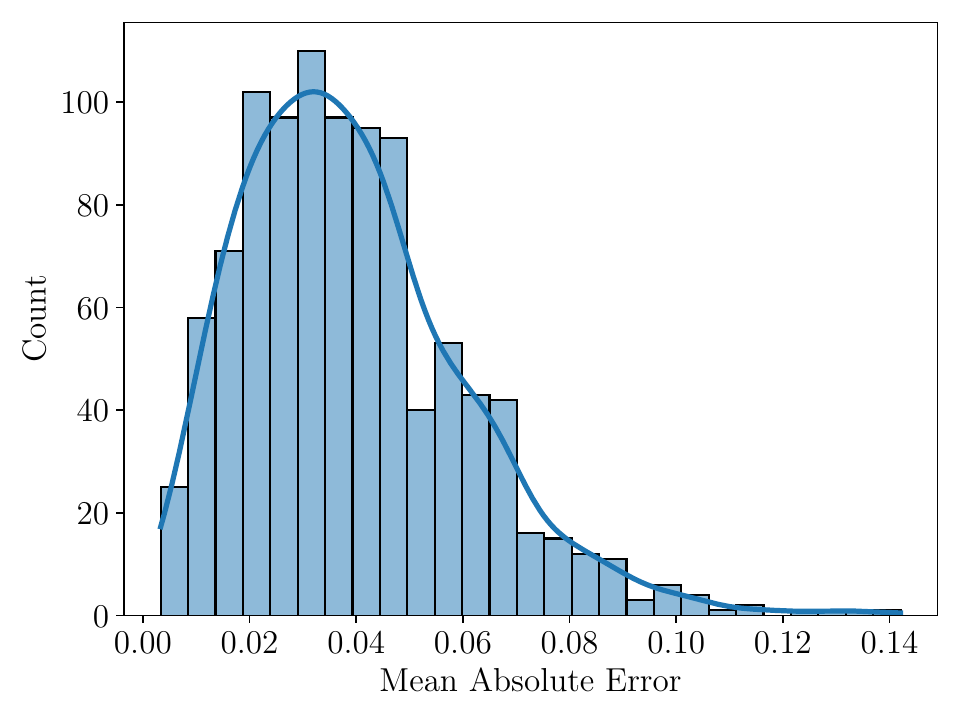}  
\end{subfigure}
\begin{subfigure}{0.45\textwidth}
    \centering
    \includegraphics[width=\linewidth]{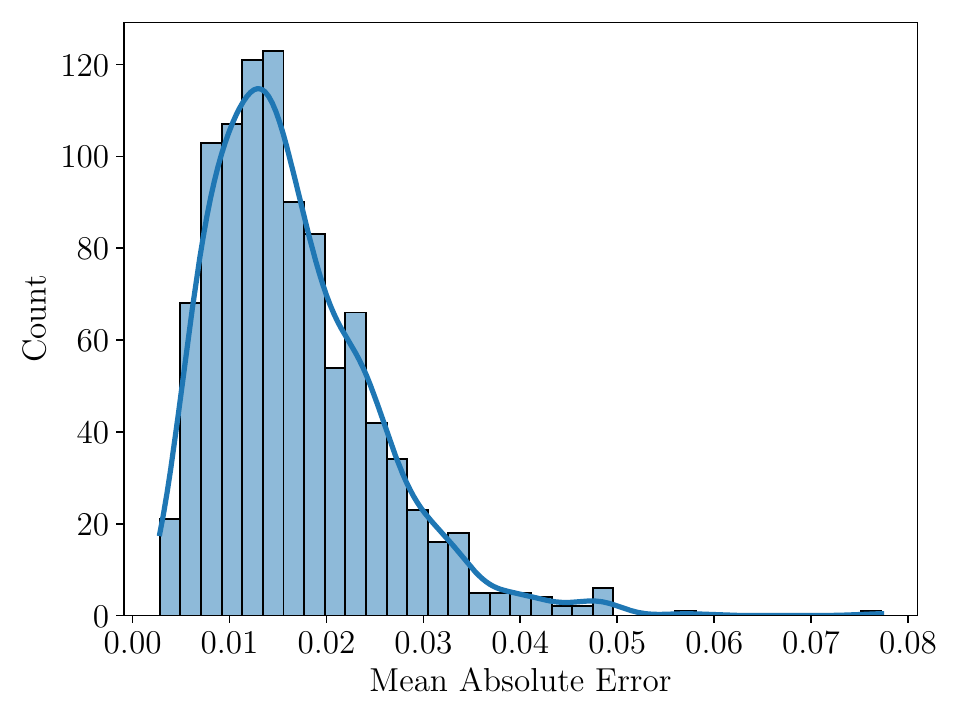} 
\end{subfigure}
\caption{Performance analysis for GRFs and Modified GRFs input function spaces: (a) GRFs basis $l=0.2$, Test error =  0.03918 $\pm$ 0.02095. (b) Modified GRFs basis $l=0.2$, Test error = 0.01653 $\pm$ 0.00856}
\label{fig:dist_grf_and_modified_grf}
\end{figure}

Unlike the sine basis, Gaussian random fields (GRFs) do not approach zero at the boundaries, leading to higher deviations in the predicted solution and physics terms, particularly at the boundaries (see Figure~\ref{fig:RD_grf_extp_samples}). Figure~\ref{fig:dist_grf_and_modified_grf}(a) illustrates the test error distribution for GRFs evaluated with the model trained with sine basis. In contrast, modified GRFs, which are zero at the boundaries similar to sine basis functions, achieve comparable prediction accuracy. Figure~\ref{fig:RD_modified_grf_extp_samples} shows that, with modified GRFs, the model accurately predicts both $u$ and the hidden physics terms, closely matching the ground truth. The test error distribution for modified GRFs is provided in Figure~\ref{fig:dist_grf_and_modified_grf}(b).

Additionally, we assess the model’s performance on modified GRFs by varying the length-scale parameter $l$ of the RBF kernel with values $l \in \{0.1, 0.15, 0.20, 0.40, 0.60\}$. As expected, Table~\ref{length_scale_error} demonstrates that prediction error increases as the length scale decreases. However, for out-of-distribution functions with boundary values approaching zero, the model maintains relatively higher accuracy compared to cases with non-zero boundary conditions.

\begin{itemize}[leftmargin=*]
\item \textbf{Noise robustness:}

Considering the impact of noise on model performance, an assessment was conducted using training data corrupted with synthetically generated random uniform noise. The noise was introduced as follows $u_{noise} = u_{clean} + U(0,\gamma \sigma_u )$ , where $U$ is sampled from uniform random distribution with minimum value 0, maximum $\gamma \sigma_u$ , $\sigma_u$ denotes standard deviation in clean data, the parameter $\gamma$ controls the noise intensity.
We train the model considering $\gamma \in \{0, 0.2, 0.5\}$, for direct comparison, both the training data volume and the training methodology were kept identical to those utilized for the optimal model presented earlier. The results are computed considering 1000 test inputs from sine basis function space. As illustrated in Table~\ref{noise_error}, the test error predictably increases with rising noise intensity. Notably, for noise intensity $\gamma < 0.2$, the error exhibits only a marginal increase, $\approx 10\%$ . This indicates that the model maintains robustness in capturing the underlying dynamics within this noise threshold.

\item \textbf{Sparsity:}

To evaluate the model's performance under conditions of data sparsity, training was conducted using only two extreme temporal data points $(t = 0 \  \text{and } t=1)$ along with 101 spatial data points. We train the model with 500 samples using the similar training procedure as shown in the previously trained optimal model. Evaluating the model on 1000 test samples yielded a mean test error of $0.00681  \pm 0.00337$. When compared against the optimal model which was trained on 500 spatio-temporal data points, these results demonstrate that the sparse data model is remarkably capable of accurately predicting the dynamics, despite its limited training data.

To assess the benefits of physics-informed training, we compare against a standard DeepONet trained on the same sparse data using identical branch and trunk network architectures. Evaluating on 1000 test samples, the DeepONet achieves a prediction error of $0.02619\pm 0.01074$ at the training time instants but degrades to $0.43818 \pm 0.21707$ at interior (unobserved) time instants. This indicates severe overfitting to the training timestamps with poor generalization to the interior time domain. Representative predictions are shown in Figure~\ref{fig:deeponet_soln}.

\end{itemize}

\begin{table}[!htb]
\caption{Reaction Diffusion system: Mean and standard deviation of the test error for all the case with input functions drawn from modified GRFs with varying length-scale parameter, $l$. The experiments were carried out for five independent runs.}
\centering 
\begin{tabular}{c c}
\hline\hline 
$l$ & Test Error \\ [0.5ex] 
\hline 
0.10 & 0.07945 $\pm$ 0.03666\\
0.15 & 0.03496 $\pm$ 0.01831\\
0.20 & 0.01653 $\pm$ 0.00856\\
0.40 & 0.00755 $\pm$ 0.00421\\
0.60 & 0.00656 $\pm$ 0.00378\\ [1ex] 
\hline 
\end{tabular}
\label{length_scale_error} 
\end{table}

\begin{table}[!htb]
\caption{Reaction Diffusion system: Mean and standard deviation of the test error over the noise level.}
\centering 
\begin{tabular}{c c}
\hline\hline 
noise intensity $\gamma$ & Test Error \\ [0.5ex] 
\hline 
0.0 & 0.00871 $\pm$ 0.00424\\
0.2 & 0.00983 $\pm$ 0.00515\\
0.5 & 0.01652 $\pm$ 0.00895\\ [1ex] 
\hline 
\end{tabular}
\label{noise_error} 
\end{table}

\noindent\textbf{System parameter identification: Diffusion coefficient}

\noindent \D{\textbf{Deterministic.}} The neural operator framework introduced in Section~\ref{subsec:parameter_identification} is employed to characterize the system, specifically to identify the diffusion coefficient of the governing PDE using solution values at selected spatial and temporal locations.

For this investigation, we considered varying diffusion coefficients and source terms in the Reaction-Diffusion equation defined in Equation~\ref{reaction_diffusion_eqn_IBC}. The equation was solved numerically using a finite difference method on a uniform grid with spatial resolution $\delta x = 0.01$ and time step $\delta t = 0.001$, creating a $101 \times 101$ grid over the $x$-$t$ domain. The solution dataset was generated using 500 different diffusion coefficient values ($D$) uniformly distributed between $0.01$ and $0.05$. Each diffusion coefficient was paired with 20 distinct source terms $f(x)$ sampled from a Gaussian Random Field (GRF), resulting in 10,000 solution fields. The dataset was split into $N_{\text{train}} = 8,500$ samples for training and $N_{\text{test}} = 1,500$ samples for testing. To emulate sparse measurements, 300 spatial-temporal locations were randomly sampled and fixed across all training and testing samples, providing known values of $u(x,t)$.
The network training hyperparameters are presented in Table \ref{tab:hyperparameter}. The training configuration employs $N_{\text{coll}} = 2500$ collocation points for PDE residual evaluation. In the loss function, a penalty coefficient of $ \lambda_{\text{coll}} = 10$ was applied to the PDE residual term, while unity weights were maintained for data loss terms.

\begin{figure}[!h]
\centering
\begin{subfigure}{0.9\textwidth}
    \centering
    \includegraphics[width=0.9\linewidth]{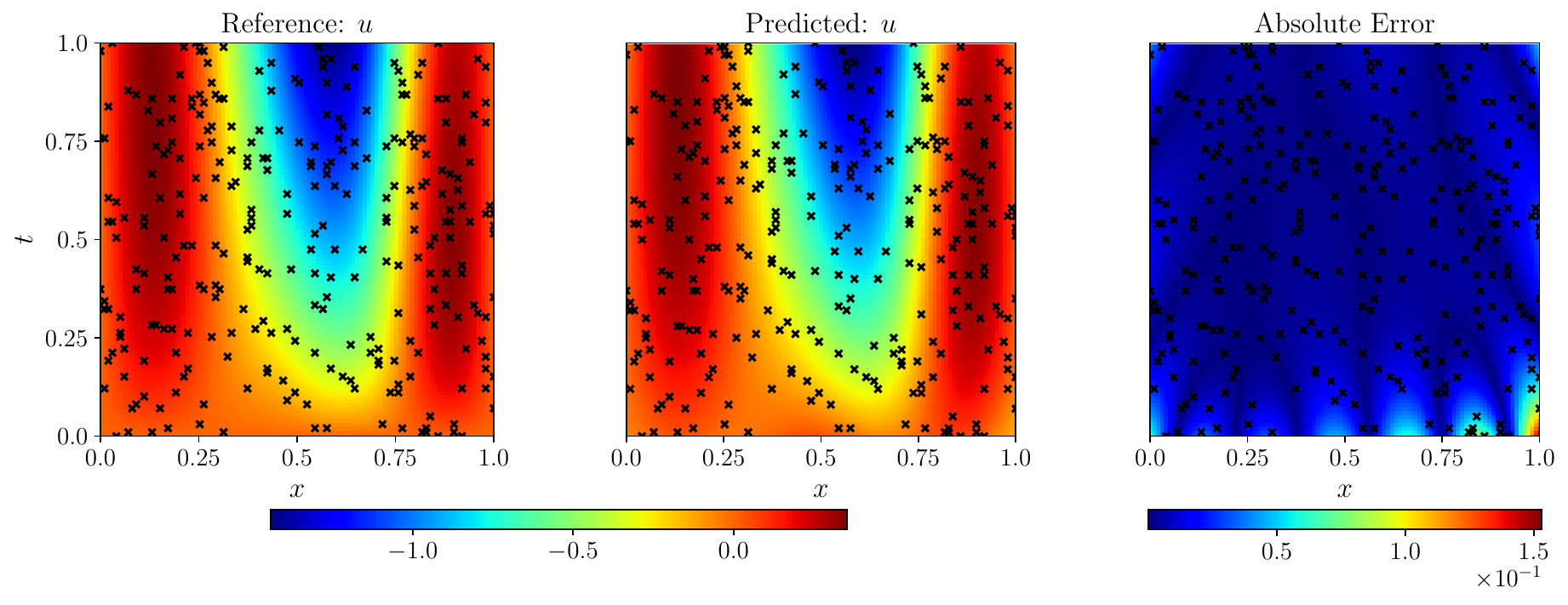}  
    \caption{Sample 1: Relative $L_2$ error of $u(x,t)$ = $0.027$. For this case, $D_{\text{true}} = 0.027$ while $D_{\text{predicted}} = 0.027$.}
    \label{react_diff_inv_sample_200}
\end{subfigure}
\begin{subfigure}{0.9\textwidth}
    \centering
    \includegraphics[width=0.9\linewidth]{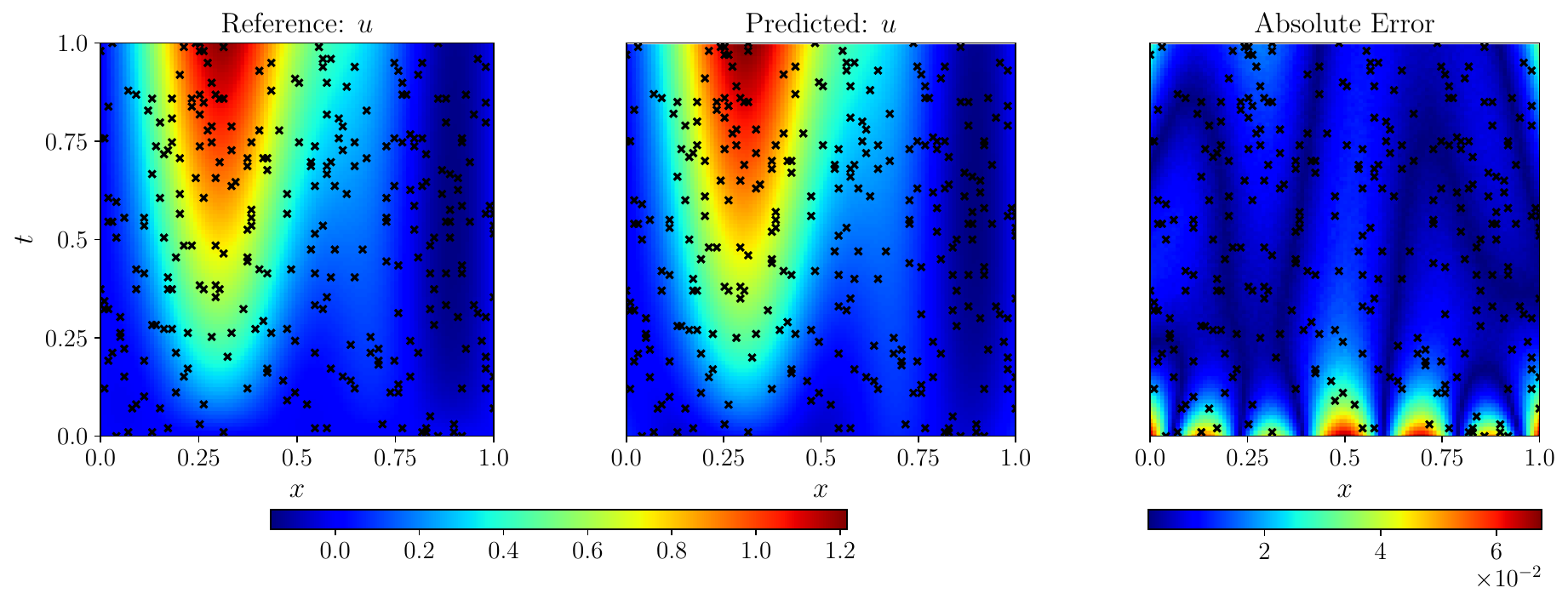}  
    \caption{Sample 2: Relative $L_2$ error of $u(x,t)$ = $0.029$. For this case, $D_{\text{true}} = 0.036$ while $D_{\text{predicted}} = 0.034$.}
    \label{Burgers_inv_sample_2000}
\end{subfigure}
\caption{Reaction Diffusion Equation: Comparison of the reference solution and the predicted solution. The black cross marks represent the location where the data of $u(x,t)$ was known during the testing. The error at these specific locations is also shown in the absolute error plot.}
\label{react_diff_inv_sample_200_n_2000}
\end{figure}

\begin{figure}[!htb]
    \centering
    \includegraphics[width=0.85\linewidth]{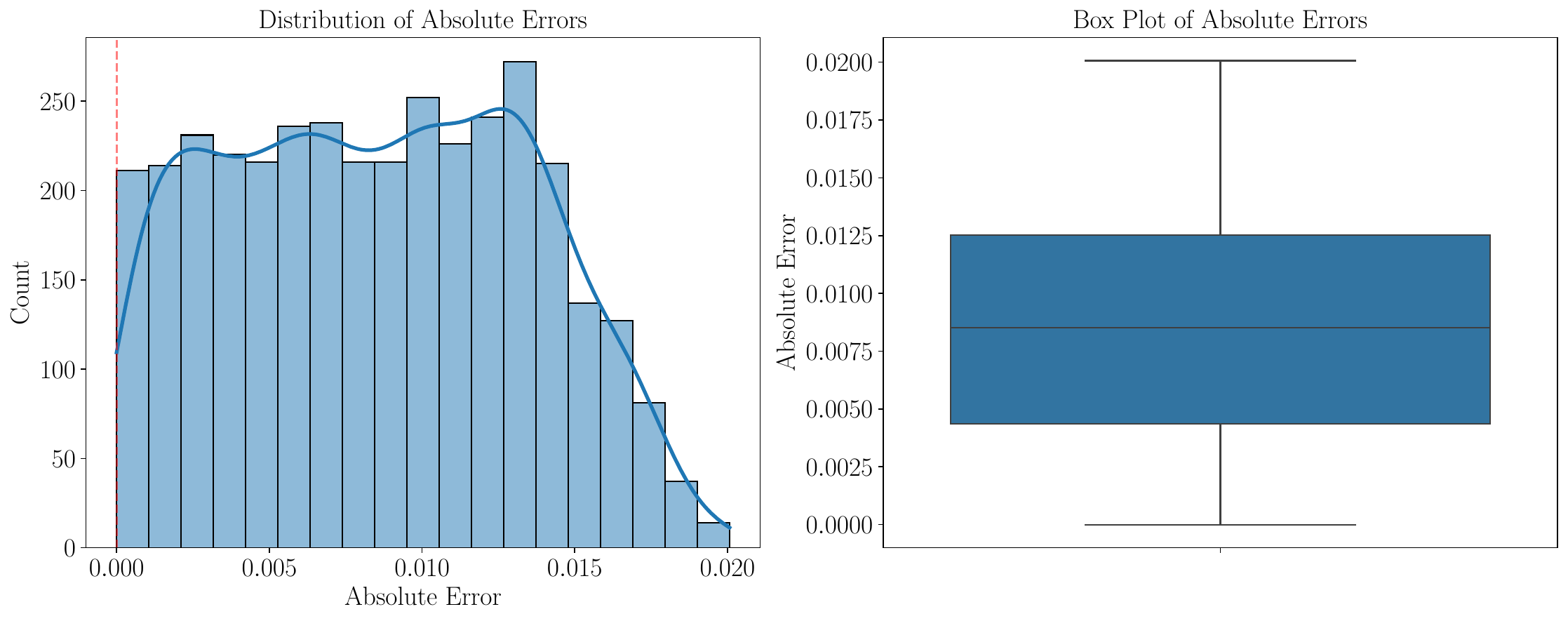}
    \caption{Reaction Diffusion Equation: Distribution of absolute error of $D$ when compared to the $D$ used to generate the ground truth using the finite difference solver. The framework approximates $D$ with an error of $8\times 10^{-3}\pm 0.008$.}
    \label{react_diff_inv_error_dist}
\end{figure}

Figure \ref{react_diff_inv_sample_200_n_2000} illustrates the framework's capability in reconstructing the solution field $u(x,t)$ and estimating the diffusion coefficient $D$ from 300 sparse measurements. Results from two representative test cases demonstrate solution reconstruction with $L_2$ errors of order $\mathcal{O}(10^{-2})$, along with accurate recovery of the true diffusion coefficient values. Figure \ref{react_diff_inv_error_dist} presents the distribution of absolute errors in predicted diffusion coefficients across the entire test set, showing that the majority of predictions maintain absolute errors below $0.005$. This indicates the framework's robust performance in parameter estimation tasks.

It is worth noting that determining the reaction coefficient from this system poses significant challenges due to the nonlinear relationship between $K$ and $u(x,t)$ in the reaction-diffusion equation. Small perturbations in the solution field $u(x,t)$ can lead to substantial errors in the estimation of $K$. Furthermore, the reaction term becomes negligible in regions where $u$ approaches zero, making the inverse problem for $K$ particularly challenging in these regions.

\begin{figure}[!htb]
    \centering
    \includegraphics[width=0.8\linewidth]{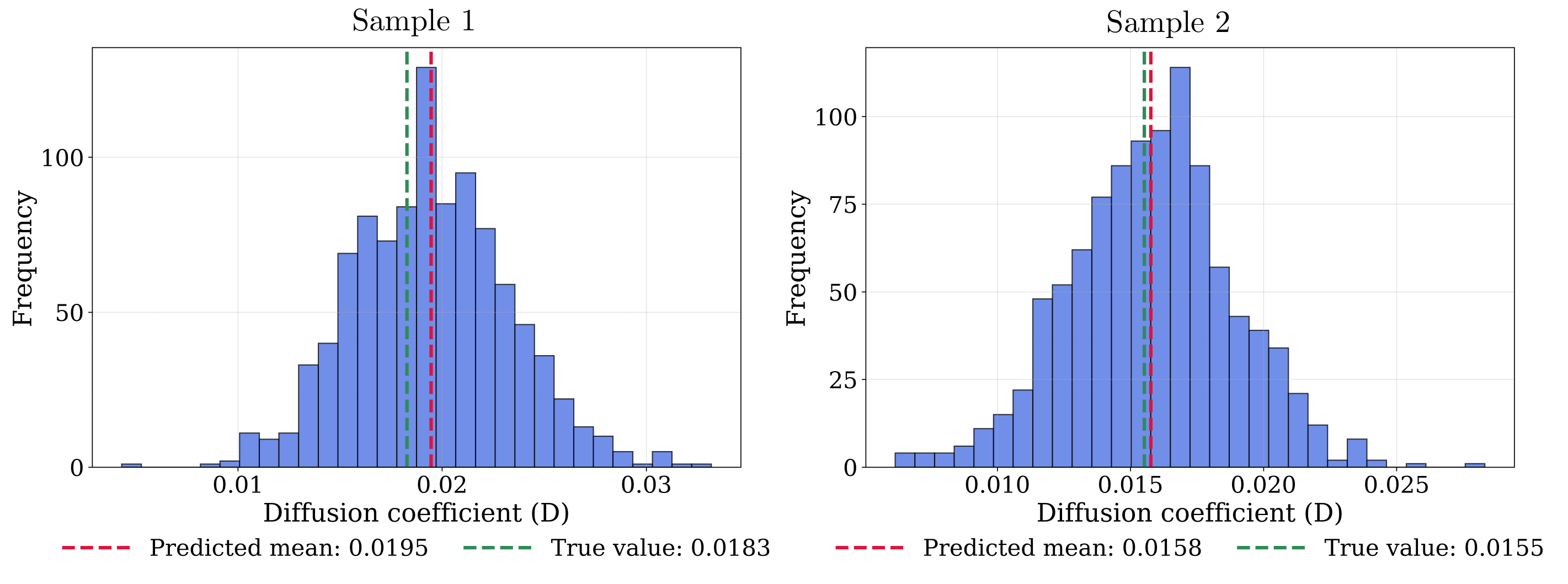}
    \caption{\D{Reaction Diffusion Equation: Predicted probability distributions of Diffusion coefficient $D$ for two representative test samples. The histogram represents 500 sampled values from the learned distribution $\mathcal{N}(\mu_D, \sigma_D^2)$. Dashed red line indicates the predicted mean, dashed green line shows the true Diffusion coefficient.}}
    \label{RD_inv_D_dist}
\end{figure}

\begin{figure}[!htb]
    \centering
    \includegraphics[width=\linewidth]{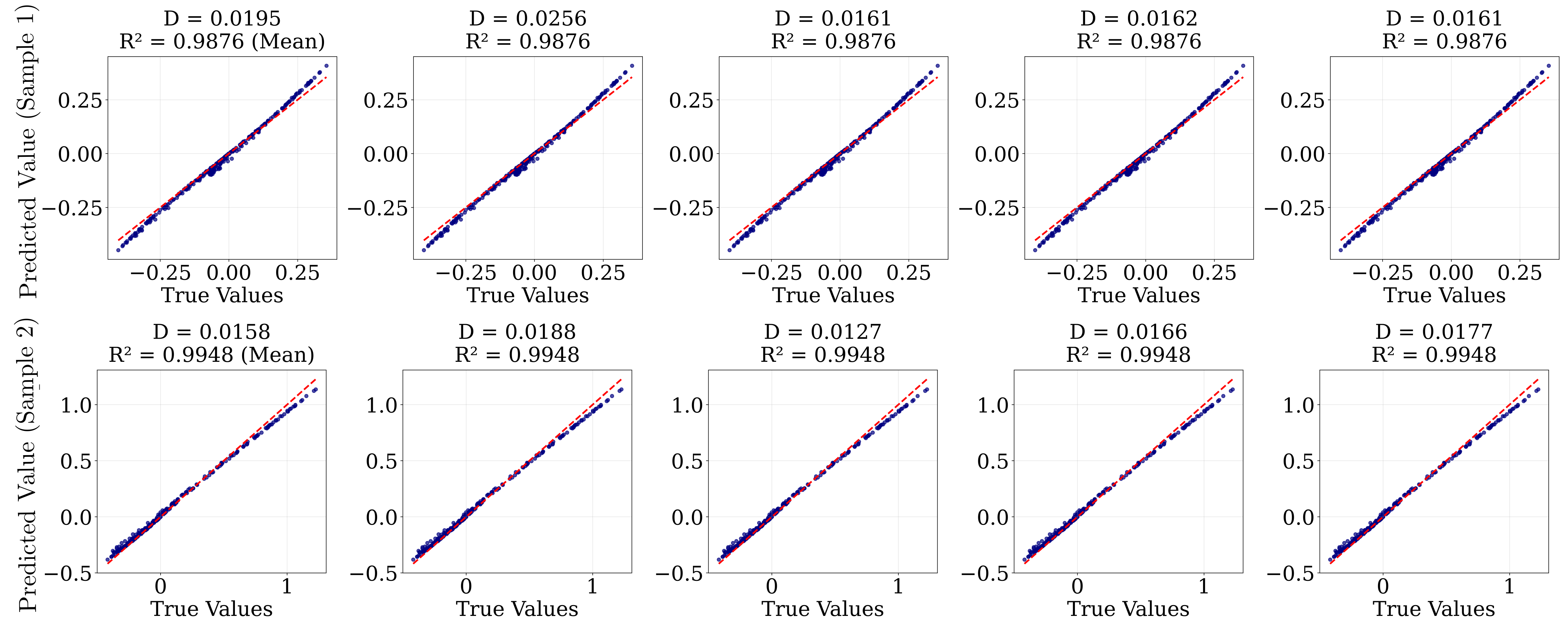}
    \caption{\D{Reaction Diffusion Equation: Validation of predicted parameter distribution for representative samples through forward finite difference simulations. Each subplot shows the true solution field $u(x,t)$ (x-axis) versus the solution obtained using a Diffusion coefficient value sampled from the predicted distribution (y-axis).}}
    \label{RD_inv_pred_u_sampled_nu}
\end{figure}

\noindent \D{\textbf{Probabilistic.} To validate the probabilistic framework's ability to capture parameter non-uniqueness, we evaluate whether parameter values sampled from the predicted distribution $\mathcal{N}(\mu_D, \sigma_D^2)$ can accurately reproduce the original sensor observations used as input to the inverse operator. For each test case, we sample 500 diffusion coefficient values from the learned distribution and perform forward simulations using a finite difference solver to obtain the corresponding solution fields. Figure~\ref{RD_inv_D_dist} illustrates the predicted parameter distributions for two representative test samples, showing the mean estimate (dashed red line) alongside the true ground truth value (dashed green line). The distribution widths reflect the epistemic uncertainty inherent in the inverse problem given the available sparse sensor data. Figure~\ref{RD_inv_pred_u_sampled_nu} presents a scatter comparison between the true solution field $u$ (used to generate sensor measurements) and solutions obtained from sampled parameter values, demonstrating strong agreement across the spatial domain. Quantitatively, we achieve an average coefficient of determination $R^2 = 0.98$ across all sampled parameters, confirming that the inverse neural operator successfully captures the distribution of diffusion coefficients consistent with the observed data. This high reconstruction fidelity validates that the probabilistic framework appropriately quantifies parameter uncertainty while ensuring physical consistency.}

\subsection{Burgers' equation}
\label{subsec:burgers}

\noindent In this example, we consider the viscous Burgers' equation defined as:
\begin{equation}\label{burgers_eqn_IBC}
    \begin{split}
&\frac{\partial u}{\partial t} (x,t)=  \nu \frac{\partial^2 u}{\partial x^2} (x,t)- u\frac{\partial u}{\partial x} (x,t) \text{ on }  \ \Omega: (x,t) \in [0,1]^2 , \\
\text{IC:  } &u(x,0)= f(x), \\
\textnormal{BC:  } &u(0,t)= u(1,t)  \text{ and } \frac{\partial u}{\partial x}(0,t) =  \frac{\partial u}{\partial x}(1,t), \\
  \end{split}
\end{equation}
where $u(x,t)$ denotes the evolving spatio-temporal velocity field defined using the space and time coordinates, $(x,t)$. In \autoref{burgers_eqn_IBC}, \textit{BC} denotes periodic boundary condition. Similar to the previous example, we will first present the results of discovering the unknown physics using DeepONet, followed by presenting the results of employing DeepONet to characterize the viscosity of the system using sparse measurements.

\vspace{6pt}
\noindent \textbf{Discovering physics using DeepONet}

\noindent In this problem, we re-write \autoref{burgers_eqn_IBC} as:
\begin{equation}\label{burgers_eqn_hidden}
\frac{\partial u}{\partial t} (x,t) =  \mathcal{N}(u,u_x,u_{xx}),  
\end{equation}
where the $\mathcal{N}$ denotes the hidden physics that is a function of the solution field $u(x,t)$ and its spatial derivatives $u_x, u_{xx}$. In this problem, our aim is to learn the solution operator, $\mathcal{G}_{\theta}$ that maps the initial condition, $f(x,0)$ to the solution field, $u(x,t)$ using the partial known physics as well as sparse labeled dataset. In this case, we have considered fixed viscosity for the system, $\nu = 0.01$. We solved equation \eqref{burgers_eqn_IBC} using a finite difference solver on a uniform grid ($\delta x = \delta t = 0.01$), resulting in a $101 \times 101$ grid spanning the $x$-$t$ domain with initial conditions sampled from a Gaussian Random Field (GRF).
 
\begin{figure}[!!htb]
\centering
\begin{subfigure}{0.45\textwidth}
    \centering
    \includegraphics[width=\linewidth]{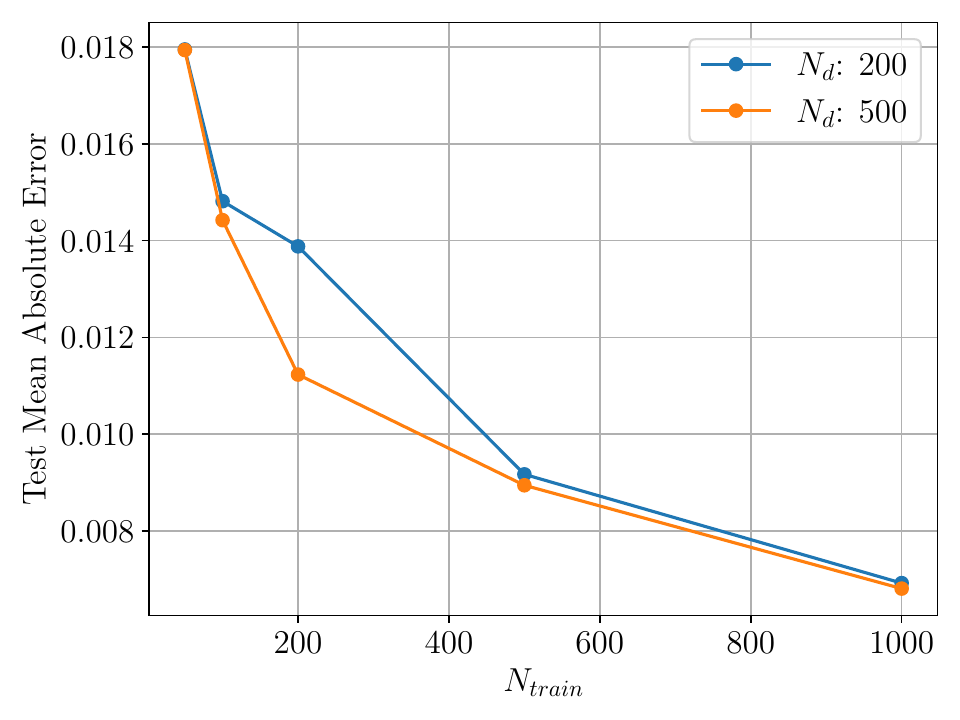}  
\end{subfigure}
\begin{subfigure}{0.45\textwidth}
    \centering
    \includegraphics[width=\linewidth]{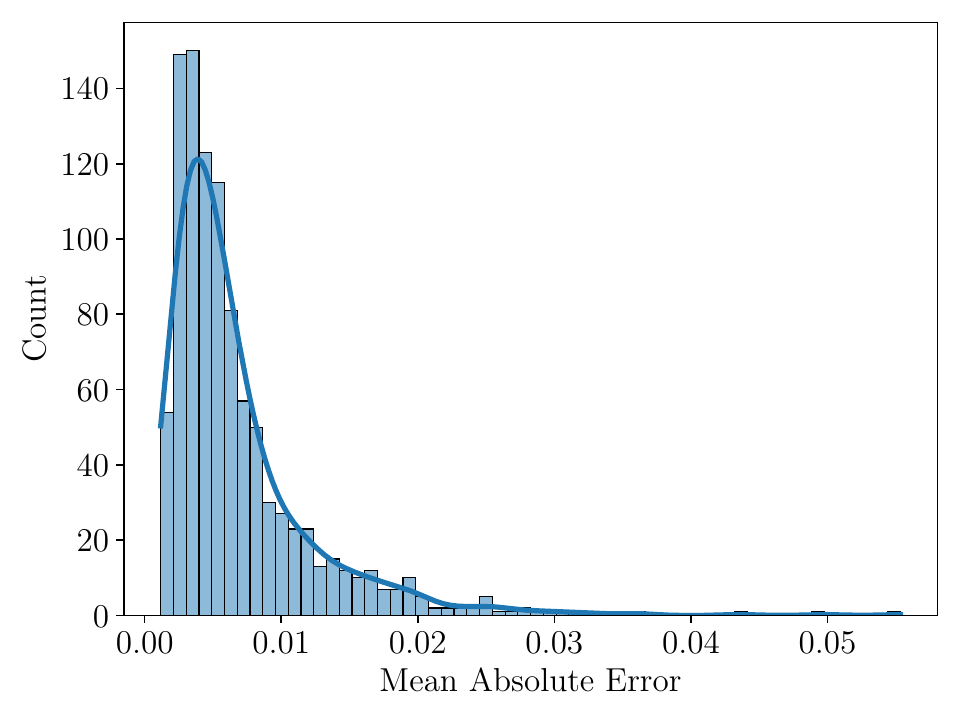}  
\end{subfigure}
\caption{Performance analysis for Burgers' equation: (a) Mean test error over varying $N_{\text{train}}$. (b) Test error distribution of optimal model that achieved an error $0.00680 \pm  0.00564$ with $N_{train} = 1000$, $N_{d} = 500$.}
\label{error_burgers}
\end{figure}

The experiments were conducted with varying numbers of training samples, $N_{train} \in \{50, 100, 200, 500, 1000\}$ and the size of labeled dataset, $N_d \in \{200, 500\}$, where $N_d$ denotes the number of locations per sample where the value of solution field is known. For testing, we used a fixed set of $N_{test} = 1,000$ samples. Figure~\ref{error_burgers}(a) illustrates the mean test error across different combinations of sample sizes and training data points. The optimal performance was achieved with $N_{train} = 1000$ and $N_d = 500$, resulting in mean absolute error of $0.00680 \pm  0.00564$.
The loss curve over the training process is shown in Figure~\ref{fig:loss_plot}(b).

\begin{figure}[!!htb]
\centering

\begin{subfigure}{0.495\textwidth}
    \centering
    \includegraphics[width=1.15\linewidth]{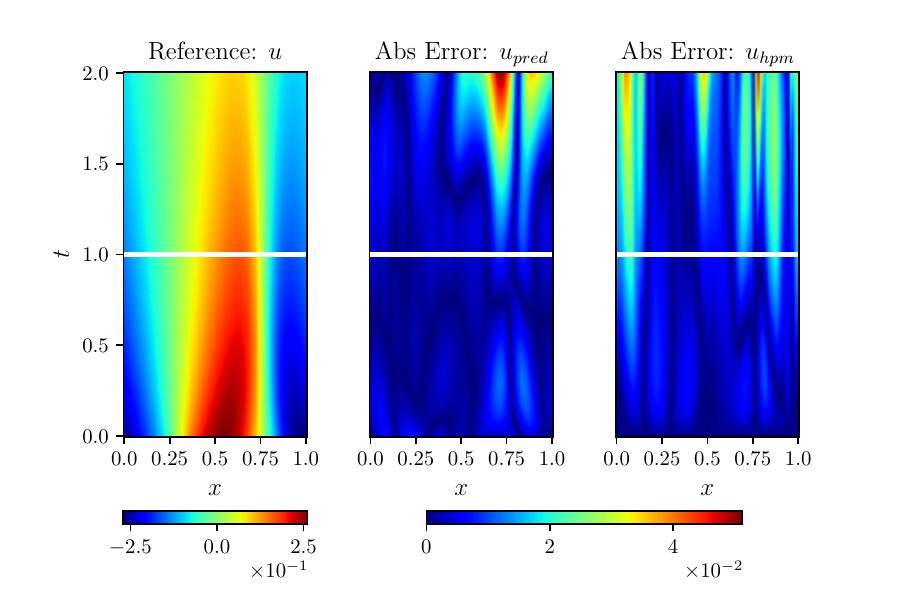}
    \caption{Sample 1}
\end{subfigure}
\hfill
\begin{subfigure}{0.495\textwidth}
    \centering
    \includegraphics[width=1.15\linewidth]{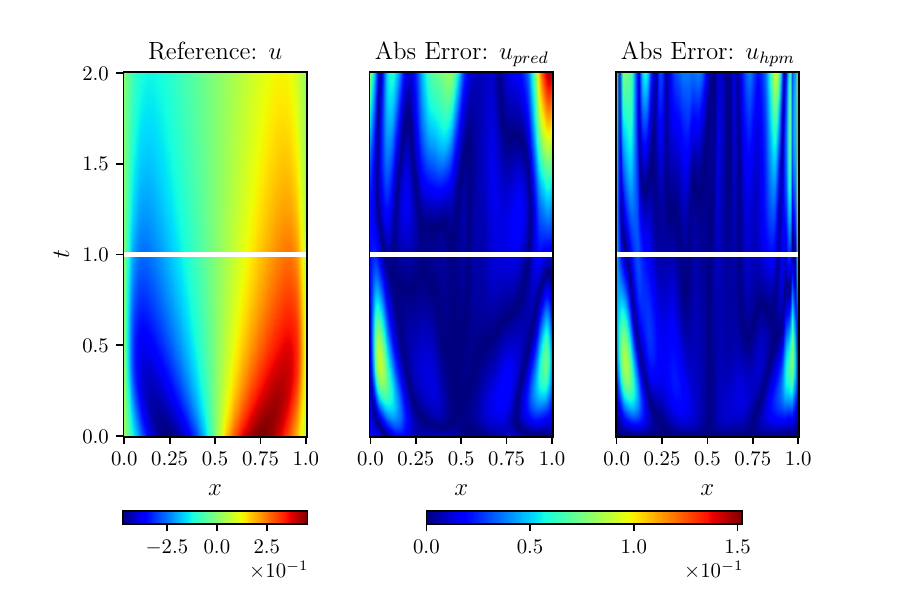}
    \caption{Sample 2}
\end{subfigure}

\caption{Burgers' equation: reference vs.\ DHPO predictions and HPM-integrated trajectories for two representative test samples. 
Training data spans $t\le1$, and the extrapolation region ($t>1$) is indicated by white horizontal lines.  
For Sample 1, mean absolute errors are  
$u_{pred}(t\le1)=0.00250$, $u_{pred}(t>1)=0.00883$,  
$u_{hpm}(t\le1)=0.00456$, $u_{hpm}(t>1)=0.01198$.  
For Sample 2, mean absolute errors are  
$u_{pred}(t\le1)=0.01268$, $u_{pred}(t>1)=0.02325$,  
$u_{hpm}(t\le1)=0.01454$, $u_{hpm}(t>1)=0.02128$.}
\label{fig:burgers_extp_samples}
\end{figure}

The distribution of test errors for this optimal configuration is presented in Figure~\ref{error_burgers}(b). As demonstrated in Figure~\ref{fig:burgers_extp_samples}, the model accurately predicts the solution when compared against ground truth. We also observe, in extrapolation region, the model accurately predicts the solution up to a specific time threshold, 
Hence we can infer that the Hidden Physics MLP is accurately capturing the dynamics, The Figure~\ref{fig:burgers_hid_phy_samples} shows the contours of PDE reference and predicted PDE terms, the model successfully approximated the unknown hidden term as,
$$\mathcal{N} \approx 0.01 \frac{\partial^2 u}{\partial x^2}(x,t) - u(x,t)\frac{\partial u}{\partial x} (x,t)$$

\noindent \textit{Relation to symbolic regression.} For comparison, we applied weak-SINDy \cite{messenger2021weak}, a sparse symbolic regression method, to 300 randomly sampled measurements from a single trajectory, recovering $u_t = 0.0101 \cdot u_{xx} - 1.0068 \cdot u \cdot u_x$ with residual $3.95 \times 10^{-3}$. While weak-SINDy efficiently discovers symbolic equations from limited data, DHPO learns generalizable operators of solution across varying initial conditions with physics discovery as an integrated component. The methods address complementary aspects of the inverse problem.

\noindent\textbf{System parameter identification: Viscosity}

\noindent \D{\textbf{Deterministic.}} We are now interested in employing the neural operator framework discussed in Section~\ref{subsec:parameter_identification} to characterize the system, more specifically to identify the viscosity of the model while being provided with the solution values at a few spatial and temporal locations.

To design the problem, we have considered varying viscosity as well as initial conditions for the Burgers' equation defined in Equation~\ref{burgers_eqn_IBC}. For data generation, we solved this equation using a finite difference solver on a uniform grid with spatial resolution $\delta x = 0.01$ and time step $\delta t = 0.01$, resulting in a $101 \times 101$ grid spanning the $x$-$t$ domain. The dataset comprises solutions corresponding to 500 different viscosity values ($\nu$) uniformly distributed between $0.01$ and $0.05$. Each viscosity value was paired with 20 distinct initial conditions sampled from a Gaussian Random Field (GRF), yielding a total of 10,000 solution fields. We randomly selected $N_{\text{train}} = 8,500$ samples for training and reserved the remaining $N_{\text{test}} = 1,500$ for testing the model. To simulate sparse measurements, we randomly sampled 300 spatial-temporal locations (fixed across all training and testing samples) from the solution field, where the value of $u(x^s,t^s)$ is known.

The hyperparameters involved in the training of the networks are presented in Table \ref{tab:hyperparameter}. The training configuration utilizes $N_{\text{coll}} = 2500$ collocation points for evaluating the PDE residual. The loss functions were weighted with unity penalty coefficients.

\begin{figure}[!h]
\centering
\begin{subfigure}{0.9\textwidth}
    \centering
    \includegraphics[width=0.9\linewidth]{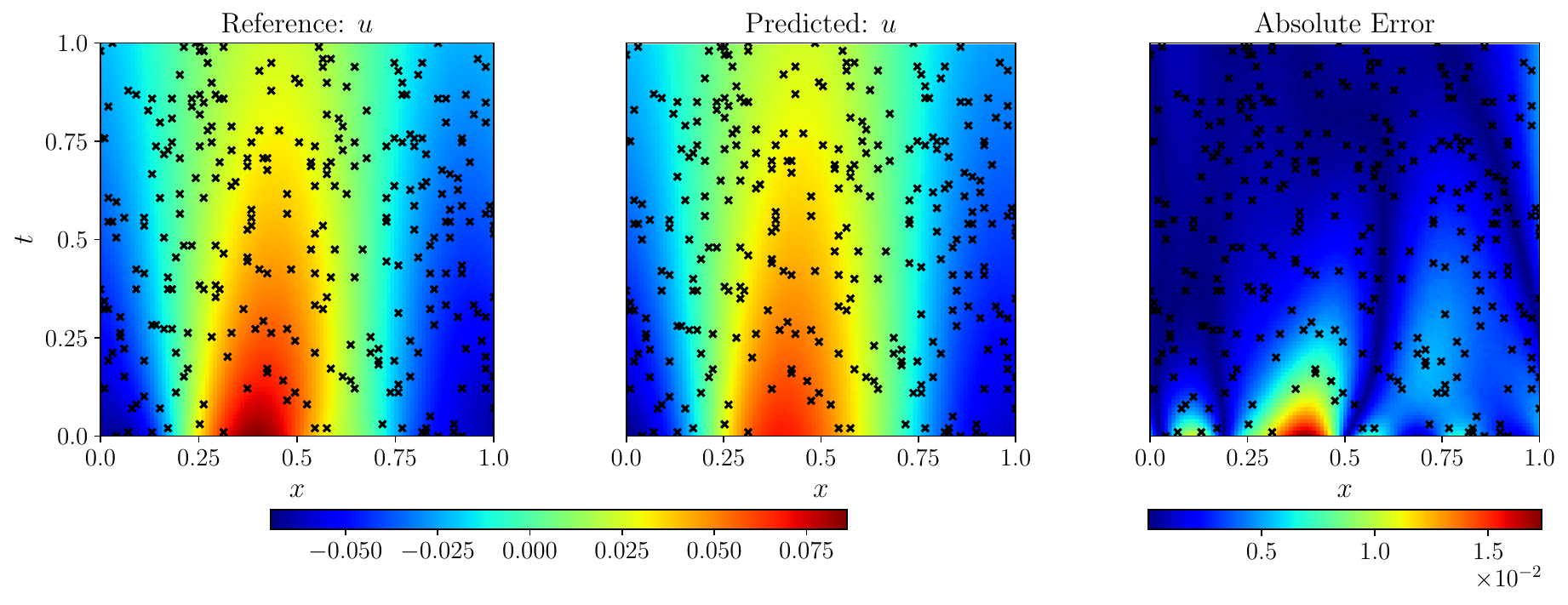}  
    \caption{Sample 1: Relative $L_2$ error of $u(x,t)$ = $0.092$. For this case, $\nu_{\text{true}} = 0.028$ while $\nu_{\text{predicted}} = 0.025$.}
    \label{Burgers_inv_sample_500}
\end{subfigure}
\begin{subfigure}{0.9\textwidth}
    \centering
    \includegraphics[width=0.9\linewidth]{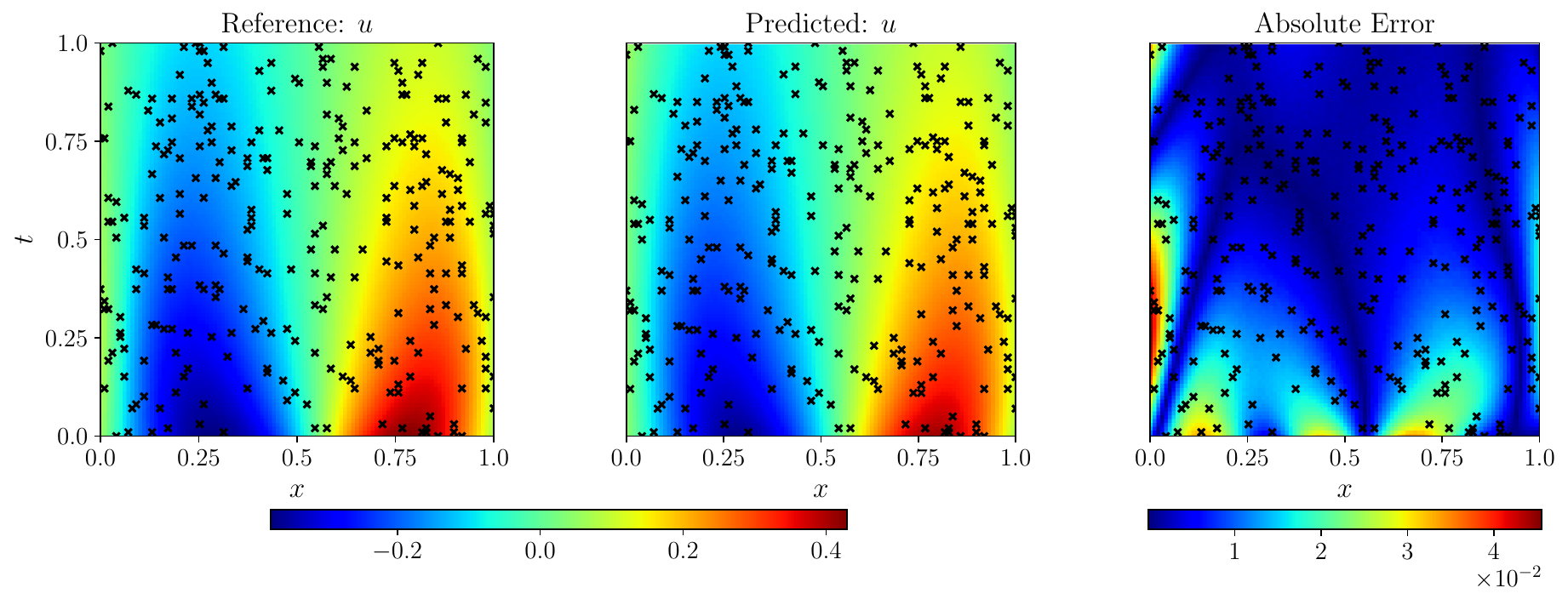}  
    \caption{Sample 2: Relative $L_2$ error of $u(x,t)$ = $0.061$. For this case, $\nu_{\text{true}} = 0.032$ while $\nu_{\text{predicted}} = 0.032$.}
    \label{Burgers_inv_sample_700}
\end{subfigure}
\caption{Burgers' Equation: Comparison of the reference solution and the predicted solution. The black cross marks represent the location where the data of $u(x,t)$ was known during the testing. The error at these specific locations is also shown in the absolute error plot.}
\label{Burgers_inv_sample_500_n_700}
\end{figure}

\begin{figure}[!htb]
    \centering
    \includegraphics[width=0.8\linewidth]{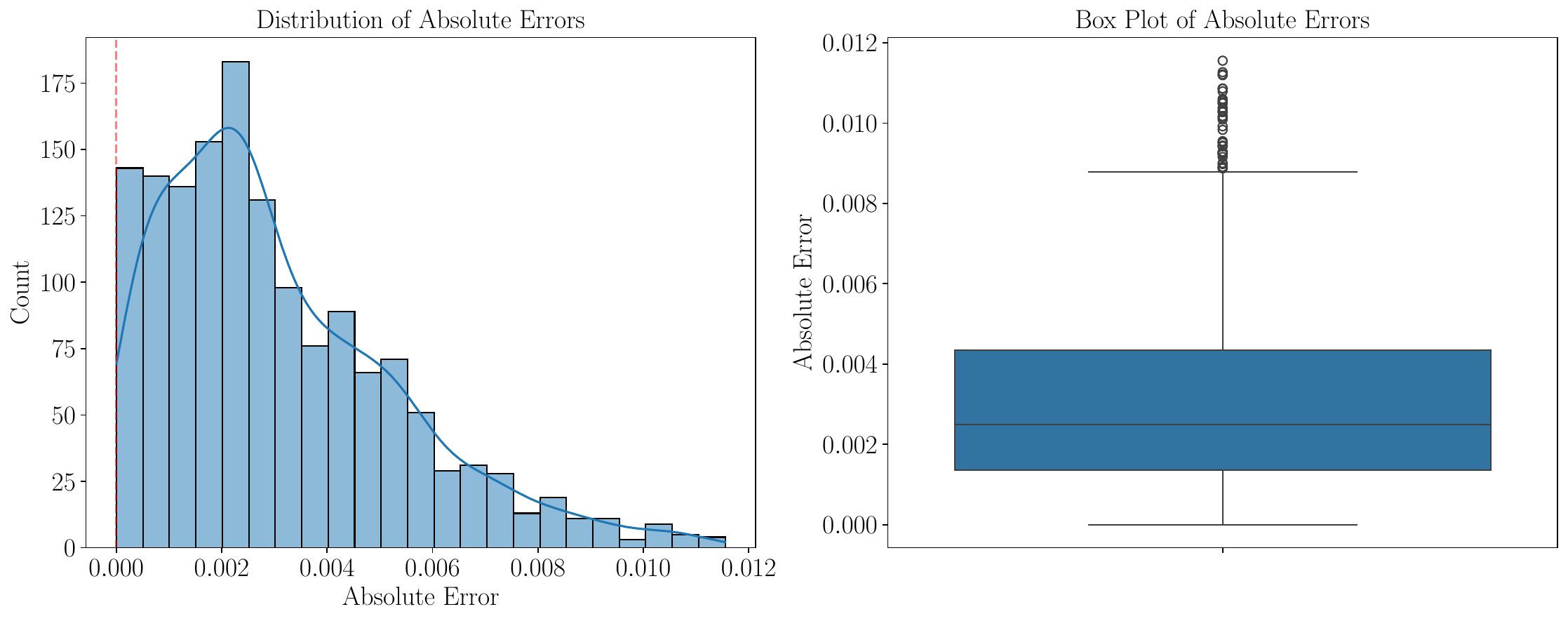}
    \caption{Burgers' Equation: Distribution of absolute error of $\nu$ when compared to the $\nu$ used to generate the ground truth using the finite difference solver. The framework approximates $\nu$ with an error of $2\times 10^{-3}\pm 0.002$.}
    \label{Burgers_inv_error_dist}
\end{figure}

Figure \ref{Burgers_inv_sample_500_n_700} demonstrates the framework's performance in reconstructing both the solution field $u(x,t)$ and estimating the viscosity parameter $\nu$ from $300$ sparse measurements. We present results from two representative test cases selected from our test set of $1,500$ samples. The framework achieves solution reconstruction with $L_2$ errors of order $\mathcal{O}(10^{-2})$, while accurately recovering the true viscosity values in both cases. To provide a comprehensive assessment of the parameter estimation accuracy, Figure \ref{Burgers_inv_error_dist} presents the distribution of absolute errors in predicted viscosity values across the entire test set. The distribution reveals that the majority of predictions exhibit absolute errors below $0.002$, demonstrating the robust performance of our framework in parameter estimation tasks.

\begin{figure}[!htb]
    \centering
    \includegraphics[width=0.8\linewidth]{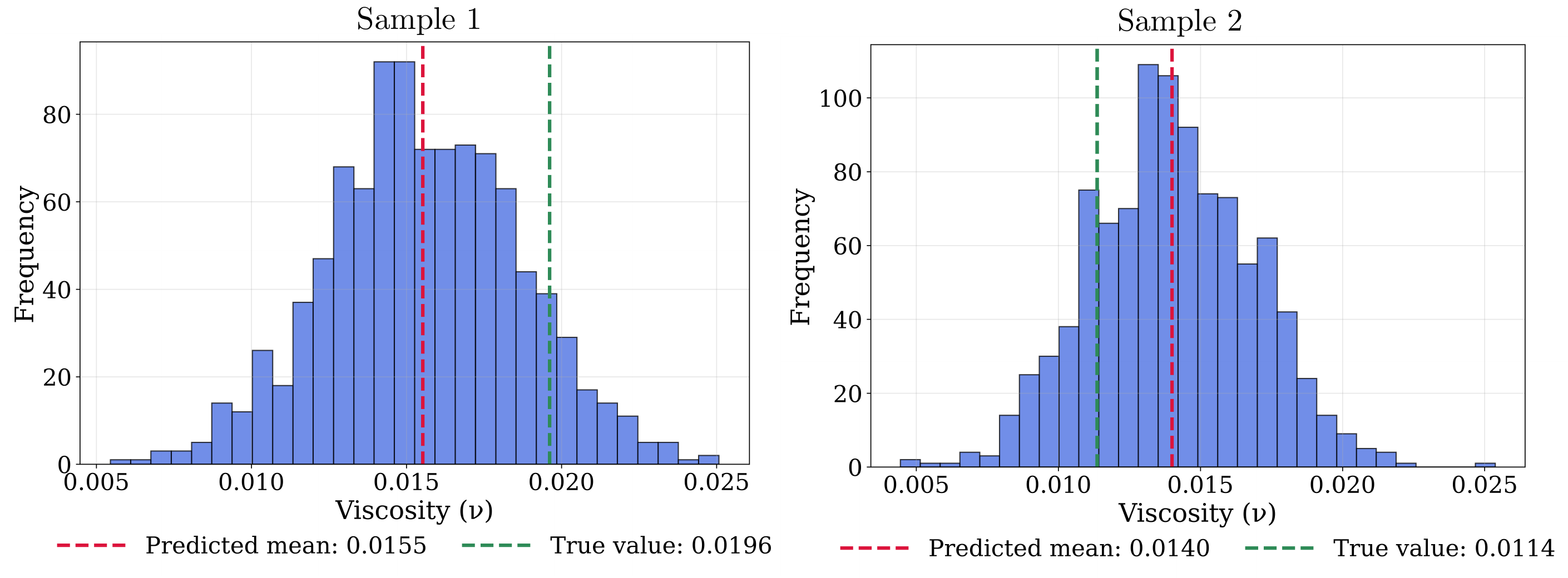}
    \caption{\D{Burgers' Equation: Predicted probability distributions of viscosity coefficient $\nu$ for two representative test samples. The histogram represents 500 sampled values from the learned distribution $\mathcal{N}(\mu_\nu, \sigma_\nu^2)$. Dashed red line indicates the predicted mean, dashed green line shows the true viscosity value.}}
    \label{fig:Burgers_inv_nu_dist}
\end{figure}

\noindent \D{\textbf{Probabilistic.} We apply the same probabilistic framework to estimate the distribution of viscosity coefficient $\nu$ for Burgers' equation. Following the validation procedure described above, we sample 500 viscosity values from the predicted distribution $\mathcal{N}(\mu_\nu, \sigma_\nu^2)$ for each test case and perform forward finite difference simulations to assess whether these sampled parameters can reproduce the original sensor observations. Figure~\ref{fig:Burgers_inv_nu_dist} presents the predicted parameter distributions for two representative test samples. In both cases, the predicted mean (dashed red line) closely approximates the true viscosity value (dashed green line), while the distribution width reflects the epistemic uncertainty arising from sparse sensor measurements. Notably, Sample 2 exhibits a narrower distribution compared to Sample 1, indicating higher confidence in parameter estimation. This variation arises from differences in the information content of the respective sensor observations.}

\begin{figure}[!htb]
    \centering
    \includegraphics[width=\linewidth]{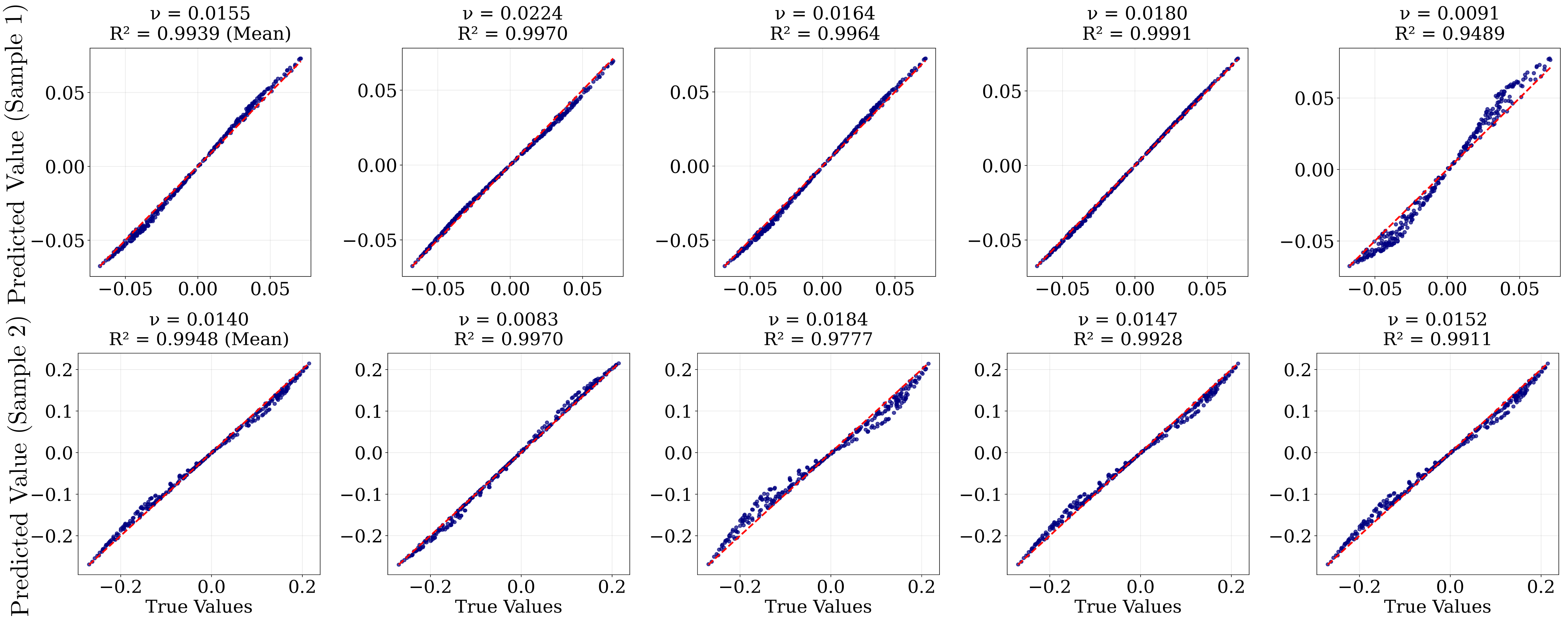}
    \caption{\D{Burgers' Equation: Validation of predicted parameter distribution for representative samples through forward finite difference simulations. Each subplot shows the true solution field $u(x,t)$ (x-axis) versus the solution obtained using a viscosity value sampled from the predicted distribution (y-axis).}}
    \label{fig:Burgers_inv_pred_u_sampled_nu}
\end{figure}

\D{Figure~\ref{fig:Burgers_inv_pred_u_sampled_nu} validates the learned distributions through forward simulation: for each sample, we evaluate 5 different sampled viscosity values spanning the predicted distribution, and for each value, we compare the resulting solution field against the true field used to generate the input sensor data. The scatter plots demonstrate exceptional agreement across all sampled parameters, with coefficient of determination values exceeding $R^2 > 0.99$ in all cases confirms the physical consistency of the learned parameter distribution.}

\noindent To validate the posterior distribution, we perform forward finite difference simulations using sampled viscosity values and compare the resulting solutions against the reference solution obtained with the true viscosity. We distinguish between coefficient of determination ($R^2$) at the 300 sensor locations and relative $L_2$ error comparing the complete solution fields. Table~\ref{tab:burgers_posterior_validation} presents both metrics for 10 viscosity values sampled from the posterior for Sample 2. \\
The sampled values span a twofold range (0.0108 to 0.0212). When each sampled viscosity is used to solve Burgers' equation forward, the resulting solutions exhibit full field $L_2$ errors from 0\% to 17\% relative to the true solution. This confirms that parameter variations significantly affect the complete PDE solution. However, at the 300 sensor locations, all sampled solutions achieve $R^2 > 0.95$ when compared to the reference sensor values, with most exceeding 0.99. Even the most extreme sample ($\nu = 0.0108$, 44\% below true value) attains $R^2 = 0.9571$ at sensors.

\begin{table}[!htb]
\centering
\footnotesize

\caption{Validation of posterior parameter distribution for Burgers' equation (Sample 2). Each row shows a viscosity value sampled from $\mathcal{N}(\mu_\nu, \sigma_\nu^2)$. For each sample, we solve Burgers' equation using finite differences and compare the resulting solution field against the reference solution obtained with the true viscosity ($\nu = 0.01922$). The coefficient of determination measures agreement at the 300 sensor locations, while the relative $L_2$ error measures agreement over the complete 101$\times$101 spatio-temporal grid.}
\label{tab:burgers_posterior_validation}
\begin{tabular}{lccc}
\hline
Sample & Viscosity ($\nu$) & $R^2$ at sensors & Full field rel. $L_2$ error \\
\hline
True (ref) & 0.01922 & 0.9990 & 0.0000 \\
$\nu_1$ (mean) & 0.01465 & 0.9863 & 0.0853 \\
$\nu_2$ & 0.01731 & 0.9961 & 0.0339 \\
$\nu_3$ & 0.01592 & 0.9919 & 0.0602 \\
$\nu_4$ & 0.01672 & 0.9945 & 0.0450 \\
$\nu_5$ & 0.02016 & 0.9995 & 0.0160 \\
$\nu_6$ & 0.01225 & 0.9705 & 0.1364 \\
$\nu_7$ & 0.01633 & 0.9933 & 0.0523 \\
$\nu_8$ & 0.02122 & 0.9992 & 0.0333 \\
$\nu_9$ & 0.01091 & 0.9583 & 0.1671 \\
$\nu_{10}$ & 0.01080 & 0.9571 & 0.1699 \\
\hline
\end{tabular}
\end{table}

\DNew{This behavior is consistent with the inherent non-uniqueness of parameter estimation from sparse observations. Multiple parameter configurations produce forward solutions that match observed sensor data equally well (high $R^2$) while showing moderate differences in the complete solution field (up to $17\%$ relative $L_2$ error). Solution variations are concentrated in unobserved regions between sensors, where measurements provide limited constraint. The posterior distribution captures the range of viscosity values consistent with the sparse observations, providing a quantitative measure of the epistemic uncertainty inherent to parameter estimation from limited data.}

\subsection{2D Heat equation}
\label{subsec:heat}

\noindent In the \D{next} example, we consider the heat equation over a L-shaped 2D computational domain $\Omega_L = \Omega \setminus R = \left[(-0.5, 0.5) \times (-0.5, 0.5)\right] \setminus \left[[0, 0.5] \times [0, 0.5]\right]$, defined as:
\begin{equation}\label{2D_heat}
    \begin{split}
&\frac{\partial u}{\partial t} (x,y,t)=  \mathcal{K} (\frac{\partial^2 u}{\partial x^2} (x,y,t) + \frac{\partial^2 u}{\partial y^2} (x,y,t)) \text{ on }  \ \Omega_L \times [0,1] , \\
\text{IC:  } &u(x,y,0)= f(x,y), \\
\textnormal{BC:  } & u(-0.5, y, t) = u(x<0, 0.5, t)  = u(0, y>0, t) = \\
&   u(x>0, 0, t) = u(0.5, y<0, t) = u(x,-0.5,t) = 0
  \end{split}
\end{equation}
where $u(x,y,t)$ denotes the evolving spatio-temporal temperature field and $\mathcal{K}$ represents of the thermal conductivity coefficient. IC in \autoref{2D_heat} denotes the initial condition sampled from GRF and BC represents the boundary condition. For data generation, we employed a finite element solver (deal.ii \cite{arndt2021deal}) on a uniform grid with spatial resolution $\delta x \approx 0.03$ and time step $\delta t = 0.01$, resulting in a total of 833 grid points covering the $x$-$y$ domain and 101 time steps.
Similar to the previous examples, we will first present the results of discovering the unknown physics using DeepONet, followed by presenting the results of employing DeepONet to characterize the thermal conductivity coefficient of the system using sparse measurements.

\vspace{6pt}
\noindent \textbf{Discovering physics using DeepONet}

\noindent We learn the mapping between different ICs to the solution field using DHPO. We consider the RHS of \autoref{2D_heat} unknown and re-write the equation as: 
\begin{equation}\label{heat_eqn_hidden}
\frac{\partial u}{\partial t} =  \mathcal{N}(u,u_x,u_y,u_{xx},u_{yy}), 
\end{equation}
where $\mathcal{N}$ is the hidden physics operator. The trunk network takes the spatio-temporal coordinates $(x,y,t)$ as input, while the branch network consists of a convolutional neural network (CNN) followed by a fully connected network (FCN). The CNN processes a $33 \times 33$ image generated from the initial condition $f(x,y)$. The outputs of the trunk and branch networks are combined via a dot product, producing a scalar output that approximates the solution field $u$. In addition, we employ a hidden-physics neural network that receives the tuple $[u, u_x, u_y, u_{xx}, u_{yy}]$ as input and outputs a scalar quantity $\mathcal{N}$, representing the unknown PDE terms. Spatial derivatives are computed using reverse mode AD.  

For training and validation, the thermal conductivity is fixed at $0.015$. A total of 500 training samples are generated by varying the initial condition. The overall loss function includes contributions from the initial condition loss, boundary condition loss, data loss, and equation residual loss. For the data loss, we sample $50/833$ spatial and $11/101$ uniformly spaced temporal points. The initial condition loss is evaluated at 200 fixed spatial locations across all samples. The boundary loss is computed at $400 \times 100$ collocation points, corresponding to 400 spatial boundary locations at 100 time instants. The equation residual loss is evaluated at $500 \times 100$ randomly sampled collocation points in space and time, with new samples drawn for each batch and epoch. The evolution of the total loss during training is shown in Figure~\ref{fig:loss_plot}(c). For testing, 1000 samples are evaluated, each on a grid of $833 \times 101$ spatio-temporal points. 

\begin{figure}[!htb]
    \centering
    \includegraphics[width=0.5\linewidth]{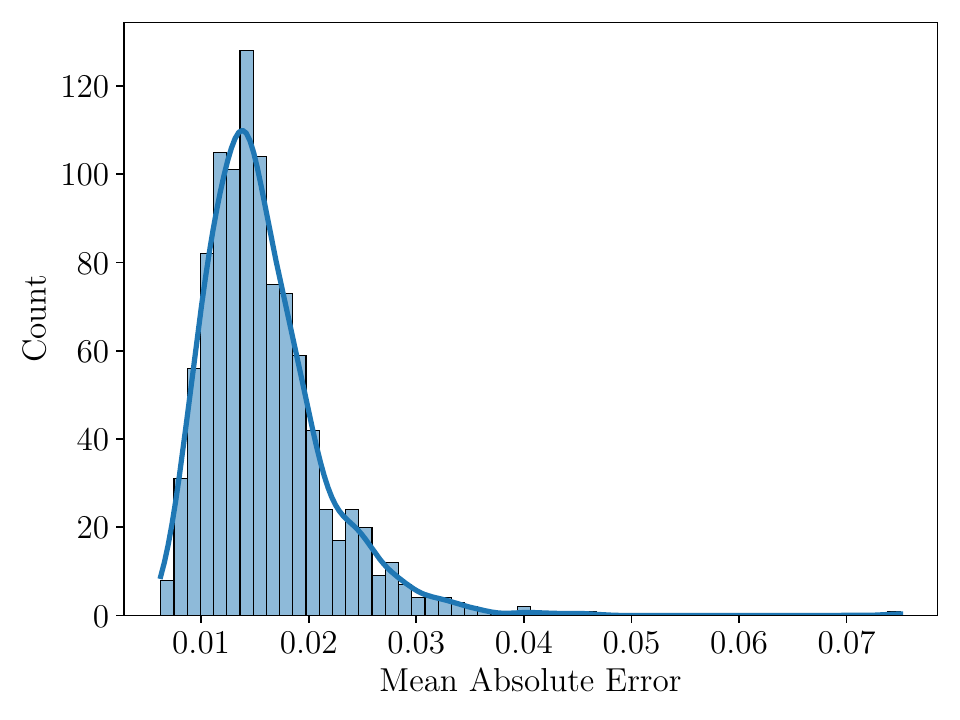}
    \caption{Heat equation: Test Error distribution using 1000 test samples, model achieved test mean absolute error of $0.01579 \pm 0.00567 $.  }
    \label{fig:dhpm_heat_2d_test_dist}
\end{figure}

\begin{figure}[h]
\centering
\includegraphics[width=\linewidth]{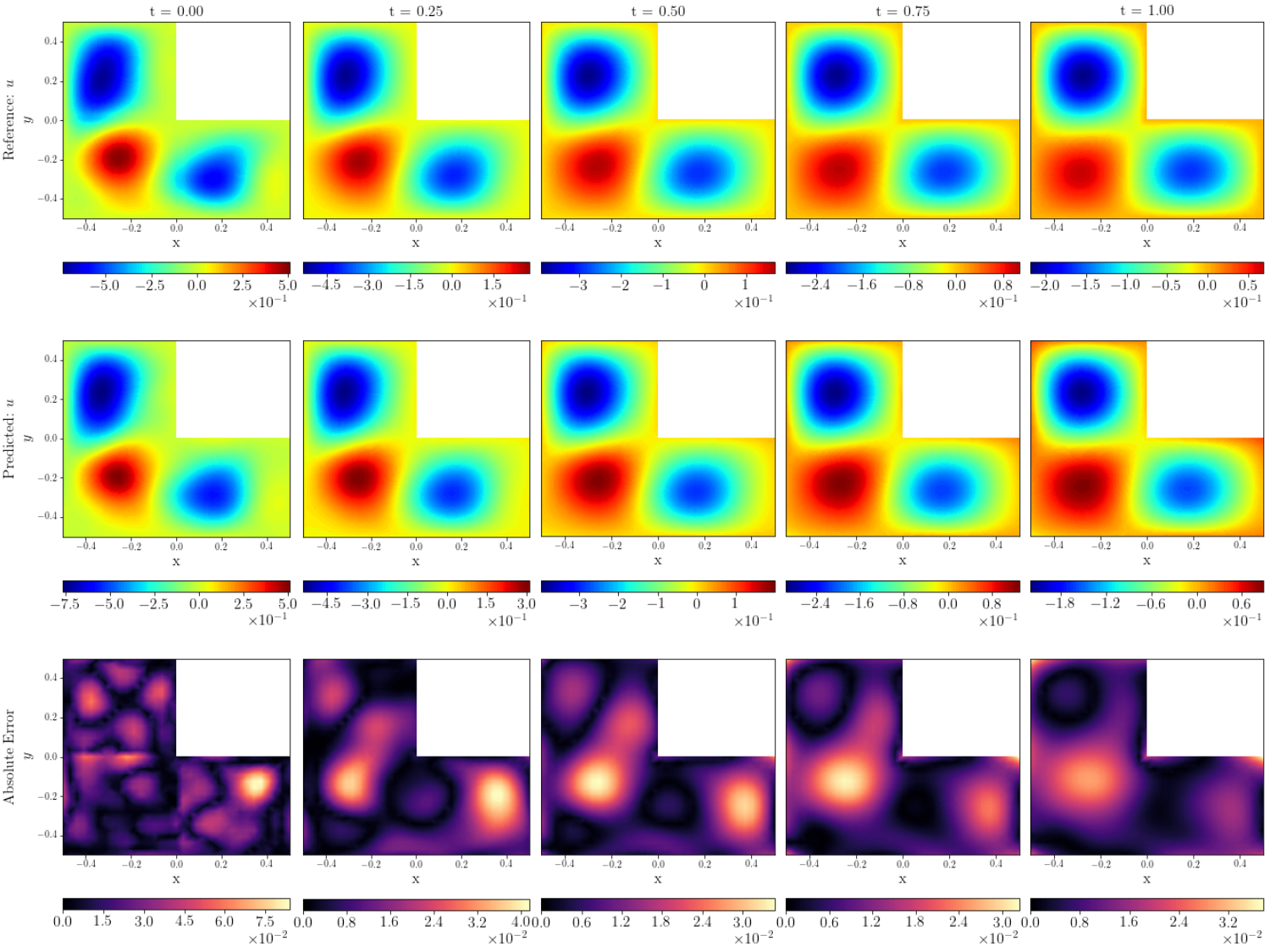}
\caption{Heat equation: reference vs predicted solution field for a representative test sample (mean absolute error = 0.01064).}
\label{fig:dhpm_heat2d_soln_single}
\end{figure}

The distribution of test errors is shown in Figure~\ref{fig:dhpm_heat_2d_test_dist}, yielding a mean absolute error of $0.01579 \pm 0.00567$. Predictions for one representative test samples are presented in Figure~\ref{fig:dhpm_heat2d_soln_single}, where the reconstructed solution fields closely match the reference, with mean absolute error  of $0.01064$, respectively.  

\vspace{6pt}
\noindent\textbf{System parameter identification: Thermal conductivity coefficient}

\noindent \noindent \D{\textbf{Deterministic.}} We now aim to apply the neural operator framework introduced in Section~\ref{subsec:parameter_identification} to characterize the system by identifying the thermal conductivity coefficient, given access to solution values at a limited number of spatial and temporal locations.

To set up this problem, we consider a Heat equation model with varying thermal conductivity coefficients and initial conditions, as defined in \autoref{2D_heat}. The dataset includes solutions for 100 different thermal conductivity values ($\mathcal{K}$), uniformly sampled in the range $[0.01, 0.2]$. Each value of $\mathcal{K}$ was paired with 100 distinct initial conditions sampled from a Gaussian Random Field (GRF), yielding a total of 10,000 solution instances. From these, we randomly selected $N_{\text{train}} = 8,000$ samples for training and used the remaining $N_{\text{test}} = 2,000$ samples for evaluation. To emulate sparse observations, we placed 40 sensors evenly across the spatial domain and recorded temperature values at 11 time steps within the interval $t \in [0, 1]$.

The hyperparameters used during training are summarized in Table~\ref{tab:hyperparameter}. The training setup includes $N_{\text{coll}} = 2000$ collocation points for computing the PDE residual, and all loss terms are assigned unit penalty weights.

\begin{figure}[!htb]
    \centering
    \includegraphics[width=0.8\linewidth]{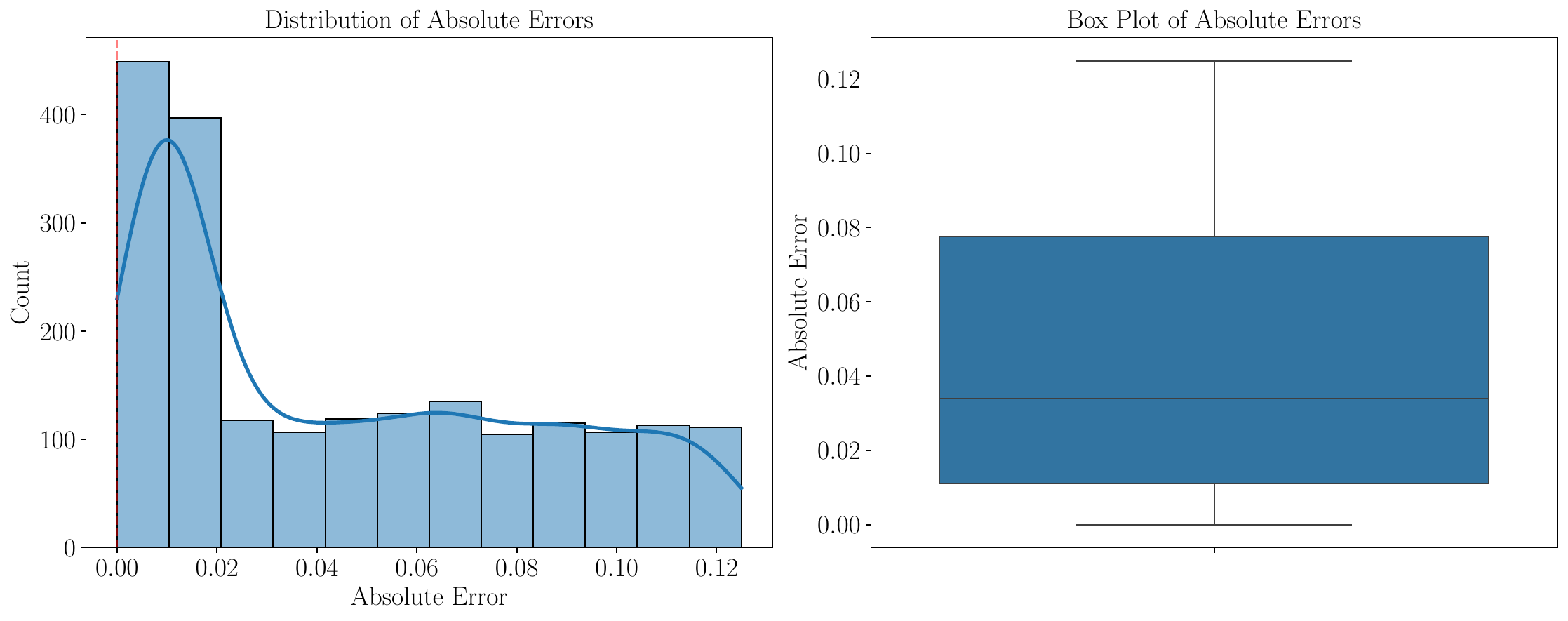}
    \caption{Heat' Equation: Distribution of absolute error of $\mathcal{K}$ when compared to the $\mathcal{K}$ used to generate the ground truth using the finite element solver. The framework approximates $\mathcal{K}$ with an error of $3\times 10^{-2}\pm 0.04$.}
    \label{heat_inv_error_dist}
\end{figure}

\begin{figure}[!htb]
    \centering
    \includegraphics[width=1.0\linewidth]{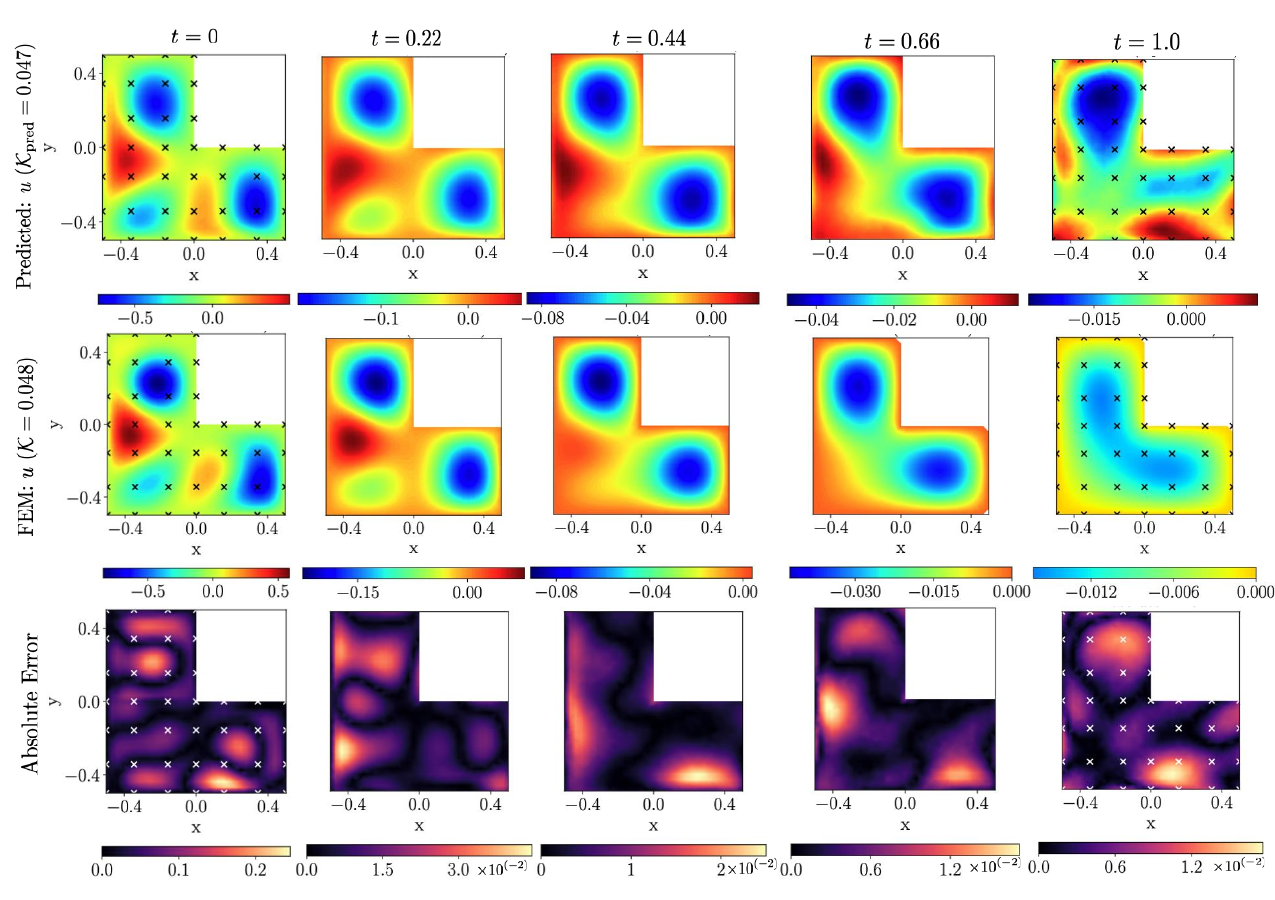}
    \caption{Heat Equation: Comparison of reference solution and predicted solution. Black and white cross marks indicate sensor locations where u(x,y,t) data was available during testing. Sensor readings were collected at 11 time steps; no sensor data is shown at t=0.22, t=0.44 and t=0.66. The corresponding absolute errors at these sensor locations are displayed in the error plot.}
    \label{Heat_inv_sample546}
\end{figure}

Figure~\ref{Heat_inv_sample546} illustrates the performance of our framework in reconstructing both the solution field $u(x,y,t)$ and the thermal conductivity coefficient $\mathcal{K}$ from a total of 440 sparse measurements. We display results from a representative test case drawn from the test set of 2,000 samples. The reconstruction achieves $L_2$ errors on the order of $\mathcal{O}(10^{-2})$, and the inferred values of $\mathcal{K}$ are reasonably accurate, despite the inverse problem's inherent non-uniqueness. To further evaluate the accuracy of parameter estimation, Figure~\ref{heat_inv_error_dist} shows the distribution of absolute errors in the predicted conductivity values across the test set. The framework approximates K with an error of $3 \times10^{-2} \pm 0.04$, confirming the robustness of our framework for thermal parameter identification. \D{The high standard deviation relative to the mean reflects the inherent variability in inverse heat conduction problems, where parameter identifiability from sparse sensors depends strongly on the thermal conductivity magnitude and initial condition structure. The bimodal distribution shows that approximately $65\%$ of cases achieve errors below $0.02$, while more challenging cases with high conductivity values or low spatial variation in initial conditions exhibit reduced identifiability. The median error of $0.028$ provides a more robust characterization of typical performance, with 72\% of predictions achieving errors below $0.03$.}

\begin{figure}[!htb]
    \centering
    \includegraphics[width=0.8\linewidth]{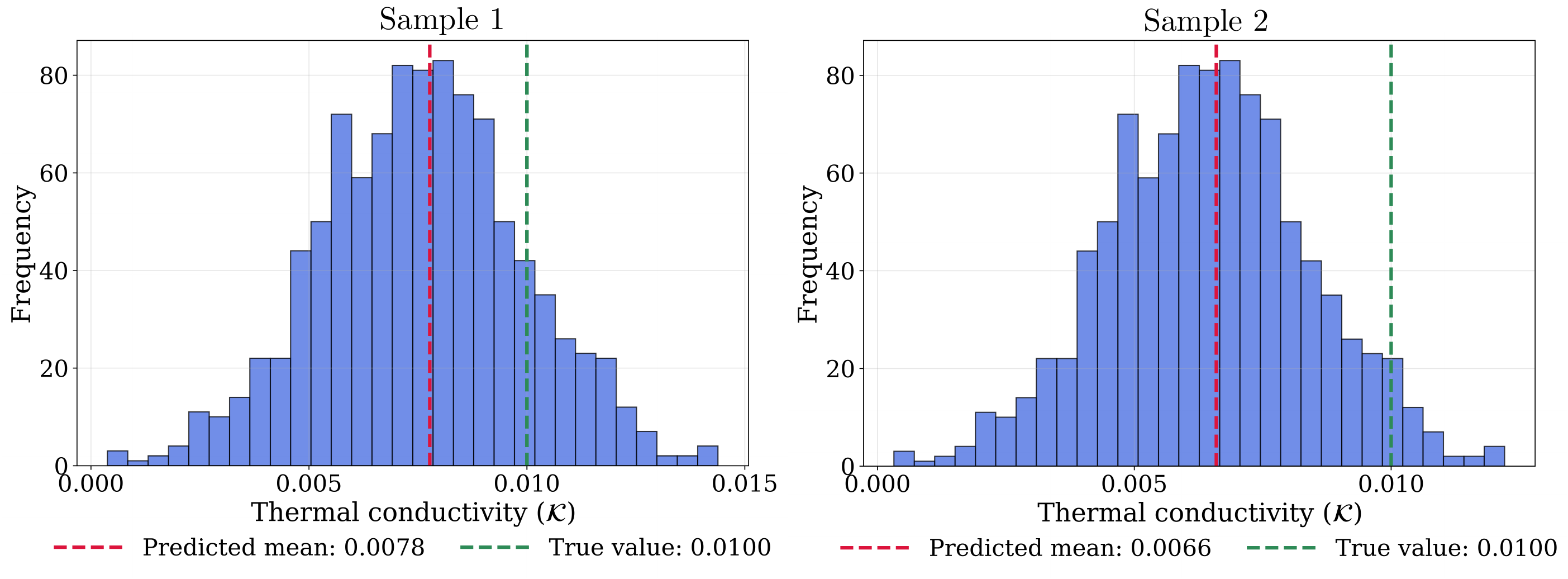}
    \caption{\D{Heat Equation: Predicted probability distributions of thermal coefficient $\mathcal{K}$ for two representative test samples. The histogram represents 500 sampled values from the learned distribution $\mathcal{N}(\mu_\mathcal{K}, \sigma_\mathcal{K}^2)$. Dashed red line indicates the predicted mean, dashed green line shows the true thermal coefficient value.}}
    \label{fig:heat_inv_k_dist}
\end{figure}

\noindent \D{\textbf{Probabilistic.} We apply the same probabilistic framework to estimate the distribution of thermal conductivity coefficient $K$ for the Heat equation. Following the validation procedure described above, we sample 500 thermal conductivity values from the predicted distribution $\mathcal{N}(\mu_K, \sigma_K^2)$ for each test case and perform forward finite element simulations to assess whether these sampled parameters can reproduce the original sensor observations. Figure~\ref{fig:heat_inv_k_dist} presents the predicted parameter distributions for two representative test samples. In both cases, the predicted mean (dashed red line) closely approximates the true thermal conductivity value (dashed green line), while the distribution width reflects the epistemic uncertainty arising from sparse sensor measurements distributed across the L-shaped domain. Both samples exhibit similar distribution widths ($\sigma_K \approx 0.002$), indicating comparable confidence levels in parameter estimation despite different initial temperature profiles.}

\begin{figure}[!htb]
    \centering
    \includegraphics[width=\linewidth]{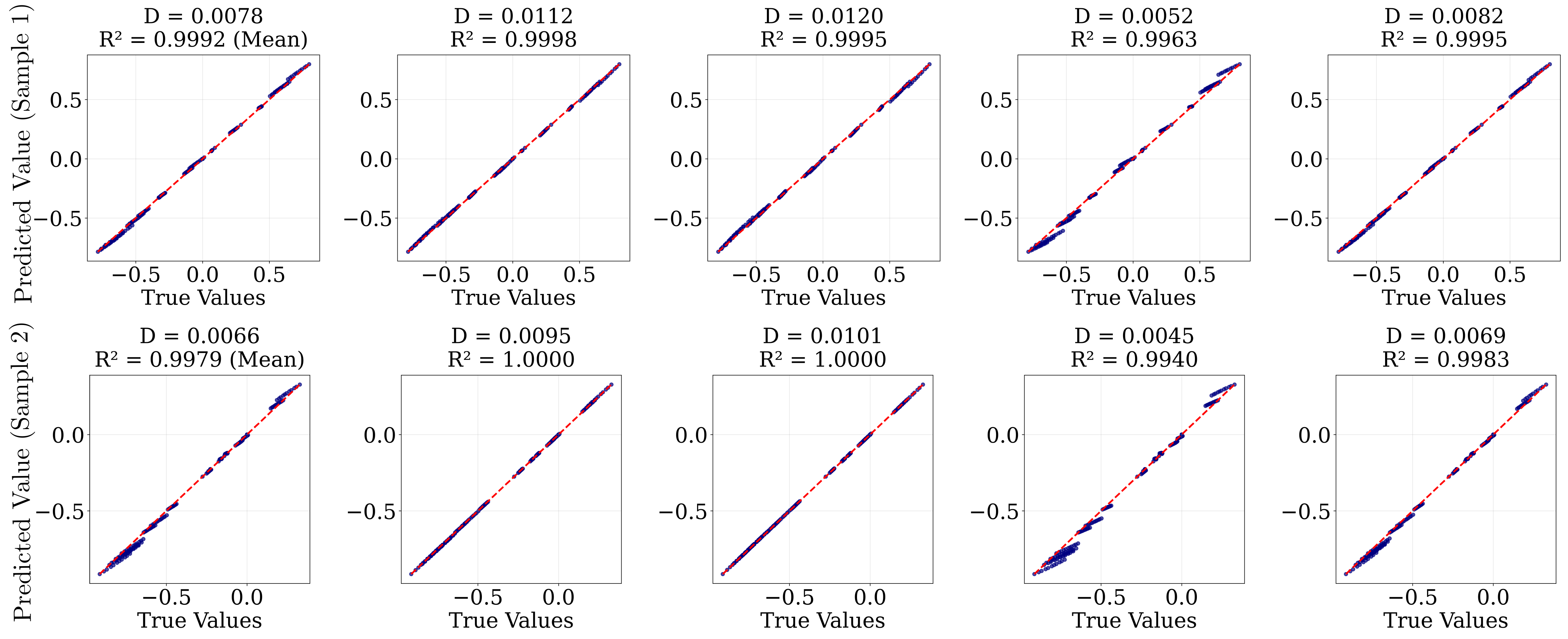 }
    \caption{\D{Heat Equation: Validation of predicted parameter distribution for representative samples through forward finite difference simulations. Each subplot shows the true solution field $u(x,t)$ (x-axis) versus the solution obtained using a thermal coefficient value sampled from the predicted distribution (y-axis).}}
    \label{fig:heat_inv_pred_u_sampled_k}
\end{figure}

\D{Figure~\ref{fig:heat_inv_pred_u_sampled_k} validates the learned distributions through forward simulation: for each sample, we evaluate 10 different sampled thermal conductivity values spanning the predicted distribution, and for each value, we compare the resulting temperature field against the true field used to generate the input sensor data. The scatter plots demonstrate exceptional agreement across all sampled parameters, with coefficient of determination values reaching $R^2 \geq 0.99$ in all cases, with several achieving perfect correlation ($R^2 = 1.0000$).}

\subsection{\D{2D Helmholtz equation: Source reconstruction}}
\label{subsec:helmholtz}

\D{In the next problem, we consider the steady state wave equation defined as:} 

\begin{equation}\label{helmholtz_eqn_IBC}
    \begin{split}
& \nabla^2 u(x,y) + k^2u(x,y) = g(x,y) \text{ on }  \ \Omega: (x,y) \in [0,1]^2 , \\
 \textnormal{BC:  } & u(0,y)= u(1,y)= u(x,0)= u(x,1) = 0   \\
  \end{split}
\end{equation}

\D{\noindent where $u(x,y)$ denotes the steady-state wave field, $g(x, y)$ represents an unknown source distribution that we aim to reconstruct, and the wave number $k$ is set to $2\pi$. To make the inverse problem tractable, $g(x,y)$ is parameterized by a Gaussian function as:}
\begin{equation}\label{gaussian_source}
g(x,y) = A \exp\left(- \frac{(x-x_0)^2 + (y-y_0)^2}{2 \sigma^2}\right)
\end{equation}
\D{where the standard deviation is fixed at $\sigma = 0.2$, while the amplitude $A$ and center location $(x_0, y_0)$ are treated as unknown parameters to be determined from sparse measurements of $u(x,y)$. For data generation, we employed a Finite difference Helmholtz solver on a uniform grid with spatial resolution $\delta x = 0.015$, resulting in $4096$ grid points. The dataset includes $800$ samples in which the three parameters $A, x_0, \text{ and } y_0$ are sampled from uniform distributions over the bounds $[2, 6]$, $[0.35, 0.65]$, and $[0.35, 0.65]$, respectively. }

\noindent \D{\textbf{Deterministic.} From the dataset, we randomly selected $N_{\text{train}} = 640$ samples for training and used the remaining $N_{\text{test}} = 160$ samples for testing. We placed $32 \times 32$ uniformly spaced sensors across the domain to capture data for source reconstruction. The hyperparameters used during training are summarized in Table~\ref{tab:hyperparameter}. The training setup includes $N_{\text{coll}} = 4096$ collocation points for computing the PDE residual. We applied a penalty weight of $10$ to the data loss term, while all other loss terms were assigned unit penalty weights.} 

\begin{figure}[!htb]
    \centering
    \includegraphics[width=0.9\linewidth]{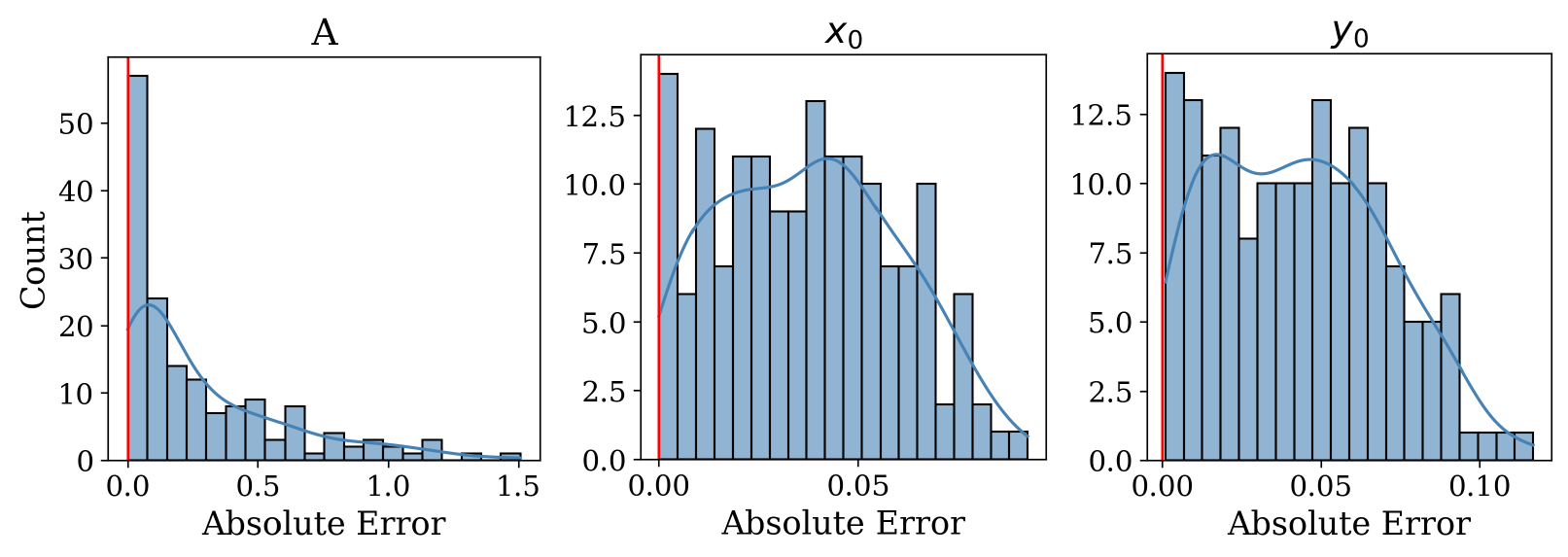}
    \caption{\D{Helmholtz Equation:
Distribution of absolute errors for the inferred parameters $A$, $x_{0}$, and $y_{0}$. The mean absolute errors across 160 samples are $2.82\times10^{-1}\pm3.19\times10^{-1}$ for $A$, $3.41\times10^{-2}\pm2.25\times10^{-2}$ for $x_{0}$, and $4.44\times10^{-2}\pm3.03\times10^{-2}$ for $y_{0}$.}
}

    \label{fig:helm_inv_error_dist}
\end{figure}

\D{Figure~\ref{fig:helm_inv_error_dist} illustrates the distribution of absolute errors obtained in the recovery of the Helmholtz source parameters $A$, $x_{0}$, and $y_{0}$ across the entire test set. The inferred amplitudes exhibit a broader spread, with a mean absolute error of $0.28\pm0.32$, reflecting the higher sensitivity of the solution field to variations in source strength. In contrast, the spatial parameters $x_{0}$ and $y_{0}$ are recovered with substantially smaller errors of $3.4\times 10^{-2}\pm2.25\times 10^{-2}$ and $4.4\times 10^{-2}\pm3.03\times 10^{-2}$, respectively. These results highlight the ability of the inverse model to accurately infer both the location and magnitude of the underlying Gaussian source from the observed PDE solutions, even in the presence of moderate variability in amplitude estimation.}

\begin{figure}[!htb]
    \centering
    \includegraphics[width=0.8\linewidth]{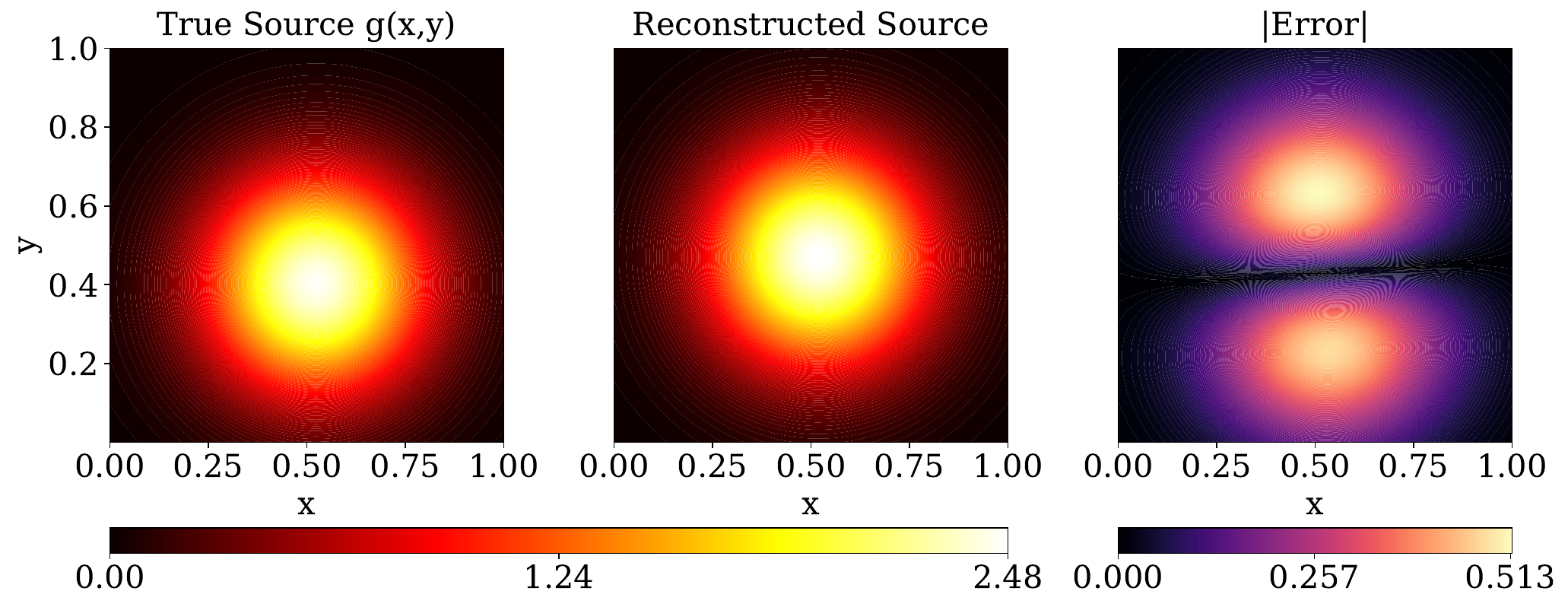}
    \caption{\D{Helmholtz equation: Source identification.
The left panel shows the true Gaussian source $g(x,y)$, the middle panel shows the reconstructed source, and the right panel shows the pointwise absolute error. The reconstructed source attains a relative $L_{2}$ error of $0.2358$. For this sample, the ground-truth source parameters are:
$A_{\text{true}} = 2.4482$, $x_{\text{true}} = 0.5237$, and $y_{\text{true}} = 0.4031$.
The predicted parameters are:
$A_{\text{pred}} = 2.4772$, $x_{\text{pred}} = 0.5192$, and $y_{\text{pred}} = 0.4699$.
The corresponding absolute errors are:
$|A_{\text{true}} - A_{\text{pred}}| = 0.02896$,
$|x_{\text{true}} - x_{\text{pred}}| = 0.00451$,
and $|y_{\text{true}} - y_{\text{pred}}| = 0.06682$.}}
    \label{fig:helmholtz_source}
\end{figure}

\begin{figure}[!htb]
    \centering
    \includegraphics[width=0.8\linewidth]{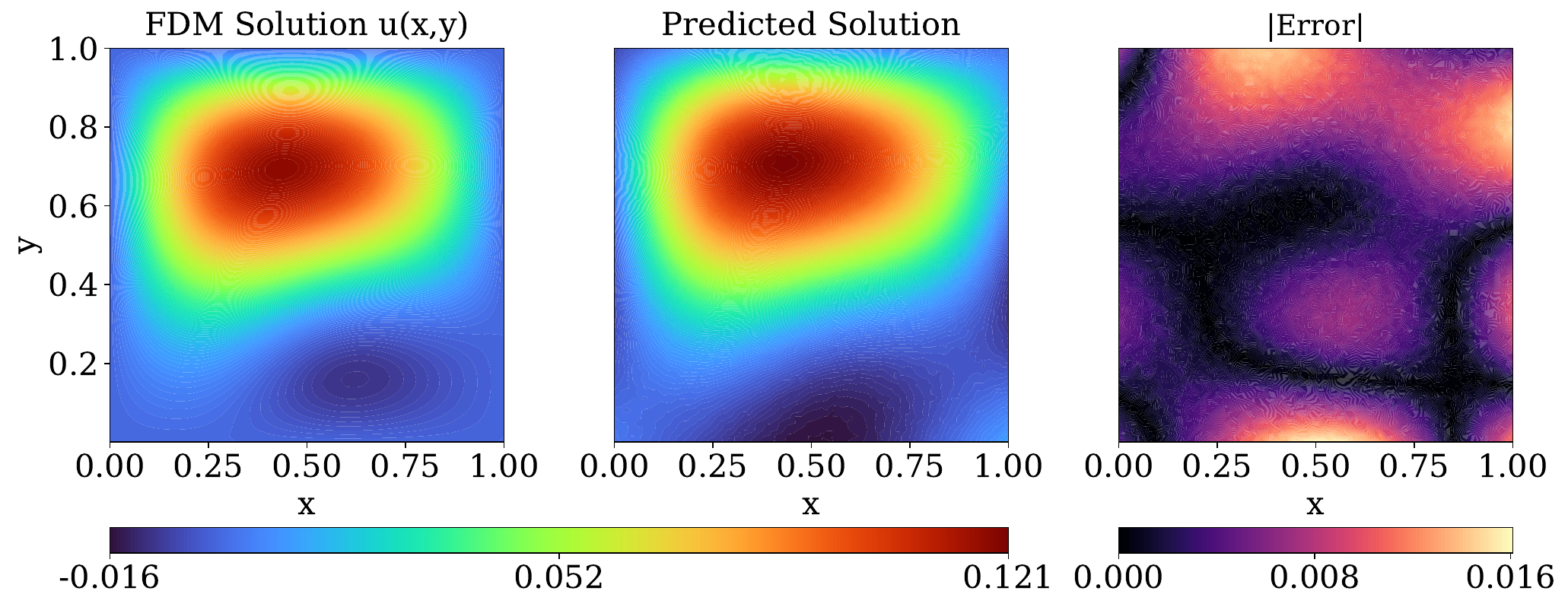}
    \caption{\D{Helmholtz equation: Solution comparison.
The left panel shows the FDM solution $u(x,y)$, the middle panel presents the predicted solution obtained from the learned operator, and the right panel displays the pointwise absolute error between the two fields. The solution field has a relative $L_{2}$ error of $0.1219$ and an MSE of $3.51\times 10^{-5}$.}
}
    \label{fig:helmholtz_sol}
\end{figure}

\D{The mean relative $L_2$ error for the predicted solution field $u(x,y)$ is $10.19\%$, while the source reconstruction achieves a mean relative $L_2$ error of $22.38\%$. For individual parameter estimation, the model achieves mean absolute errors of $0.33$ for amplitude $A$, $0.036$ for $x_0$, and $0.045$ for $y_0$, corresponding to mean relative errors of $6.85\%$, $7.68\%$, and $9.81\%$, respectively, indicating strong predictive performance for the amplitude and reasonable accuracy for the spatial location parameters. Figures~\ref{fig:helmholtz_source} and \ref{fig:helmholtz_sol} present comparisons of the reconstructed source and the corresponding solution prediction for a representative test sample.}

\noindent \D{\textbf{Probabilistic.} We apply the probabilistic framework to the Helmholtz source reconstruction problem, which involves estimating three parameters simultaneously: the amplitude $A$, and the center coordinates $(x_0, y_0)$ of a Gaussian source. This represents a more challenging multi-parameter inverse problem compared to the single-parameter cases above. Following the validation procedure, we sample 500 parameter sets from the predicted joint distribution, where each parameter follows $A \sim \mathcal{N}(\mu_A, \sigma_A^2)$, $x_0 \sim \mathcal{N}(\mu_{x_0}, \sigma_{x_0}^2)$, and $y_0 \sim \mathcal{N}(\mu_{y_0}, \sigma_{y_0}^2)$, and perform forward Finite difference Helmholtz simulations to assess data reconstruction capability. }

\begin{figure}[!htb]
    \centering
    \includegraphics[width=\linewidth]{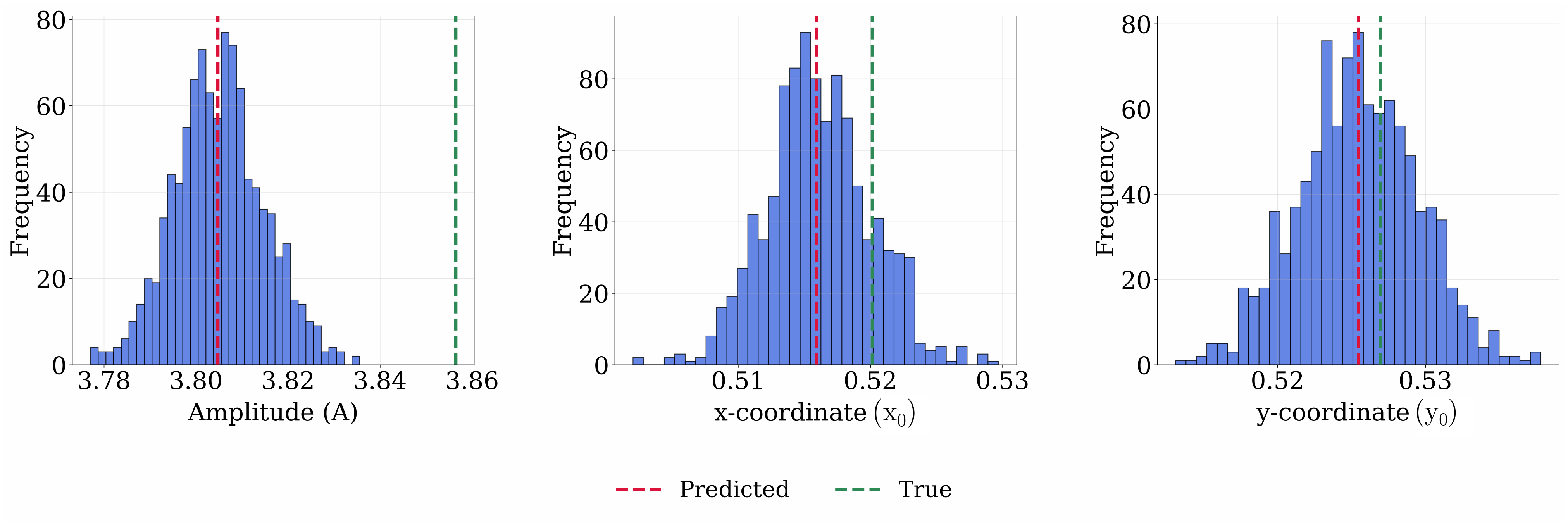}
    \caption{\D{Helmholtz Equation: Predicted probability distributions of Amplitude $A$, x-coordinate of center $x_0$ and y-coordinate of center $y_0$ for two representative test samples. The histogram represents 500 sampled values from the learned distribution $\mathcal{N}(\mu_A, \sigma_A)$, $\mathcal{N}(\mu_{x_0}, \sigma_{x_0})$ and $\mathcal{N}(\mu_{y_0}, \sigma_{y_0})$. Dashed red line indicates the predicted means, dashed green line shows the true parameter values.}}
    \label{fig:Helmholtz_inv_param_dist}
\end{figure}

\begin{figure}[!htb]
    \centering
    \includegraphics[width=\linewidth]{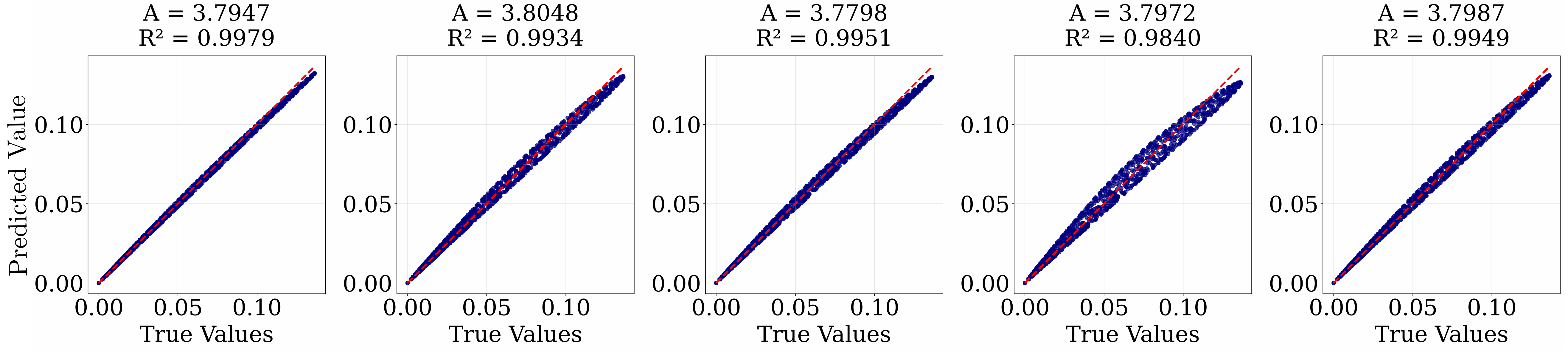}
    \caption{\D{Helmholtz Equation: Validation of predicted parameter distribution for representative samples through forward finite difference simulations. Each subplot shows the true solution field $u(x,t)$ (x-axis) versus the solution obtained using a value sampled from the predicted distributions of parameters (y-axis).}}
    \label{fig:Helmholtz_inv_pred_u_sampled_params}
\end{figure}

\D{Figure~\ref{fig:Helmholtz_inv_param_dist} presents the predicted distributions for all three parameters. The predicted means (dashed red lines) accurately capture the true parameter values (dashed green lines) for the spatial coordinates; however, the predicted mean for amplitude shows a noticeable offset from the true value. We hypothesize that this discrepancy might be due to a multimodal distribution of the amplitude for this problem. Notably, the spatial location parameters $x_0$ and $y_0$ exhibit relatively narrow distributions ($\sigma_{x_0} \approx 0.004$, $\sigma_{y_0} \approx 0.005$), indicating high confidence in source localization, while the amplitude parameter $A$ shows a broader distribution ($\sigma_A \approx 0.02$), reflecting greater uncertainty in estimating the source strength. This difference arises from the higher sensitivity of the wave field to source location compared to source amplitude, small spatial shifts produce more distinct measurement patterns than amplitude variations. Figure~\ref{fig:Helmholtz_inv_pred_u_sampled_params} validates the learned distributions through forward simulation: we sample 5 complete parameter sets $(A, x_0, y_0)$ from the joint distribution and compare the resulting wave fields against the true field. The scatter plots demonstrate exceptional agreement across all sampled parameter sets, with coefficient of determination values consistently exceeding $R^2 > 0.99$ for all cases, confirming that the probabilistic framework successfully captures the manifold of source parameters consistent with the observed wave field measurements.}


\textbf{Remark on parametric vs. non-parametric reconstruction.} The parametric Gaussian representation in Equation~\ref{gaussian_source} is a deliberate choice motivated by the fundamental ill-posedness of the inverse source problem. Direct reconstruction of the source field $g(x,y)$ on our $64 \times 64$ computational grid would require recovering 4096 field values from only 1024 sensor measurements. This yields a severely underdetermined system with information ratio 1:4.

Such problems possess a nullspace of dimension at least 3072, meaning infinitely many distinct source distributions produce identical sensor observations~\cite{isakov2006inverse}. The singular value decomposition (SVD) of the forward operator reveals that high-frequency source components correspond to rapidly decaying singular values~\cite{hansen2010discrete}. These components generate evanescent waves that decay exponentially with distance, rendering them effectively invisible to sensors. Even modest measurement noise causes error amplification proportional to $1/\sigma_{\min}$, violating the discrete Picard condition~\cite{hansen1987truncated, colton1998inverse}.

\begin{figure}[!htb]
    \centering
    \begin{subfigure}[b]{\linewidth}
        \centering
        \includegraphics[width=0.8\linewidth]{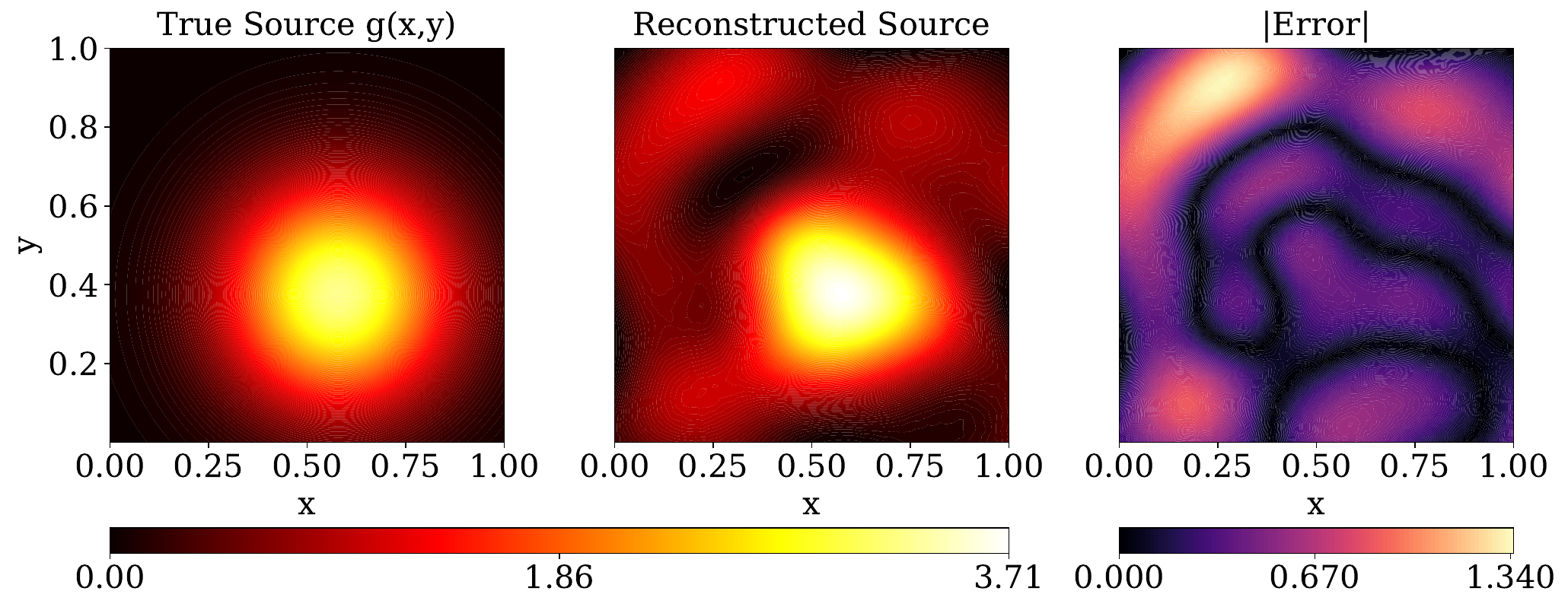}
        \caption{\DNew{MLP-based inverse operator: source reconstruction}}
        \label{fig:helmholtz_source_NP_MLP}
    \end{subfigure}
    
    \vspace{0.3cm}
    
    \begin{subfigure}[b]{\linewidth}
        \centering
        \includegraphics[width=0.8\linewidth]{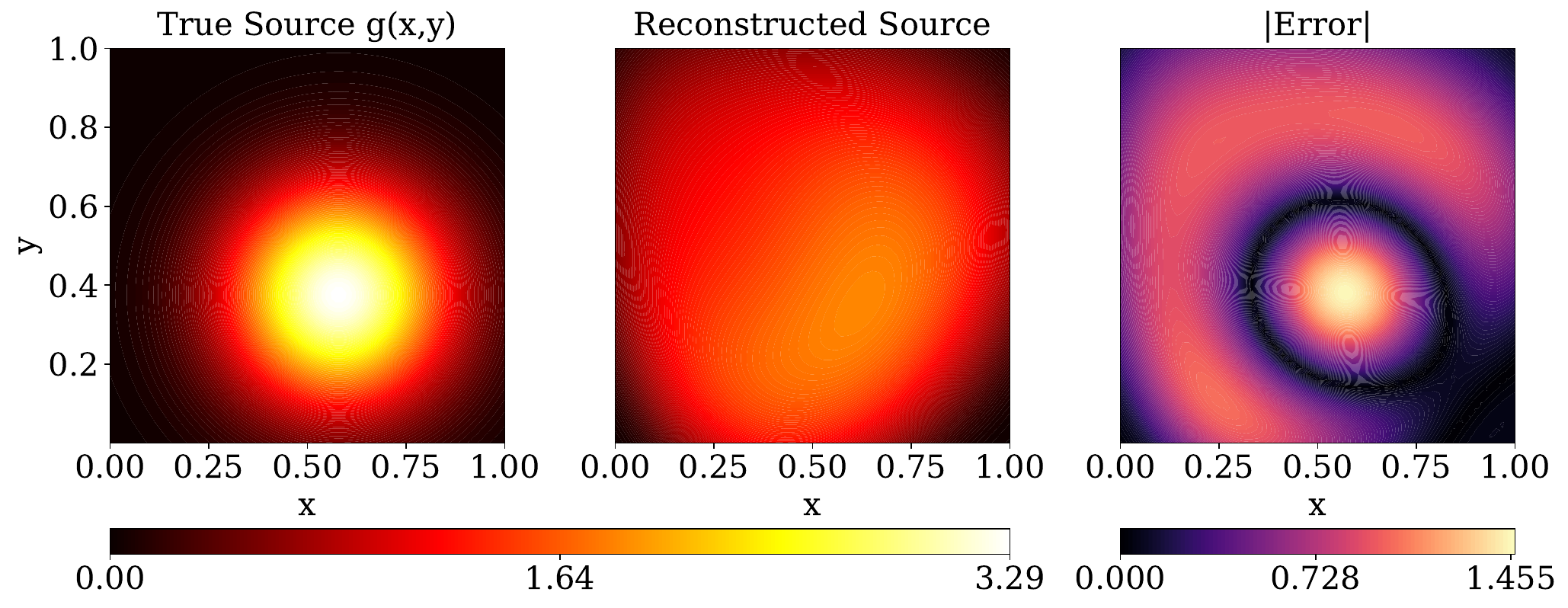}
        \caption{\DNew{DeepONet-based inverse operator: source reconstruction}}
        \label{fig:helmholtz_source_NP_DON}
    \end{subfigure}
    
    \caption{Helmholtz equation: Non-parametric source reconstruction comparison. (a) MLP-based approach with relative $L_2$ error of $0.3372$. (b) DeepONet-based approach with relative $L_2$ error of $0.5970$. For each subfigure: left panel shows the true Gaussian source $g(x,y)$, middle panel shows the reconstructed source, and right panel shows the pointwise absolute error. Both approaches correctly identify the approximate source location but fail to recover detailed spatial structure, confirming fundamental non-uniqueness.}
    \label{fig:helmholtz_source_NP_comparison}
\end{figure}

In contrast, our parametric formulation reduces the problem to the well-posed space $\mathbb{R}^3$ with information ratio $341:1$. This exploits the physical prior that acoustic sources, antennas, and seismic events are typically spatially localized~\cite{bao2005inverse, li2017recovering}. Alternative regularization strategies exist, including Tikhonov smoothness penalties~\cite{willoughby1979solutions}, total variation constraints~\cite{rudin1992nonlinear}, and sparsity-based methods~\cite{candes2006robust}. However, all require comparable prior assumptions to make the problem well-posed. Our parametric representation offers the advantage of direct physical interpretability: the parameters $(A, x_0, y_0)$ correspond to source strength and location, which are the quantities of interest in practical applications such as seismic source localization~\cite{waldhauser2000double} and electromagnetic inverse scattering~\cite{chen2018computational}.

To validate this theoretical analysis, we attempted direct field reconstruction on the full $64 \times 64$ grid (4096 field values). We modified our framework so the inverse operator outputs 4096 values instead of 3 parameters, keeping all other architectural choices identical. We conducted two experiments: (1) MLP-based inverse operator: A multilayer perceptron maps sensor measurements directly to source field values. (2) DeepONet-based inverse operator: A DeepONet architecture where the branch network processes sensor locations and the trunk network processes evaluation points $(x,y)$.

\begin{figure}[!htb]
    \centering
    \begin{subfigure}[b]{\linewidth}
        \centering
        \includegraphics[width=0.8\linewidth]{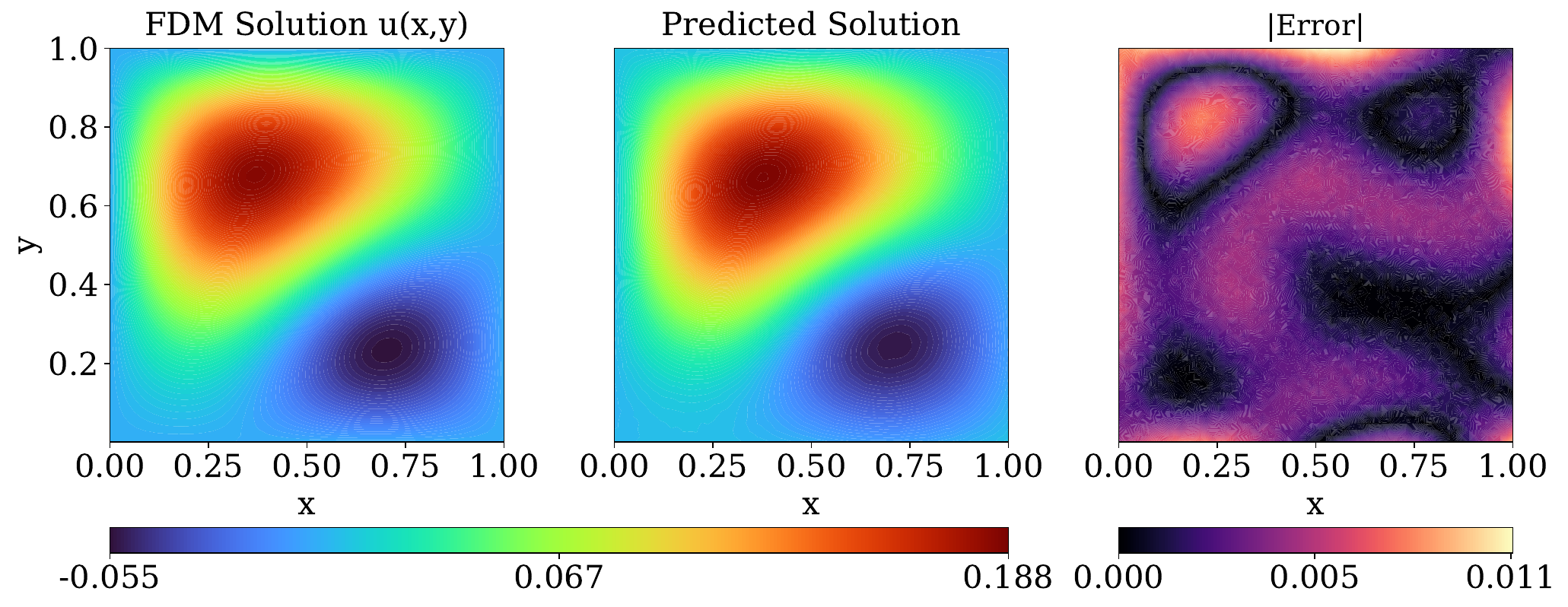}
        \caption{\DNew{MLP-based inverse operator: solution field}}
        \label{fig:helmholtz_sol_NP_MLP}
    \end{subfigure}
    
    \vspace{0.3cm}
    
    \begin{subfigure}[b]{\linewidth}
        \centering
        \includegraphics[width=0.8\linewidth]{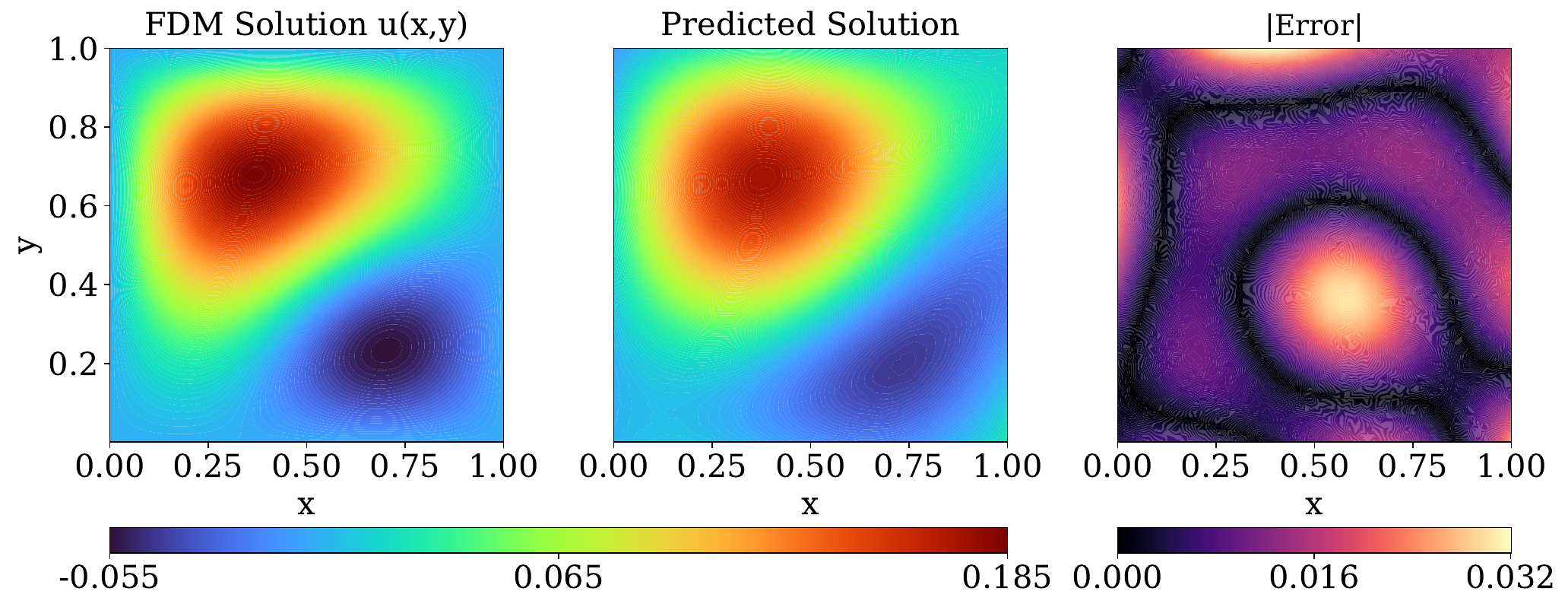}
        \caption{\DNew{DeepONet-based inverse operator: solution field}}
        \label{fig:helmholtz_sol_NP_DON}
    \end{subfigure}
    
    \caption{Helmholtz equation: Solution field comparison for non-parametric reconstruction. (a) Solution from MLP-reconstructed source with relative $L_2$ error of $0.0497$. (b) Solution from DeepONet-reconstructed source with relative $L_2$ error of $0.1523$. For each subfigure: left panel shows the reference FDM solution $u(x,y)$, middle panel shows the predicted solution, and right panel shows the pointwise absolute error. Multiple source configurations produce similar wave fields at sensor locations, demonstrating the non-uniqueness of the inverse problem.}
    \label{fig:helmholtz_solution_NP_comparison}
\end{figure}

Figures~\ref{fig:helmholtz_source_NP_comparison} and~\ref{fig:helmholtz_solution_NP_comparison} present the results. For the MLP approach, the reconstructed source achieves relative $L_2$ error of $0.34$. While the inverse operator correctly identifies the approximate region of maximum source intensity, it fails to capture the detailed spatial structure of the true Gaussian source (Figure~\ref{fig:helmholtz_source_NP_MLP}). The solution field reconstructed from this source achieves relative $L_2$ error of $0.050$ (Figure~\ref{fig:helmholtz_sol_NP_MLP}), indicating that multiple source configurations can produce similar wave fields at the sensor locations.

The DeepONet-based inverse operator produces smoother reconstructions but exhibits spatial diffusion, with source error of $0.60$ (Figure~\ref{fig:helmholtz_source_NP_DON}). The corresponding solution field error is $0.15$ (Figure~\ref{fig:helmholtz_sol_NP_DON}). The increased smoothness reflects the inherent bias of operator networks toward continuous functions but does not improve accuracy. Both architectures struggle to recover the true source structure despite different inductive biases.

These results confirm that the ill-posedness is a fundamental property of the Helmholtz inverse source problem, independent of the neural network architecture. Without explicit regularization through parametrization, smoothness penalties, or other priors, stable high-fidelity source reconstruction from sparse measurements is not achievable. The parametric approach remains the most appropriate choice for this problem class, combining mathematical well-posedness with physical interpretability.

\subsection{\D{Robustness of parameter identification} }

\D{To evaluate the practical applicability of the proposed frameworks, we examine their robustness under two realistic scenarios: (i) noisy observations and (ii) sparse data availability. We analyze the parameter identification framework across representative benchmark problems.}

\D{Table~\ref{tab:noise_robustness_det} presents parameter identification accuracy under varying noise levels. Gaussian noise $\eta \sim \mathcal{N}(0, \sigma^2)$ was added to sensor measurements during both training and testing, with $\sigma$ representing the percentage of signal standard deviation. The results demonstrate remarkable robustness enabled by physics-informed training. For 1D problems (Burgers' and Reaction-Diffusion), parameter estimation maintains absolute errors on the order of $10^{-3}$ even at 5\% noise, with relative solution errors below 12\%. The Heat equation shows moderate degradation with thermal conductivity errors around $5.5 \times 10^{-2}$ at 5\% noise, while the Helmholtz source reconstruction achieves spatial localization errors below $5 \times 10^{-2}$ across all noise levels, though amplitude estimation exhibits larger errors due to the parameter's inherent ambiguity. Critically, solution reconstruction quality remains strong across all problems (relative $L_2$ errors $< 0.6$ for Heat and $< 0.3$ for Helmholtz), confirming that the physics-informed framework effectively regularizes the inverse problem through embedded governing equations, enabling stable parameter estimation from noisy sparse measurements.}

\D{Table~\ref{tab:sparsity_robustness_det} demonstrates the framework's robustness to sparse sensor data. For Reaction-Diffusion and Burgers' equations, both parameter identification accuracy and solution reconstruction improve monotonically with increasing sensor density, with relative $L_2$ errors decreasing from 0.088 to 0.050 and 0.048 to 0.029, respectively. The Heat equation exhibits strong robustness, with parameter error improving by over 50\% (from $6.6 \times 10^{-2}$ to $3.0 \times 10^{-2}$) while maintaining consistent solution accuracy (relative $L_2$ error $\approx$ 0.57-0.71). The Helmholtz problem shows an interesting pattern: spatial parameters ($x_0$, $y_0$) remain accurately identified even at the sparsest configuration (64 sensors), while the amplitude parameter $A$ exhibits sensitivity to sensor density, with error increasing from $2.0 \times 10^{-2}$ to $2.1$ at very sparse configurations. This reflects the higher sensitivity of wave field reconstruction to source amplitude under limited measurements. Importantly, even at the sparsest sensor configurations, all problems achieve parameter errors suitable for reasonable solution reconstruction, demonstrating the effectiveness of physics-informed training in data-limited scenarios.}

\begin{table}[h!]
\centering
\footnotesize
\caption{ \D{Parameter identification accuracy i.e. average absolute test error in the test cases under varying noise levels. Sensor measurements were corrupted with Gaussian noise $\eta \sim \mathcal{N}(0, \sigma^2)$ where $\sigma$ represents the percentage of signal standard deviation. All experiments use 300 sensors.}}
\label{tab:noise_robustness_det}
\begin{tabular}{lccccc}
\hline
\multirow{2}{*}{Problem} & \multirow{2}{*}{Parameter} & \multicolumn{4}{c}{Noise Level (\%)} \\
\cmidrule(lr){3-6}
& & 0\% & 1\% & 3\% & 5\% \\
\hline
\multirow{2}{*}{Reaction-Diffusion} 
& $|D_{\text{true}} - D_{\text{pred}}|$ & $8.0 \times 10^{-3}$ & $8.0 \times 10^{-3}$ & $9.0 \times 10^{-3}$ & $9.0 \times 10^{-3}$ \\
& Rel. $L_2$ error ($u$) & 0.050 & 0.057 & 0.111  & 0.112 \\
\hline
\multirow{2}{*}{Burgers'} 
& $|\nu_{\text{true}} - \nu_{\text{pred}}|$ & $2.0 \times 10^{-3}$ & $2.0 \times 10^{-3}$  & $3.0 \times 10^{-3}$ & $4.0 \times 10^{-3}$ \\
& Rel. $L_2$ error ($u$) & 0.029 & 0.037 & 0.044 & 0.049 \\
\hline
\multirow{2}{*}{Heat} 
& $|K_{\text{true}} - K_{\text{pred}}|$ & $3.0 \times 10^{-2}$ & $5.6 \times 10^{-2}$ & $6.6 \times 10^{-2}$  & $5.5 \times 10^{-2}$  \\
& Rel. $L_2$ error ($u$) & 0.570 & 0.512 & 0.534 & 0.581 \\
\hline
\multirow{4}{*}{Helmholtz} 
& $|A_{\text{true}} - A_{\text{pred}}|$ & $2.8 \times 10^{-1}$ & $2.0 \times 10^{0}$ & $2.5 \times 10^{0}$ & $2.1 \times 10^{0}$  \\

& $|x_{0,\text{true}} - x_{0,\text{pred}}|$ & $3.4 \times 10^{-2}$ & $3.3 \times 10^{-2}$ &  $2.6 \times 10^{-2}$ & $3.5 \times 10^{-2}$ \\

& $|y_{0,\text{true}} - y_{0,\text{pred}}|$ & $4.4 \times 10^{-2}$ & $3.7 \times 10^{-2}$  &  $4.5 \times 10^{-2}$ & $3.9\times 10^{-2}$ \\
& Rel. $L_2$ error ($u$) & 0.099 & 0.266 & 0.089  & 0.281 \\
\hline
\end{tabular}
\end{table}

\begin{table}[h!]
\centering
\footnotesize
\setlength{\tabcolsep}{4pt}
\caption{\D{Robustness of deterministic parameter identification to data sparsity. Results show mean absolute parameter error and relative $L_2$ error for the state variable $u$ over test samples under varying sensor counts. All measurements are noise-free.}}
\label{tab:sparsity_robustness_det}
\begin{tabular}{@{}llcccc@{}}
\toprule
\textbf{Problem} & \textbf{Metric} & \textbf{Config 1} & \textbf{Config 2} & \textbf{Config 3} & \textbf{Config 4} \\
\midrule
\multirow{2}{*}{Reaction-Diffusion} 
& $|D_{\text{true}} - D_{\text{pred}}|$ & \textbf{50}: $5.0 \times 10^{-3}$ &  \textbf{100}: $6.0 \times 10^{-3}$ & \textbf{200}: $7 \times 10^{-3}$ & \textbf{300}: $8.0 \times 10^{-3}$ \\
 & Rel. $L_2$ error ($u$) & 0.088 & 0.077 & 0.073 & 0.050 \\
\midrule
\multirow{2}{*}{Burgers'} 
& $|\nu_{\text{true}} - \nu_{\text{pred}}|$ & \textbf{50}: $3.0 \times 10^{-3}$ & \textbf{100}: $2.0 \times 10^{-3}$ & \textbf{200}: $2.0 \times 10^{-3}$ & \textbf{300}: $2.0 \times 10^{-3}$ \\
& Rel. $L_2$ error ($u$)& 0.048 &  0.034 & 0.035 & 0.029 \\
\midrule
\multirow{2}{*}{Heat} 
& $|K_{\text{true}} - K_{\text{pred}}|$ & \textbf{165}:  $6.6 \times 10^{-2}$ & \textbf{231}: $7.0 \times 10^{-2}$ & \textbf{352}: $7.0 \times 10^{-2}$ &\textbf{440}: $3.0 \times 10^{-2}$ \\
& Rel. $L_2$ error ($u$) & 0.71 & 0.59 & 0.56 & 0.570 \\
\midrule
\multirow{4}{*}{Helmholtz} 
& $|A_{\text{true}} - A_{\text{pred}}|$ & \textbf{64}: $2.0 \times 10^{-2}$ & \textbf{144}: $1.3 \times 10^{-2}$ & \textbf{256}: $1.48 \times 10^{-1}$& \textbf{1024}: $2.1 \times 10^{0}$ \\
& $|x_0^{\text{true}} - x_0^{\text{pred}}|$ & $1.8 \times 10^{-2}$ & $1.8 \times 10^{-2}$ & $1.9 \times 10^{-2}$ & $3.5 \times 10^{-2}$ \\
& $|y_0^{\text{true}} - y_0^{\text{pred}}|$ & $2.9 \times 10^{-2}$ & $2.1 \times 10^{-2}$ & $2.2 \times 10^{-2}$ & $3.9 \times 10^{-2}$ \\
& Rel. $L_2$ error ($u$) & 0.113  & 0.113 & 0.121 & 0.281 \\
\bottomrule
\end{tabular}
\end{table}

\section{\D{Discussions} and Conclusion}
\label{sec:discussions&conclusion}

\subsection{\D{Computational efficiency and scalability analysis}} \label{subsec:comp_efficiency}

\D{While the benchmark problems in this study are low-dimensional and individual forward simulations can be solved efficiently using traditional numerical methods (requiring $0.05$ to $0.12$ seconds depending on discretization), the proposed operator learning framework offers substantial computational advantages for inverse problems and parameter estimation tasks. Traditional approaches to parameter identification require iterative optimization, where each parameter estimation scenario involves multiple forward solver evaluations, typically $500$ or more iterations for convergence, along with gradient computation via finite differences or adjoint methods. For our benchmark problems, this translates to approximately $25$ to $60$ seconds per parameter estimation scenario, assuming a minimum of $500$ iterations. When parameter estimation must be performed across multiple scenarios $N$, the total computational cost scales as $\mathcal{O}(N \times n_{\text{iter}} \times t_{\text{forward}})$, making it prohibitively expensive for large-scale parameter space exploration. For instance, estimating parameters for $1,500$ test scenarios using traditional iterative optimization would require an estimated $10.4 $ to $25$ hours of computation time, and this assumes relatively fast convergence without accounting for difficult optimization landscapes that may require significantly more iterations.}

\begin{table}[!htb]
\centering
\footnotesize
\caption{\D{Computational cost: training runtime and inference time for all models.}}
\label{tab:comp_cost}
\setlength{\tabcolsep}{5pt}
\renewcommand{\arraystretch}{1.1}
\begin{tabular}{lcccc}
\toprule
\multirow{2}{*}{Problem} &
DHPO &
INV-DON (Det.) &
INV-DON (Prob.) &
Inference \\
& Train [hrs] & Train [s] & Train [s] & [ms / query] \\
\midrule
Reaction–Diffusion 1D     & 3  & 371.32 & 263.68 & 0.104 \\
Burgers' 1D                 & 70 & 446.30 & 450.08 & 0.101 \\
Heat 2D (L-shaped)         & 19 & 144.14 & 224.46 &  0.217\\
Helmholtz 2D (Source ID)   & --- & 185.30 & 233.57 & 0.188 \\
\bottomrule
\end{tabular}
\end{table}

\D{In contrast, our operator learning framework requires a one-time training phase ranging from $2.4$ to $7.5$ minutes depending on the problem complexity, after which parameter estimation for each new scenario requires only rapid inference (approximately $0.1$ to $0.24$ miliseconds per query) without any iterative optimization. As demonstrated in Table \ref{tab:comp_cost}, it should be noted that DHPO training times are reported in hours, whereas INV-DON training times are reported in seconds, due to differences in implementation (TensorFlow 2 for DHPO versus JAX for INV-DON), hyperparameter choices (e.g., batch size, number of training samples, and epochs), and GPU hardware (V100 for DHPO versus A100 for INV-DON).These implementation differences do not affect the inference performance, which remains consistently fast across all methods. Our framework achieves parameter estimation for $1,500$ test scenarios in approximately $7.5$ minutes total time for the Burgers' equation and $6.2$ minutes for the reaction diffusion system, representing speedups of over $80\times$ to $240\times$ compared to traditional iterative methods. For the 2D heat equation with $2,000$ test scenarios, the total time is approximately $3.7$ minutes compared to an estimated $13.9$ to $33.3$ hours using traditional optimization, yielding speedups of $225\times$ to $540\times$. This advantage becomes particularly pronounced for applications requiring real-time parameter estimation, extensive parameter space exploration for uncertainty quantification, or repeated inverse problems with different observation sets. We acknowledge that for single forward problem evaluations, traditional numerical methods are computationally efficient and well-established. However, the inverse problems addressed in this work, specifically parameter identification from sparse observations, require fundamentally different computational considerations, as traditional optimization based methods must repeatedly evaluate the forward solver whereas our operator learning framework learns the inverse mapping directly, enabling rapid parameter estimation across multiple scenarios without iterative optimization.}

\subsection{Conclusion}
In this work, we presented two complementary operator-learning frameworks for solving inverse problems in computational science and engineering. The Deep Hidden Physics Operator (DHPO) extends hidden-physics modeling into the operator-learning paradigm, enabling the discovery of unknown PDE terms across diverse equation families. In parallel, the proposed parameter identification framework leverages pretrained DeepONets to reconstruct solution fields from sparse sensor data and infer governing parameters in a physics-consistent manner. Together, these approaches address two fundamental challenges, the hidden-physics discovery and parameter estimation within a unified data-efficient framework.

Our results on canonical benchmarks, including the reaction-diffusion system, Burgers’ equation, and the 2D heat equation, demonstrate that the methods achieve robust performance even under sparse and noisy observations. Specifically, DHPO successfully recovers hidden physics with relative solution errors on the order of $10^{-2}$, while the parameter identification framework achieves absolute parameter errors on the order of $10^{-3}$. Compared to existing methods such as PINNs and DHPM, the proposed frameworks offer improved generalization across PDE families, eliminate the need for repeated retraining, and enhance data efficiency.

These findings highlight the promise of combining neural operator learning with physics-informed constraints for advancing data-driven discovery in the sciences. Future directions include extending the frameworks to three-dimensional problems, incorporating uncertainty quantification for rigorous error bounds, and exploring hybrid architectures that integrate other neural operators (e.g., Fourier or wavelet-based operators) for enhanced scalability. More broadly, the operator-learning perspective opens new opportunities for robust, generalizable, and interpretable solvers in computational mechanics and beyond.

\subsection{\D{Limitations and future directions}}

\D{While our approach demonstrates robust performance across various inverse problems, the current framework assumes fixed sensor locations during both training and inference. This represents a practical limitation, as many real-world inverse problems involve variable numbers and spatial distributions of sensors depending on measurement constraints or sensor availability. The fixed-location assumption is inherited from the DeepONet architecture, where the branch network processes sensor measurements at predetermined locations. Recent work on set-based operator networks (SetONet) \cite{tretiakov2025setonet} addresses this limitation by incorporating permutation-invariant set encoders that can handle arbitrary sensor configurations. Extending our Neural Inverse Operator framework to accommodate variable sensor placements represents a promising direction for future research, potentially combining our operator-to-function learning approach with SetONet's flexible sensor encoding mechanisms. Such an extension would enhance the practical applicability of the method to scenarios with heterogeneous sensor networks or adaptive sensing strategies.}

\section*{Acknowledgments}

VK and BP would like to acknowledge the computing support provided by The Deep Learning and GPU computing Platform (BD/DPA-BDP1) at Robert Bosch, GmbH, Renningen, Germany. DRS and SG  would like to acknowledge computing support provided by the Advanced Research Computing at Hopkins (ARCH) core facility at Johns Hopkins University and the Rockfish cluster. ARCH core facility (\url{rockfish.jhu.edu}) is supported by the National Science Foundation (NSF) grant number OAC1920103. DRS and SG would like to acknowledge Dr. Krishna Kumar, Dr. Xingjian Li, and Dr. Hassan Iqbal for the insightful discussions
during the journal club meetings. DRS and SG were partially supported by JHU Discovery award 2024 and NSF grant number 2436738.

\section*{Author contributions}
\noindent Conceptualization: VK, DRS, BP, SG  \\
Investigation: VK, DRS, BP, SG \\
Visualization: VK, DRS, BP, SG  \\
Supervision: BP, SG \\
Writing - original draft: VK, DRS \\
Writing - review \& editing: VK, DRS, BP, SG 

\section*{Data and code availability}
\noindent The code and data that support the findings of this study will be available from the corresponding author on reasonable request upon publication.

\section*{Competing interests}
\noindent The authors declare no competing interest

\bibliography{references}
\setcounter{figure}{0}
\setcounter{table}{0}
\appendix
\section{Hyperparameter tuning}

In the deep learning community, Weights \& Biases (WandB) is a popular online platform that provides tools for monitoring and controlling hyperparameters during model training runs. Sweeps, a crucial component of WandB, allow for the systematic tuning of model hyperparameters. The Sweeps configuration dictionary contains the predefined hyperparameter search space. Depending on this setup, a sweep could include several runs, with grid search, random search, or Bayesian optimization used for hyperparameter sampling. Sarkar {\it{et al.}} \cite{sarkar4993297adaptive} showcases the use of WandB for hyperparameter tuning on PINNs training. We have utilized the WandB tool for hyperparameter tuning of our framework for the parameter identification problem of the Burgers' equation. The report of the hyperparameter tuning can be found at the link {\small\url{https://api.wandb.ai/links/droysar1-johns-hopkins-university/pzs4hfup}}.

\begin{table}[!!htb]
\centering
\caption{Architecture and training parameters for all the examples presented in this work.}
\label{tab:hyperparameter}
\setlength{\tabcolsep}{5pt}
\tiny
\renewcommand{\arraystretch}{1.1}
\begin{tabular}{@{}lccccccc@{}}
\toprule
\textbf{Equation} & \textbf{\begin{tabular}[c]{@{}c@{}}Branch\\ Network\end{tabular}} & \textbf{\begin{tabular}[c]{@{}c@{}}Trunk\\ Network\end{tabular}} & \textbf{\begin{tabular}[c]{@{}c@{}}MLP or\\ FNN\end{tabular}} & \textbf{\begin{tabular}[c]{@{}c@{}}Activation\\ Function\end{tabular}} & \textbf{\begin{tabular}[c]{@{}c@{}}Training\\ Iterations\end{tabular}} & \textbf{\begin{tabular}[c]{@{}c@{}}Learning Rate\\ (Initial / Decay)\end{tabular}} & \textbf{\begin{tabular}[c]{@{}c@{}}Batch\\ Size\end{tabular}} \\
\midrule
\textbf{Hidden physics} &  &  &  &  &  & \\
\hdashline
\addlinespace
Reaction-Diffusion          & $[101,128,128,128,50]$ & $[2,128,128,128,50]$ & $[3,128,128,128,1]$ & \begin{tabular}[c]{@{}c@{}c@{}} Branch $-$ ReLU\\ Trunk $-$ Tanh \\ MLP $-$ Tanh \end{tabular} & 10k & $1e-4$ & 10 \\
Burgers'          & \begin{tabular}[c]{@{}c@{}}$[101,128,128,$\\ $128,128,50]$ \end{tabular}
 & \begin{tabular}[c]{@{}c@{}}
 $[2,128,128,$ \\$128,128,50]$ \end{tabular}& $[3,256,256,256,1]$ & \begin{tabular}[c]{@{}c@{}c@{}} Branch $-$ ReLU\\ Trunk $-$ Tanh \\ MLP $-$ Tanh \end{tabular}  & 10k & $1e-4$ & 1 \\
 Heat Equation (2D) & \begin{tabular}[c]{@{}c@{}} CNN: $33 \times 33$, \\ Channels $[1,40,60]$, \\ Kernel $3\times 3$, stride $2$, \\ Maxpool $2 \times 2$ \\ FCN: $[240,64,64,64]$ \end{tabular} 
& $[3,128,128,128,64]$ 
& $[5,128,128,128,1]$ 
& \begin{tabular}[c]{@{}c@{}} Branch $-$ ReLU, Linear \\ Trunk $-$ Tanh, Linear \\ MLP $-$ Tanh, Linear \end{tabular} 
& 2000 
& $1e-4$ 
& 1 \\
\midrule
\multicolumn{2}{@{}l}{\textbf{\begin{tabular}[c]{@{}l@{}}System parameter identification\end{tabular}}} &  &  &  &  & & \\
\hdashline
\addlinespace
Reaction-Diffusion & $[300, 64, 64, 64, 100]$ & $[2, 64, 64, 64, 100]$ & $[300, 64, 64, 64, 1]$ & Tanh & 80k & \begin{tabular}[c]{@{}c@{}}$1e-3$\\ exponential, \\ rate = 0.9\end{tabular}  & $3500$  \\ \\
Burgers'      & $[300, 64, 64, 64, 100]$ & $[2, 64, 64, 64, 100]$ & $[300, 64, 64, 64, 1]$& Tanh & 80k & \begin{tabular}[c]{@{}c@{}}$1e-3$\\ exponential, \\ rate = 0.9\end{tabular}  & $2500$  \\
Heat      & $[440, 64, 64, 64, 100]$ & $[3, 64, 64, 64, 100]$ & $[440, 32, 16, 8, 1]$& \begin{tabular}[c]{@{}c@{}}Branch-Tanh\\ Trunk-Tanh \\MLP - GeLU \end{tabular} & 80k & \begin{tabular}[c]{@{}c@{}}$1e-3$\\ exponential, \\ rate = 0.9\end{tabular}  & $75$  \\
Helmholtz      & $[1024, 32, 32, 32, 100]$ & $[2, 32, 32, 32, 100]$ & $[1024, 32, 32, 32, 3]$& \begin{tabular}[c]{@{}c@{}}Branch-Tanh\\ Trunk-Tanh \\MLP - GeLU \end{tabular} & 80k & \begin{tabular}[c]{@{}c@{}}$1e-3$\\ exponential, \\ rate = 0.9\end{tabular}  & $75$  \\
\bottomrule
\end{tabular}
\end{table}

\begin{figure}[!htb]
\subfloat[]{\label{}\includegraphics[width=.45\linewidth]{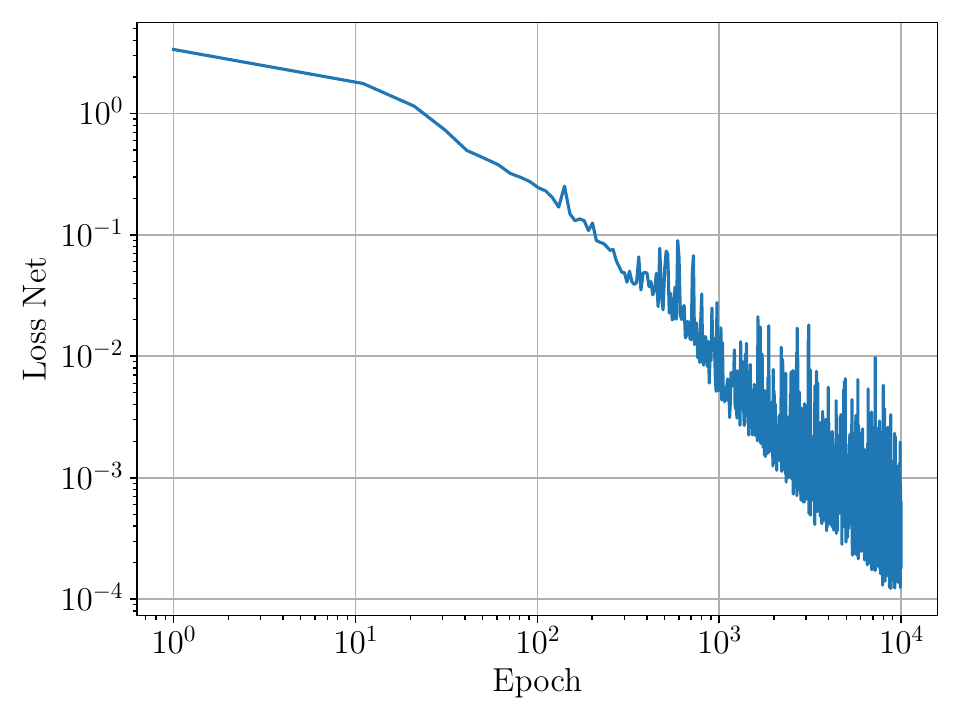}}\hfill
\subfloat[]{\label{}\includegraphics[width=.45\linewidth]{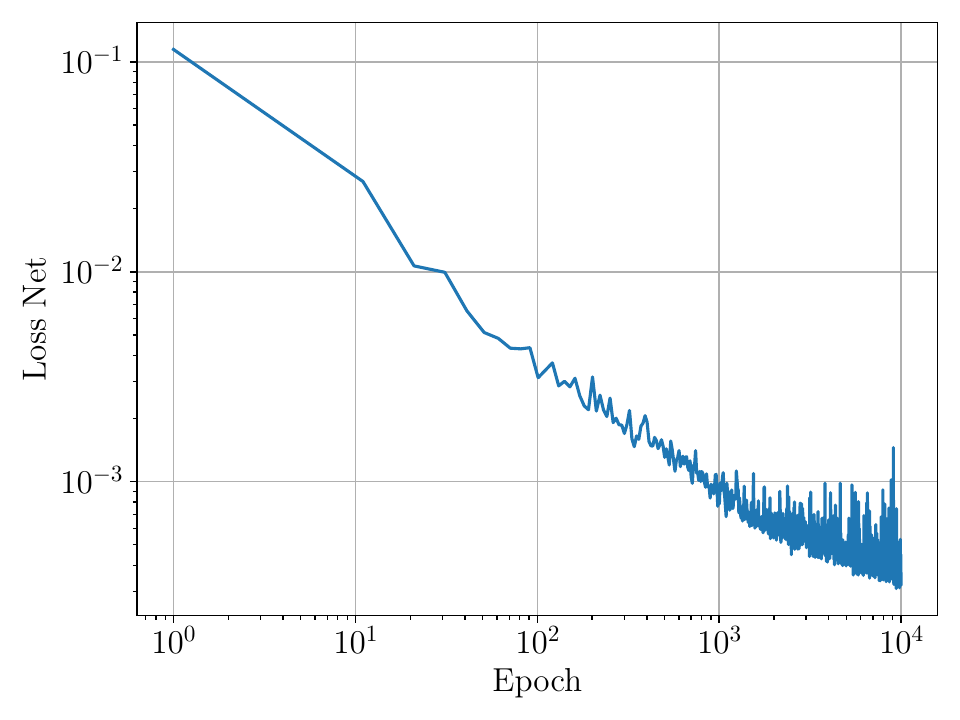}}\par 
\subfloat[]{\label{}\includegraphics[width=.45\linewidth]{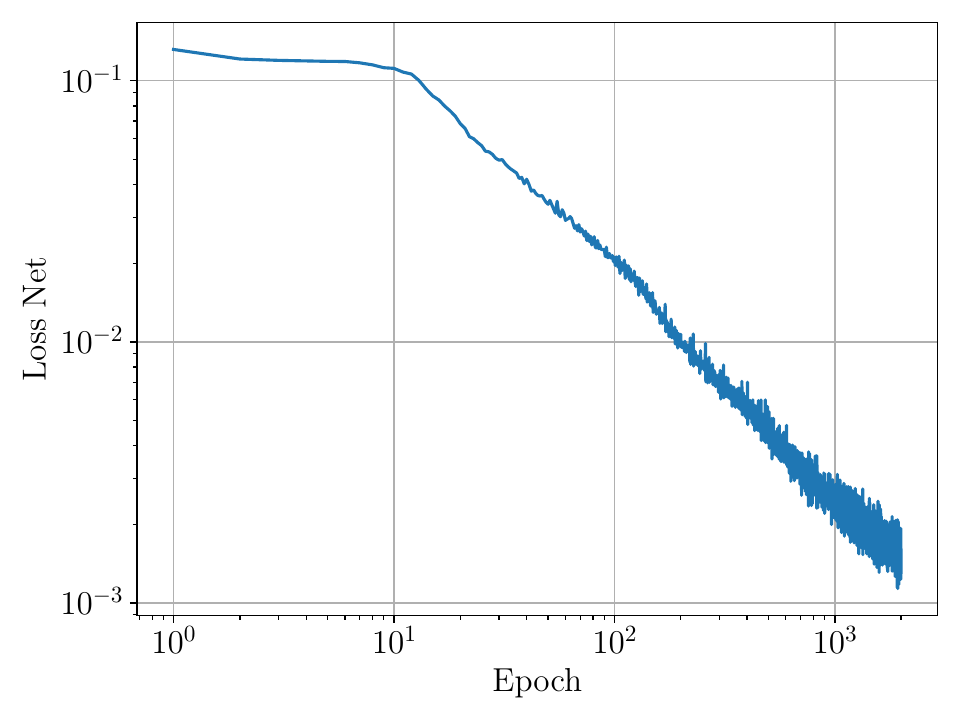}}
\caption{Loss function over the training process of optimal model: (a) Reaction diffusion equation with $N_{train} = 500, N_d = 500.$ (b) Burgers' equation with $N_{train} = 1000, N_d = 500.$ (c)
Heat equation with $N_{train} = 500, N_d = 550.$}
\label{fig:loss_plot}
\end{figure}

\begin{figure}[!htb]
\centering
\includegraphics[width=\linewidth, height=0.8\textheight, keepaspectratio]{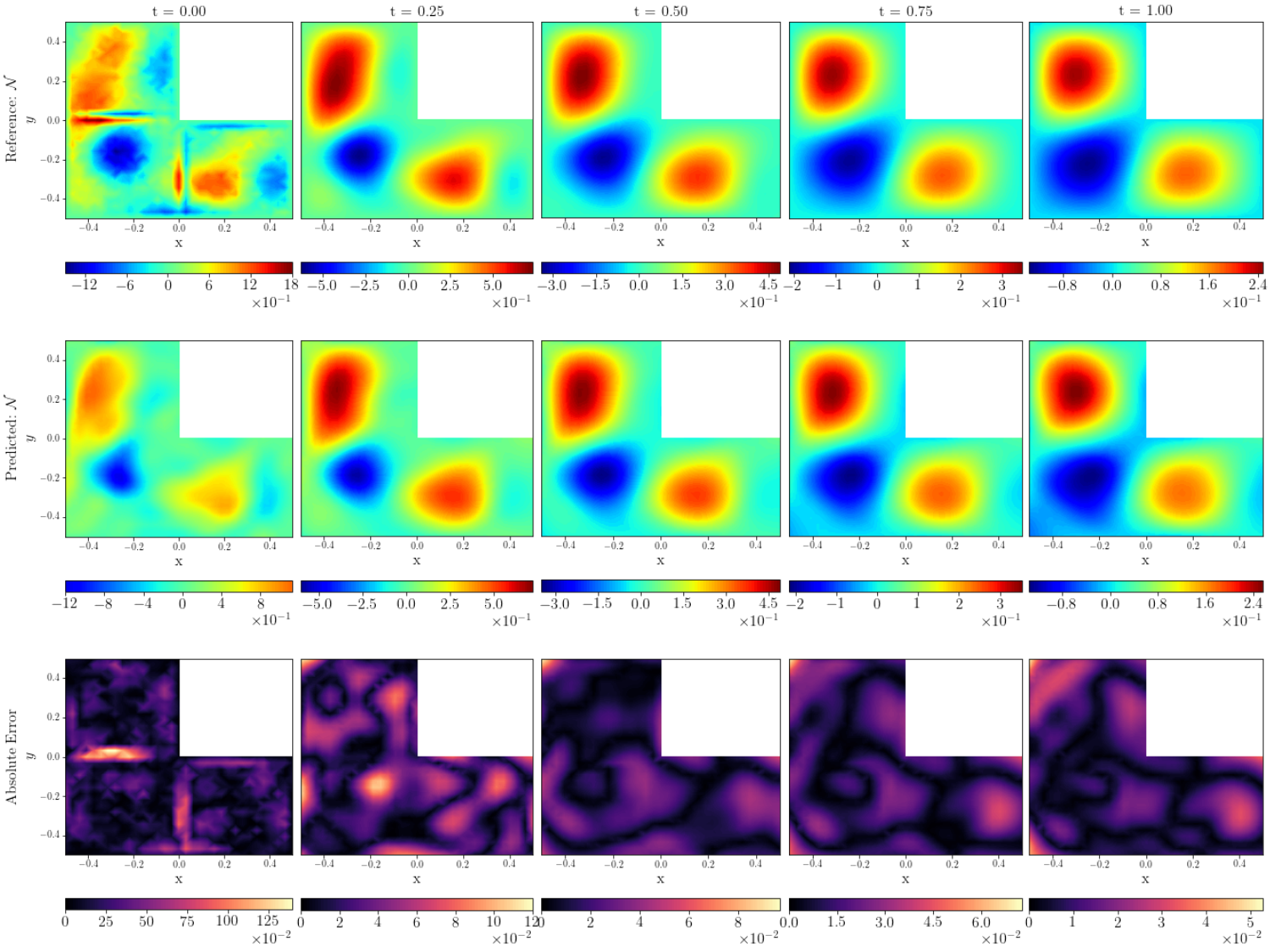}
\caption{Sample 1: hidden physics solution comparison, mean absolute error = 0.03177}
\label{fig:dhpm_RHSeqn_pred_sample1}
\end{figure}

\begin{figure}[!htb]
\centering
\includegraphics[width=\linewidth, height=0.8\textheight, keepaspectratio]{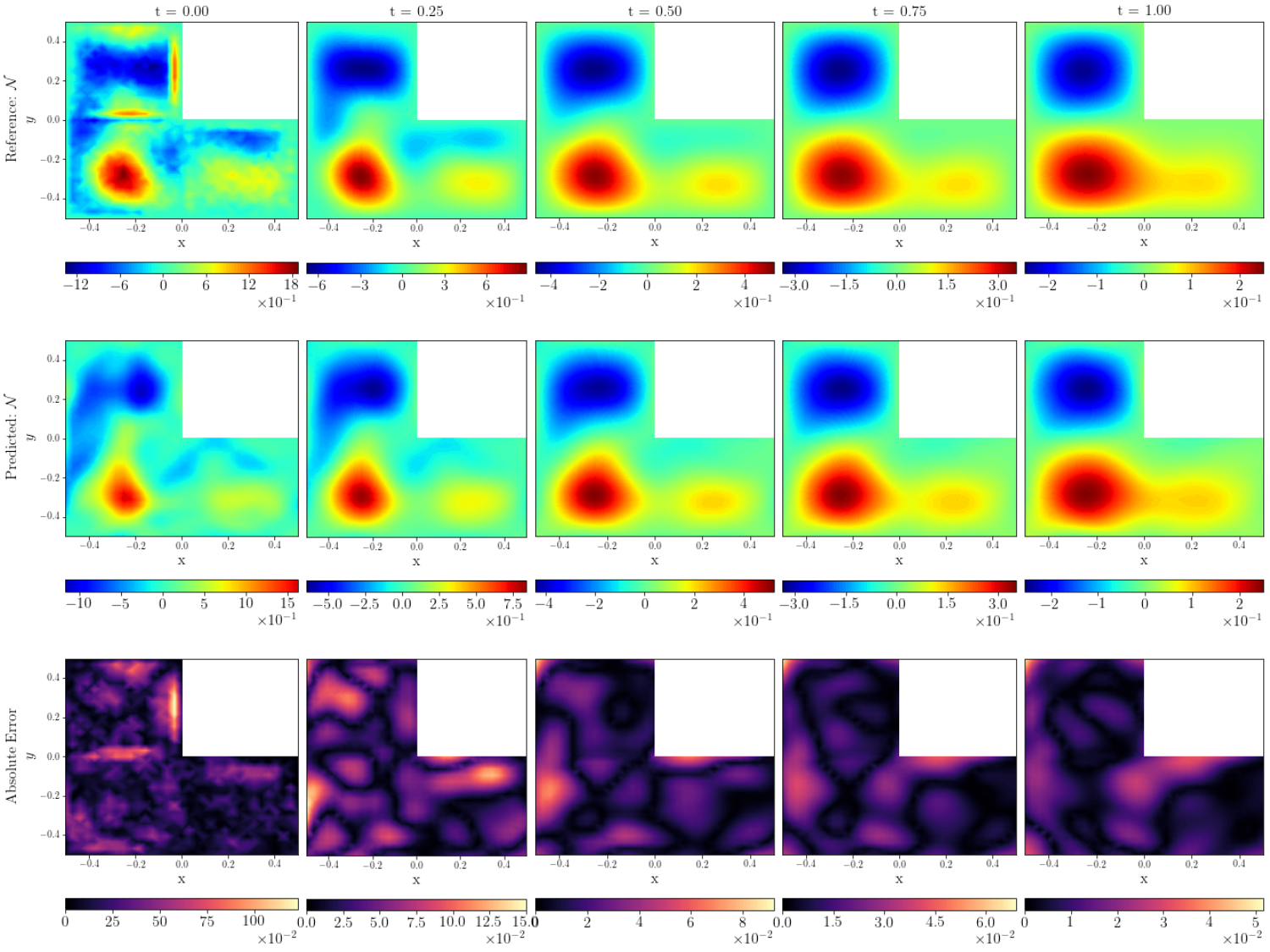}
\caption{Sample 2: hidden physics solution comparison, mean absolute error = 0.03415}
\label{fig:dhpm_RHSeqn_pred_sample2}
\end{figure}

\begin{figure}[!htb]
\centering
\vspace{-20pt}
\begin{subfigure}{\textwidth}
    \centering
    \includegraphics[width=0.9\linewidth]{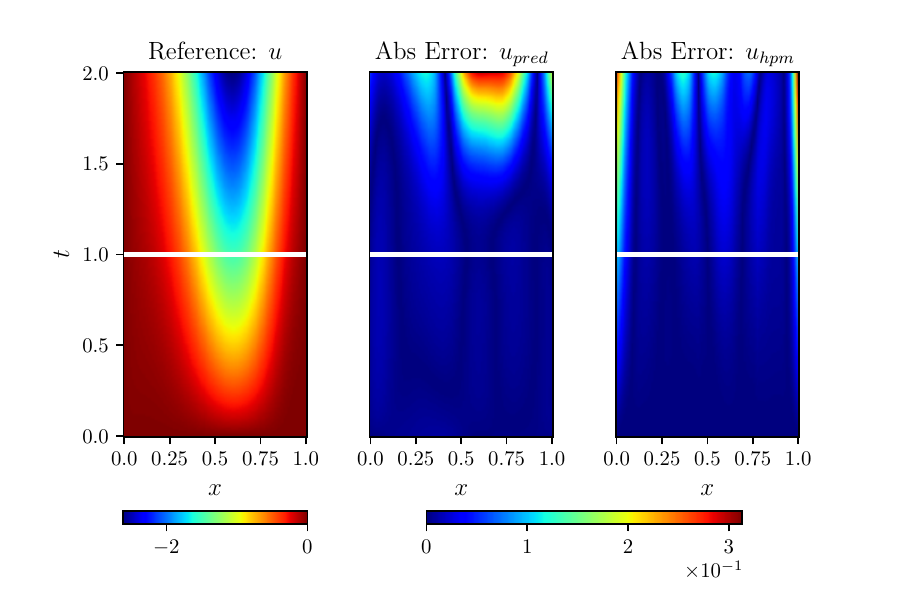} 
    \caption{Sample 1: solution field accuracy comparison, mean absolute error $u_{pred}(t\leq 1)  = 0.00507,  u_{pred}(t>1) = 0.04674, \\  u_{hpm}(t\leq 1)= 0.00627, u_{hpm}(t>1)  = 0.03299$ }
\end{subfigure}
\begin{subfigure}{\textwidth}
    \centering
    \includegraphics[width=0.9\linewidth]{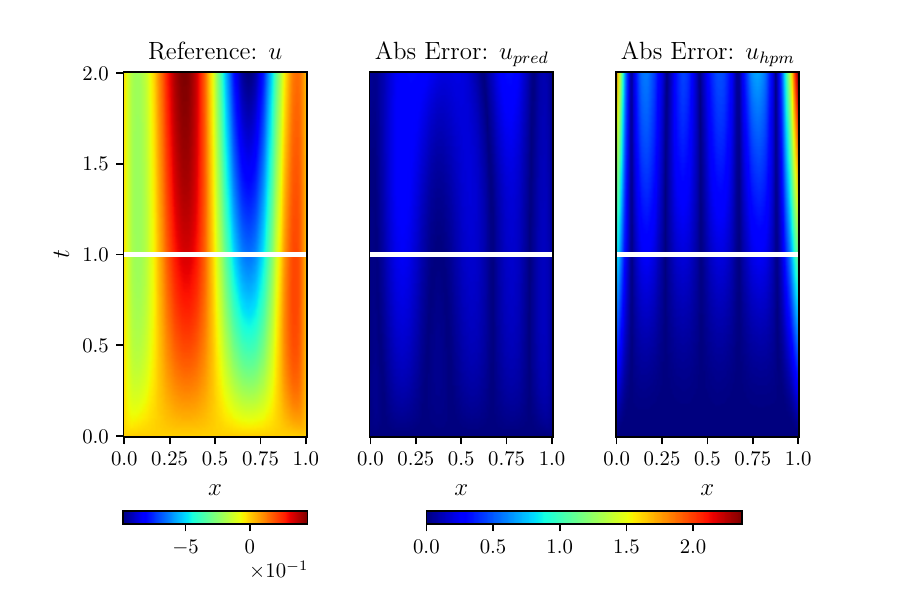}  
    \caption{Sample 2: solution field accuracy comparison, mean absolute error $u_{pred}(t\leq 1)  = 0.06620,  u_{pred}(t>1) = 0.13954, \\  u_{hpm}(t\leq 1) = 0.08864, u_{hpm}(t>1)  =  0.35198$ }
\end{subfigure}
\caption{Reaction diffusion equation: Comparison of DHPO solution \& HPM integrated solution for two representative test samples with input function from GRF function space $l=0.2$ (nonzero at boundaries). It should be noted that training data is collected from t = 0 to  1., extrapolation time region separated by the white horizontal lines.}
\label{fig:RD_grf_extp_samples}
\end{figure}

\begin{figure}[!htb]
\centering
\vspace{-20pt}
\begin{subfigure}{\textwidth}
    \centering
    \includegraphics[width=0.9\linewidth]{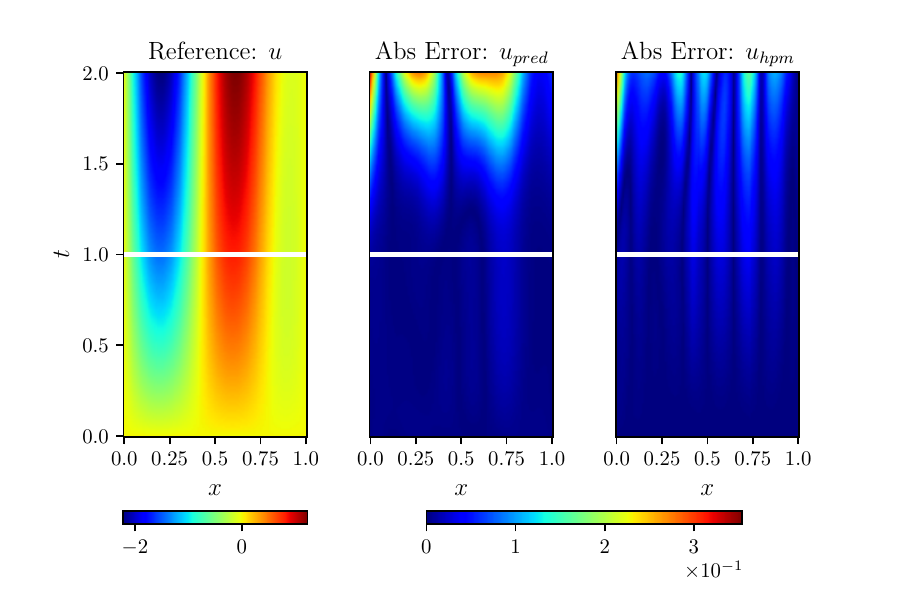} 
    \caption{Sample 1: solution field accuracy comparison, mean absolute error $u_{pred}(t\leq 1) = 0.00458, u_{pred}(t>1) = 0.05970, \\  u_{hpm}(t\leq 1) = 0.0059, u_{hpm}(t>1) =  0.03579$ }
\end{subfigure}
\begin{subfigure}{\textwidth}
    \centering
    \includegraphics[width=0.9\linewidth]{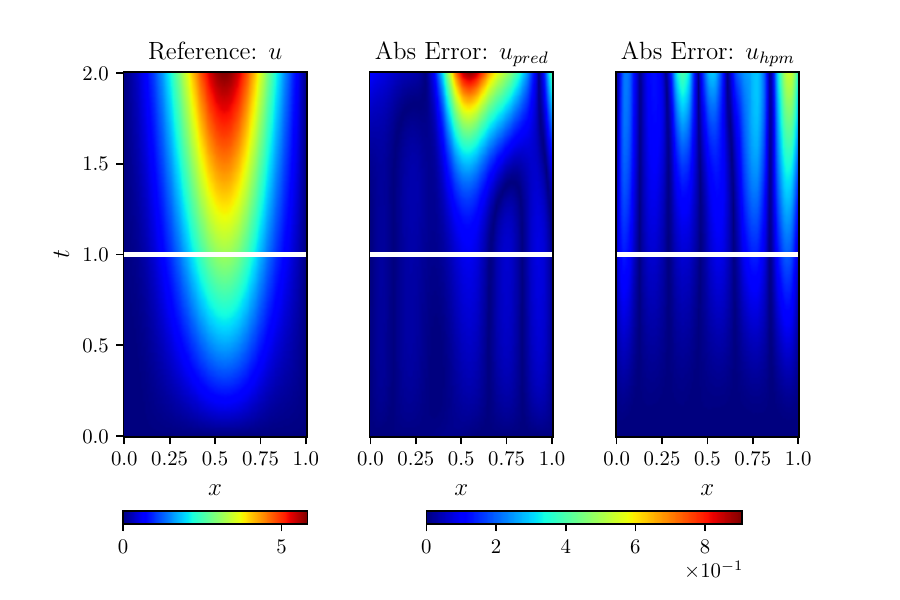}  
    \caption{Sample 2: solution field accuracy comparison, mean absolute error $u_{pred}(t\leq 1) =  0.02584, u_{pred}(t>1) = 0.12762, \\  u_{hpm}(t\leq 1) = 0.02967, u_{hpm}(t>1) =  0.13450$ }
\end{subfigure}
\caption{Reaction diffusion equation: 
Comparison of DHPO solution \& HPM integrated solution for two representative test samples with input function from Modified GRF function space $l=0.2$ (zero at boundaries). It should be noted that training data is collected from t = 0 to  1., extrapolation time region separated by the white horizontal lines.}
\label{fig:RD_modified_grf_extp_samples}
\end{figure}

\begin{figure}[!htb]
\centering
\begin{subfigure}{\textwidth}
    \centering
    \includegraphics[width=0.75\linewidth]{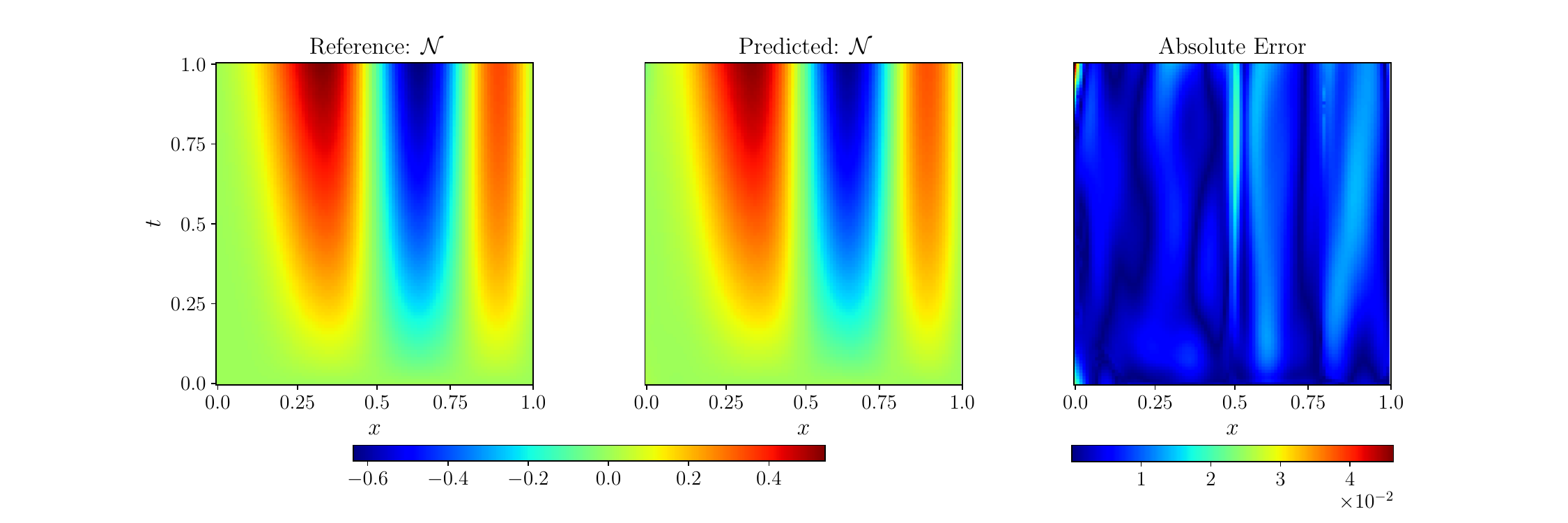}  
    \caption{Sample 1: hidden physics solution comparison, mean absolute error = 0.00567.}
\end{subfigure}
\begin{subfigure}{\textwidth}
    \centering
    \includegraphics[width=0.8\linewidth]{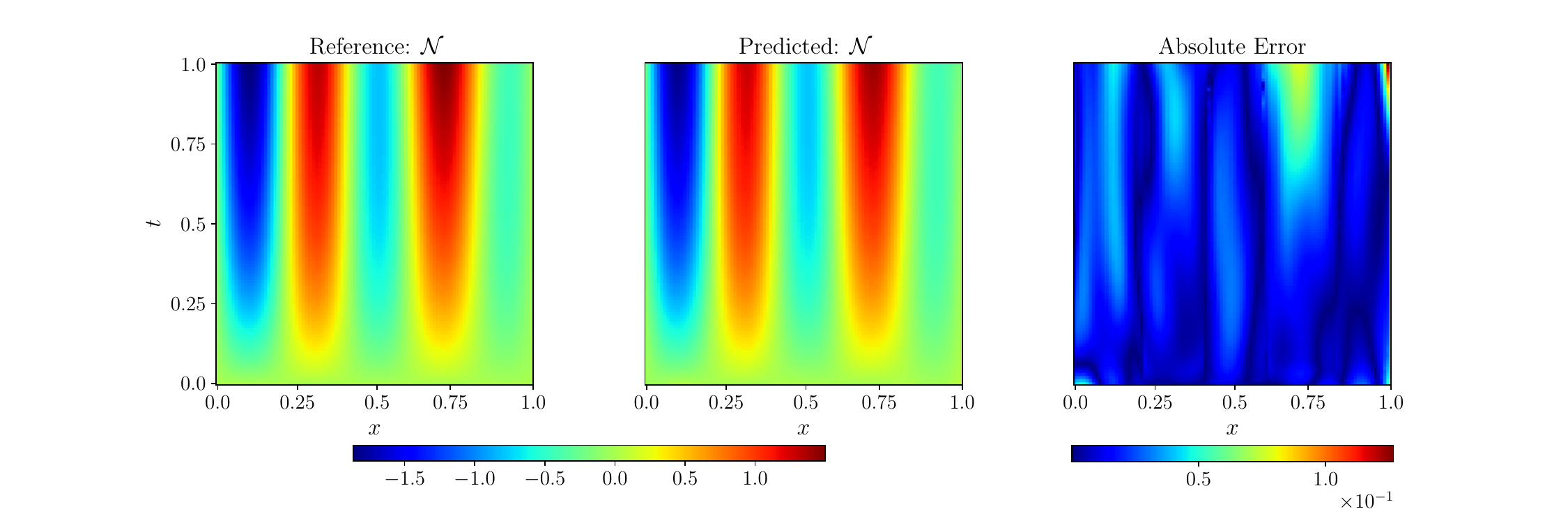}  
    \caption{Sample 2: hidden physics solution comparison, mean absolute error = 0.01787.}
\end{subfigure}
\caption{Reaction diffusion equation: Comparison of reference and predicted PDE terms $\mathcal{N}$ with input function from sine function space. It should be noted that, the Reference $\mathcal{N}$ is computed using finite difference with the gradients being computed using second order scheme.}
\label{fig:RD_sine_hid_phy_samples}
\end{figure}

\begin{figure}[!htb]
\centering
\begin{subfigure}{\textwidth}
    \centering
    \includegraphics[width=0.8\linewidth]{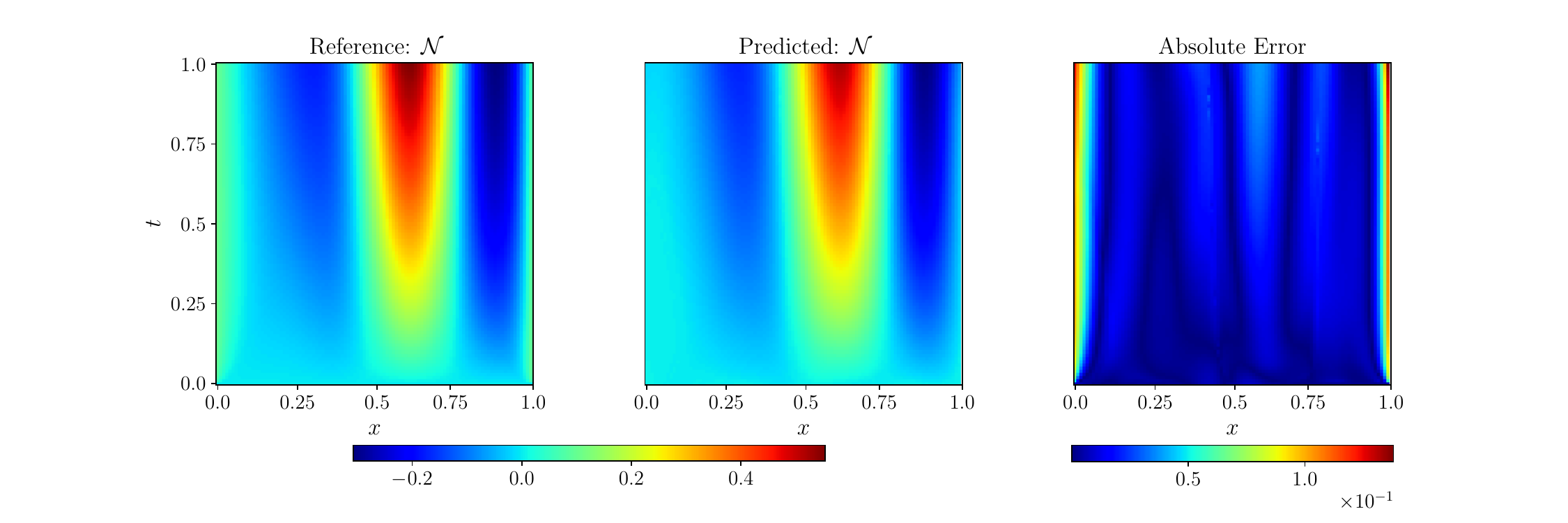}  
    \caption{Sample 1: hidden physics solution comparison, mean absolute error = 0.01587.}
\end{subfigure}
\begin{subfigure}{\textwidth}
    \centering
    \includegraphics[width=0.8\linewidth]{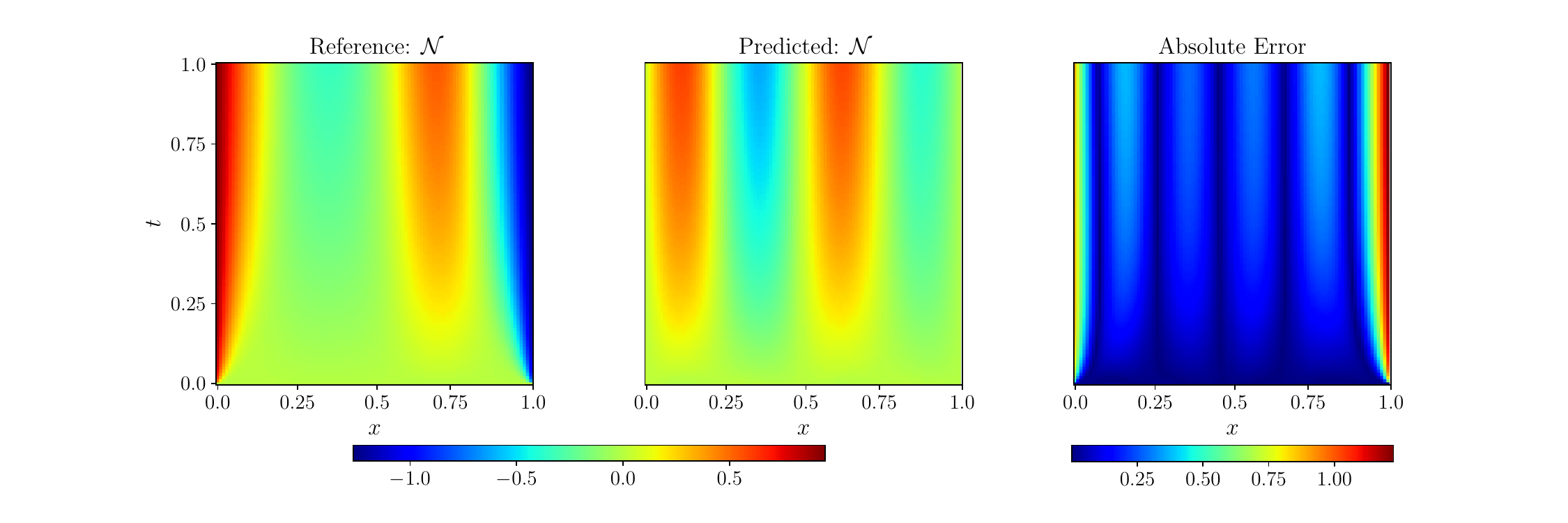}  
    \caption{Sample 2: hidden physics solution comparison, mean absolute error = 0.21618.}
\end{subfigure}
\caption{Reaction diffusion equation: Comparison of reference and predicted PDE terms with input function from GRF function.}
\label{fig:RD_grf_hid_phy_samples}
\end{figure}

\begin{figure}[!htb]
\centering
\begin{subfigure}{\textwidth}
    \centering
    \includegraphics[width=0.8\linewidth]{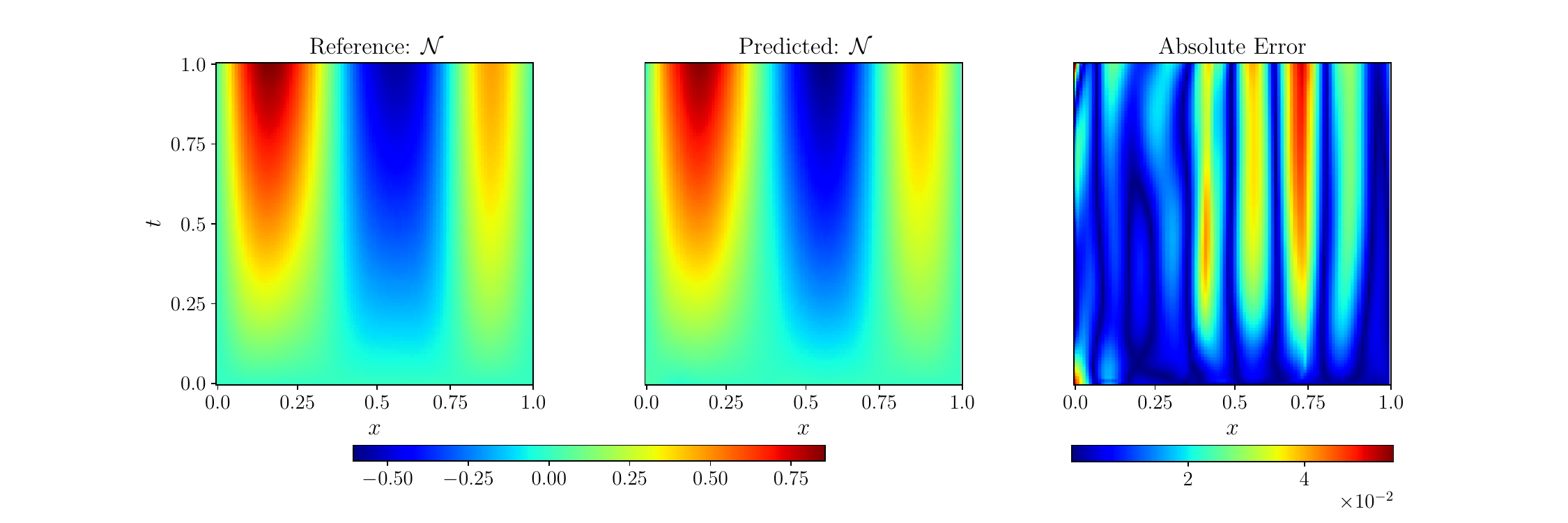}  
    \caption{Sample 1: hidden physics solution comparison, mean absolute error = 0.01435.}
\end{subfigure}
\begin{subfigure}{\textwidth}
    \centering
    \includegraphics[width=0.8\linewidth]{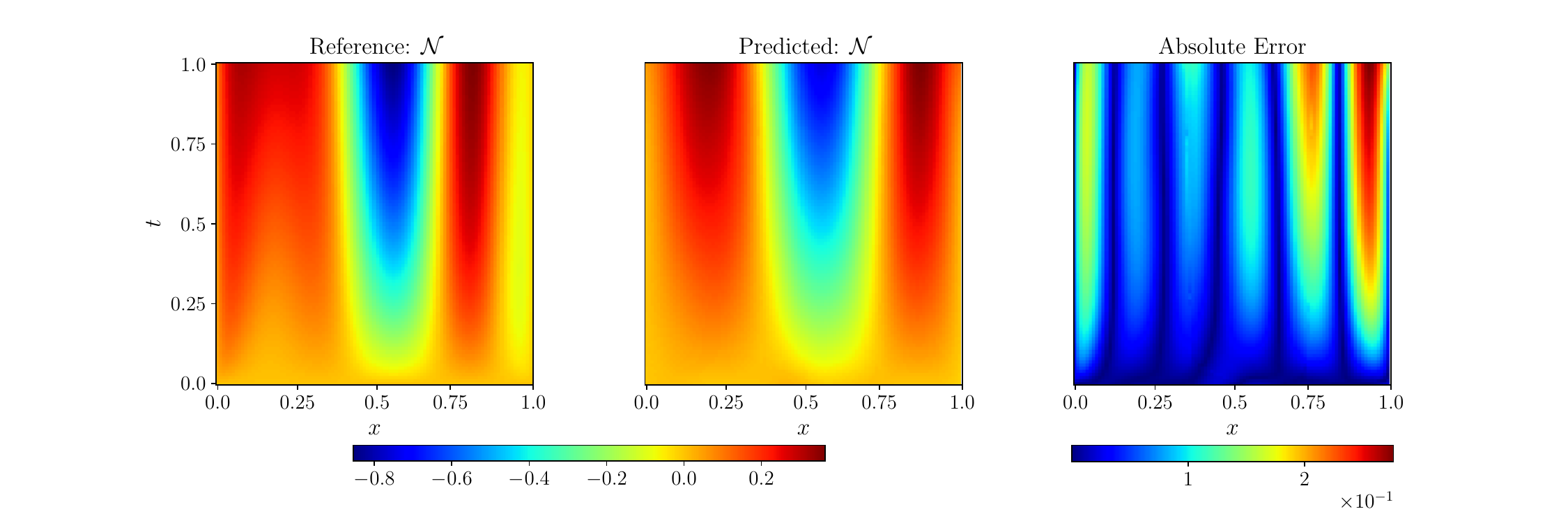}  
    \caption{Sample 2: hidden physics solution comparison, mean absolute error = 0.07512.}
\end{subfigure}
\caption{Reaction diffusion equation: Comparison of reference and predicted PDE terms with input function from Modified GRF function.}
\label{fig:RD_modified_grf_hid_phy_samples}
\end{figure}

\begin{figure}[!htb]
\centering
\begin{subfigure}{\textwidth}
    \centering
    \includegraphics[width=0.8\linewidth]{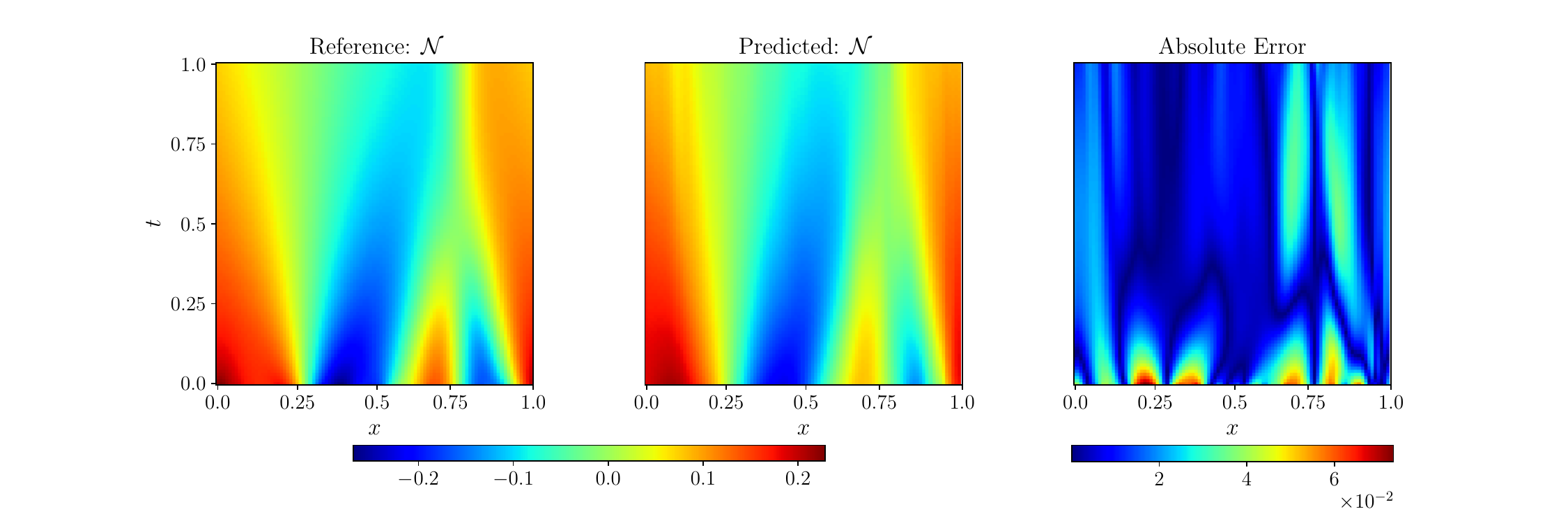}  
    \caption{Sample 1: hidden physics solution comparison, mean absolute error = 0.01267.}
\end{subfigure}
\begin{subfigure}{\textwidth}
    \centering
    \includegraphics[width=0.8\linewidth]{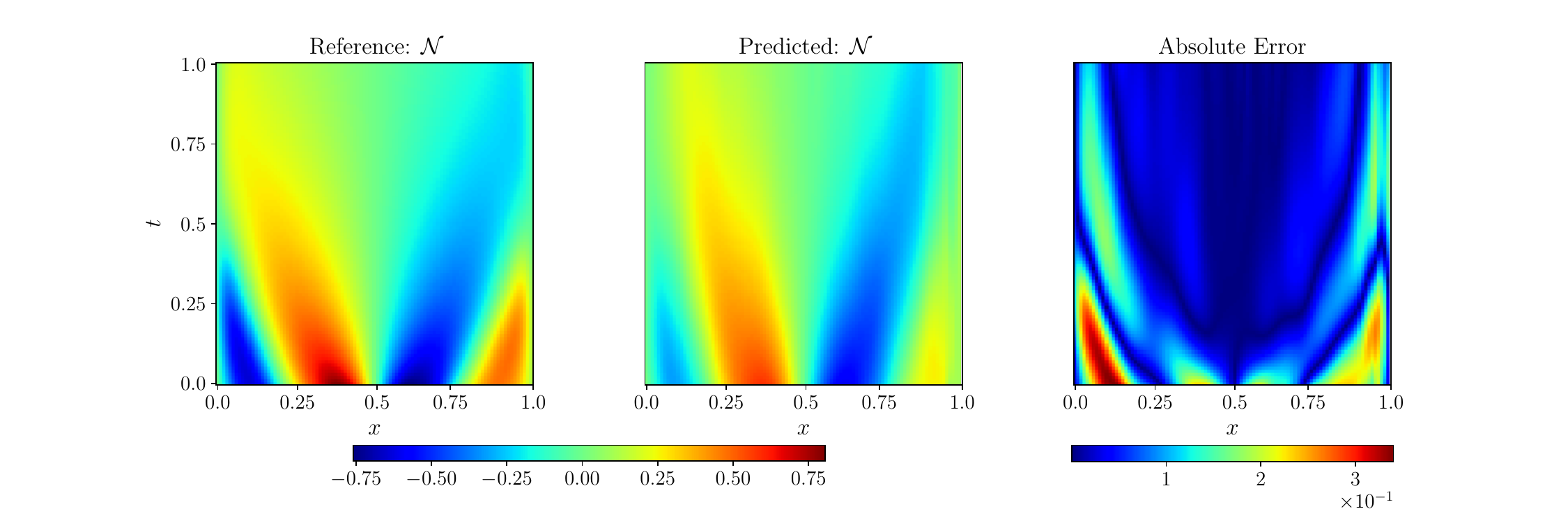}  
    \caption{Sample 2: hidden physics solution comparison, mean absolute error = 0.05602.}
\end{subfigure}
\caption{Burgers' equation: Comparison of reference and predicted PDE terms.}
\label{fig:burgers_hid_phy_samples}
\end{figure}

\begin{figure}[!htb]
\centering
    \centering
    \includegraphics[width=0.8\linewidth]{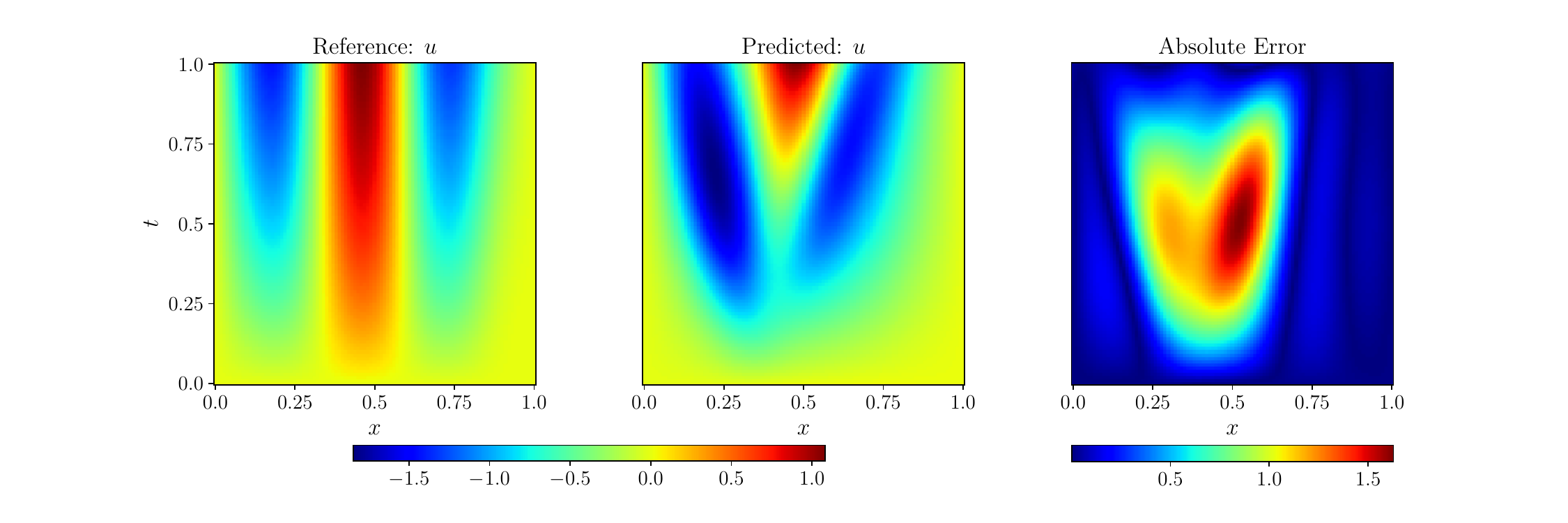}  
    \caption{DeepONet solution: Sparse training solution comparison.  mean absolute error at training time instants $ (t=0,1)$ is 0.03029, mean absolute error at testing time instants $(0<t<1)$ is 0.38671.  
    }
    \label{fig:deeponet_soln}
\end{figure}

\end{document}